\documentclass[11pt]{article}

\usepackage[final]{acl}

\usepackage{times}
\usepackage{latexsym}

\usepackage[T1]{fontenc}
\usepackage[utf8x]{inputenc}

\usepackage{microtype}

\usepackage{inconsolata}

\usepackage{graphicx}

\usepackage{tabularray}
\usepackage{booktabs}
\usepackage{multirow}
\usepackage{xcolor}
\usepackage{colortbl}
\usepackage{subcaption}
\usepackage{wrapfig}
\usepackage{pdflscape}
\usepackage{tcolorbox}
\usepackage{amssymb}
\usepackage{amsmath}
\usepackage{placeins}
\usepackage[inline]{enumitem}

\newtcolorbox{promptbox}[1]{colback=gray!5!white,colframe=gray!75!black,fonttitle=\bfseries\scriptsize,fontupper=\ttfamily\footnotesize,title=#1}

\newcommand{\yes}{\colorbox{green!30}{yes}}
\newcommand{\no}{\colorbox{red!30}{no}}

\title{M$\mathbf5$ -- A Diverse Benchmark to Assess the Performance of Large Multimodal Models Across Multilingual and Multicultural Vision-Language Tasks}

\author{Florian Schneider\textsuperscript{1} \\
  Language Technology Group \\
  Universität Hamburg, Germany \\
  \texttt{\href{mailto:florian.schneider-1@uni-hamburg.de}{florian.schneider-1@uni-hamburg.de}} \\ \And
  Sunayana Sitaram \\
  Microsoft Research India \\
  Bangalore, India \\
  \texttt{\href{mailto:sitaram@microsoft.com}{sitaram@microsoft.com}}
}

\begin{document}
\maketitle
\begin{abstract}
Since the release of ChatGPT, the field of Natural Language Processing has experienced rapid advancements, particularly in Large Language Models (LLMs) and their multimodal counterparts, Large Multimodal Models (LMMs).
Despite their impressive capabilities, LLMs often exhibit significant performance disparities across different languages and cultural contexts, as demonstrated by various text-only benchmarks.
However, current research lacks such benchmarks for multimodal visio-linguistic settings.
This work fills this gap by introducing M5, the first comprehensive benchmark designed to evaluate LMMs on diverse vision-language tasks within a multilingual and multicultural context.
M5 includes eight datasets covering five tasks and $41$ languages, with a focus on underrepresented languages and culturally diverse images.
Furthermore, we introduce two novel datasets, M5-VGR and M5-VLOD, including a new Visio-Linguistic Outlier Detection task, in which all evaluated open-source models fail to significantly surpass the random baseline.
Through extensive evaluation and analyses, we highlight substantial task-agnostic performance disparities between high- and low-resource languages.
Moreover, we show that larger models do not necessarily outperform smaller ones in a multilingual setting.
\end{abstract}
\footnotetext[1]{
    This works was done during a research internship with Microsoft Research India (Bangalore).
}
\section{Introduction}
\label{sec:intro}
Since the release of ChatGPT, Natural Language Processing has experienced a significant surge in interest and research, with a particular focus on LLMs finetuned to follow human instructions.
Besides proprietary models like GPT-4~\citep{achiam2023gpt4}, Claude~\citep{bai2022claude}, or Gemini~\citep{anil2023gemini}, there are also successful open-source variants such as Llama~\citep{touvron2023llama}, Phi~\citep{gunasekar2023phi,abdin2024phi3}, or Mistral~\citep{jiang2023mistral}.
\begin{figure}[!ht]
    \centering
    \includegraphics[width=1.\linewidth]{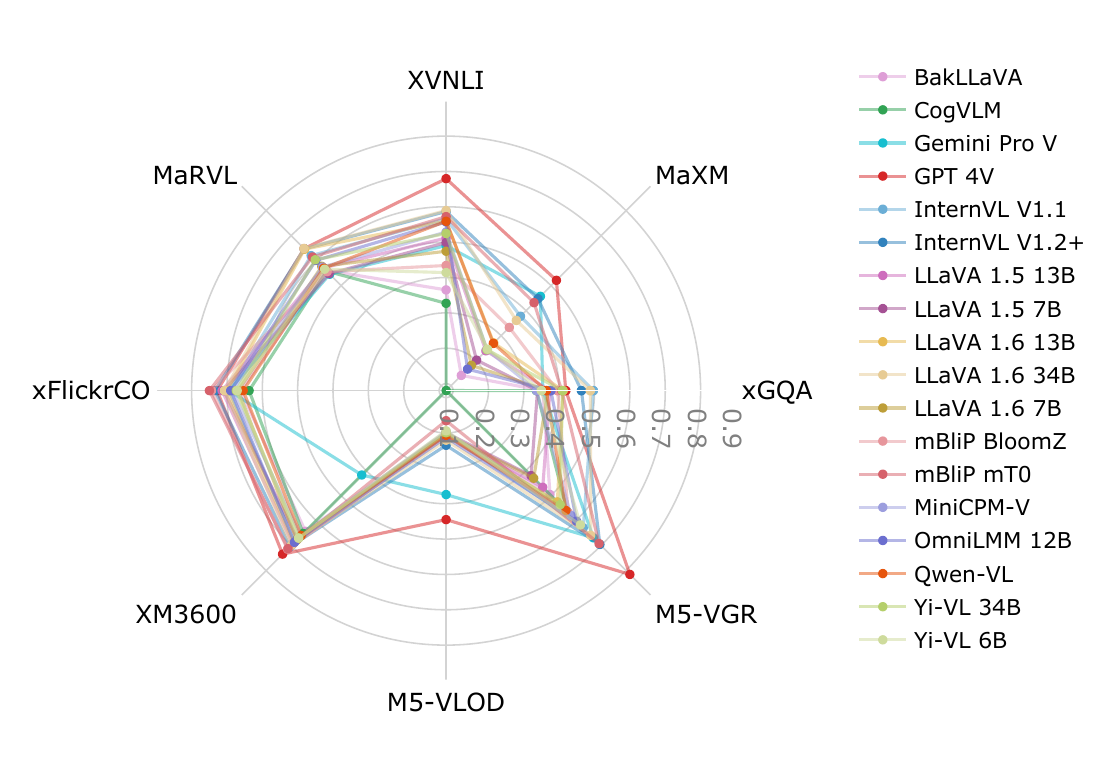}
    \vspace*{-1.2cm}
    \caption{An overview of the average performance of the models on the datasets included in the M5 benchmark. For xFlickrCO and XM3600, we report BERTScore F1. For the other datasets, the accuracy metric is reported.}
    \vspace{-.6cm}
    \label{fig:m5b_fig1_radar}
\end{figure}
While LLMs often demonstrate impressive performance on a wide range of tasks, quantifying and measuring this performance is challenging.
Nevertheless, recent evaluation studies have shown that LLMs generally perform well in English but much worse in other languages~\citep{ahuja2023mega,ahuja2023megaverse,holtermann2024multiq}.

In this work, we focus on multimodal variants of LLMs, Large Multimodal Models (LMMs), such as GPT 4V~\citep{openai2023gpt4v}, Gemini Pro V~\citep{anil2023gemini}, or the popular open-source model, LLaVA~\citep{ liu2023improvedllava,liu2023llava}.
LLMs are not text-only but are also capable of processing images in addition to text.
\begin{figure*}[!ht]
    \centering
    \includegraphics[width=.95\linewidth]{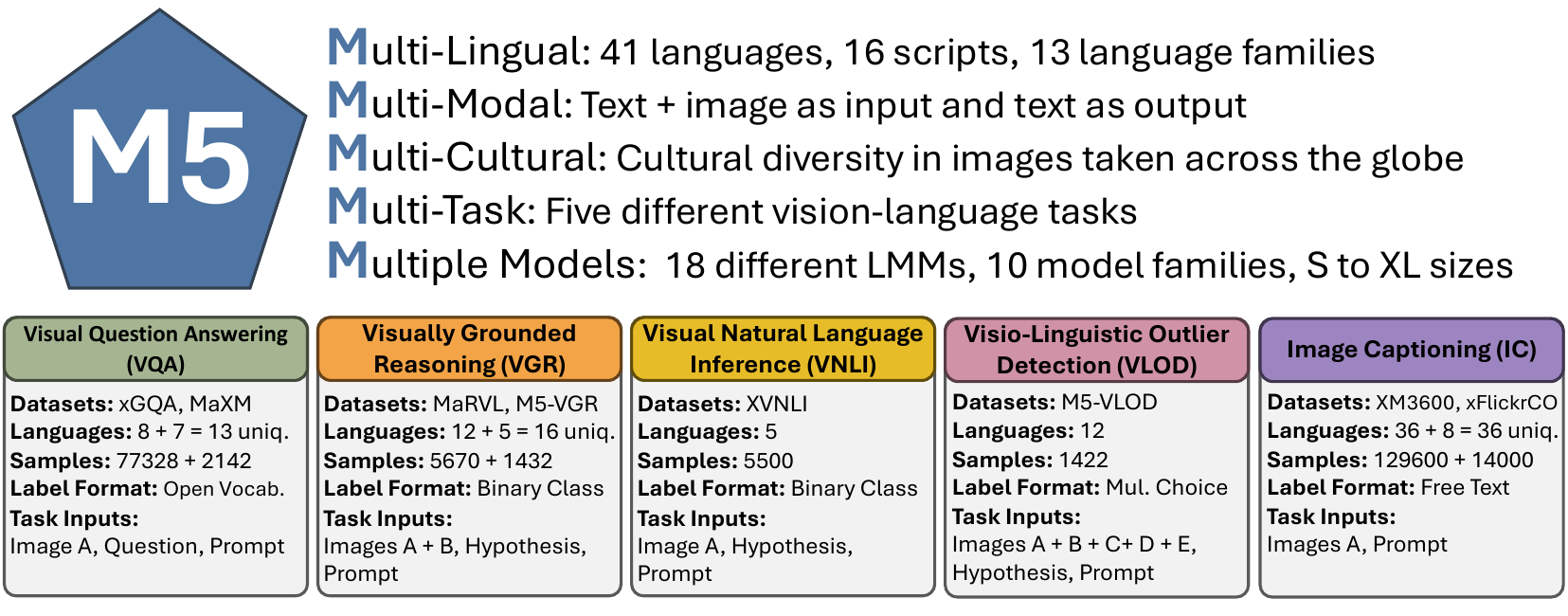}
    \vspace{-.4cm}
    \caption{An informative overview of the M5 Benchmark introduced in this work.}
    \label{fig:m5b_overview}
    \vspace{-.4cm}
\end{figure*}
Most open-source LMMs comprise three major components: an LLM, a vision-encoder model, and a mapping network that projects image embeddings into the text embedding space.
With this architecture, where an LLM serves as the core, we argue that LMMs inherently suffer from the same issue as LLMs: they generally perform much worse in non-English languages.
However, existing benchmarks are either text-only~\cite{ahuja2023mega} or multimodal but monolingual~\cite{yue2023mmmu}, thus unable to prove this hypothesis. 
In other words, current research lacks multimodal multilingual benchmarks to examine LMMs' multilingual capabilities.
In this work, we fill this gap by introducing the M5 Benchmark, taking a significant step towards identifying and measuring the performance disparities of current LMMs between various languages.
Figure~\ref{fig:m5b_overview} and Figure~\ref{fig:m5b_fig1_radar} present a high-level summary of our benchmark.
Moreover, we introduce two new evaluation datasets, including a novel vision-language task.
Both datasets focus on African and Asian cultures, which are underrepresented or even non-existent in previous benchmarks.
Our exhaustive analyses additionally investigate the influence of different factors on the performance, such as the models' size or language fidelity.
%
% \vspace{-0.1cm}
\paragraph{Major Contributions}
The major contributions of this work are
\begin{enumerate*}[label=(\alph*)]
    \item M5, the first multimodal benchmark to assess the performance of current LMMs across five tasks, eight datasets, and $41$ languages;
    \item Two novel datasets spanning $10$ underrepresented African and Asian languages, English and German, with images depicting the respective cultures.
    \item A novel vision-language task: Visio-Linguistic Outlier Detection (VLOD);
    \item A large-scale evaluation of $18$ recent LLMs and a thorough analysis of their multilingual performance.
    \item A public release of our codebase and all datasets in a uniform schema to foster future research for more equitable and accessible LMMs or AI in general\footnote{We will release all code and data upon acceptance.}.
\end{enumerate*}

\section{Related Work}
\label{sec:related_work}
\paragraph{Large Multi-Modal Models}
\label{sec:related_work_lmms}
This work focuses on the multimodal counterpart of large language models (LLMs), often referred to as Large Multimodal Models (LMMs).
LMMs are language models capable of processing and ``understanding'' data other than text.
While this generally subsumes images, video, audio, or more, we concentrate on visio-linguistic LMMs, i.e., models that take text and/or images as input and generate textual output.

The vast majority of open-source LMMs comprise three major components: a pretrained generative LLM as the core, a pretrained vision-encoder model that computes semantically rich image embeddings, and a shallow mapping network that learned to project image embeddings into the text embedding space.
One of this architecture's successful open-source implementations with a recent LLM, i.e., the Llama-based Vicuna~\cite{vicuna2023, touvron2023llama},  is LLaVA~\cite{liu2023llava}, from which many others took inspiration also regarding the training data and process.
Besides this, LMMs also exist, which use Cross-Attention~\cite{wang2023cogvlm, bai2023qwenvl}, Q-Formers~\cite{li2023blip2, geigle2023mblip}, Adapters~\cite{eichenberg2022magma}, or Preceiver Resamplers~\cite{alayrac2022flamingo,awadalla2023openflamingo} to process image embeddings.
For an overview including architectural details and the number of parameters of the $18$ LMMs' components we employed in this work, please see Table~\ref{tab:models_architecture_overview}.

\paragraph{Evaluation Benchmarks}
\label{sec:related_work_benchmarks}
With the recent surge in the research of LLMs and LMMs, analyzing the models' performances is crucial yet challenging.
Popular benchmarks like BIG-Bench~\cite{srivastava2023bigbench}, HELM~\cite{liang2022helm}, or MMLU~\cite{hendrycks2020mmlu} are the defacto-standard to evaluate LLMs on text-only tasks primarily in English.
% 
%However, it is essential to create multilingual LLMs to enable reliable LLM performance for all humans.
%
Efforts like MEGA, MEGAVERSE, or MultiQ~\cite{ahuja2023mega,ahuja2023megaverse,holtermann2024multiq} extended these monolingual benchmarks to a large set of diverse languages and showed that the LLMs' performance in English versus non-English languages differs significantly.

Similarly, efforts have been made to evaluate multimodal models.
Benchmarks like MMMU~\cite{yue2023mmmu}, MME~\cite{fu2023mme}, or MMBench~\cite{liu2023mmbench} assess the performance of LMMs on a vast number of text-image tasks.
However, these benchmarks primarily focus on English, with some tasks available in Chinese.
Like MMMU, there is CMMMU~\cite{zhang2024cmmmu}, which focuses on text-image tasks in Chinese.
Nonetheless, evaluating state-of-the-art LMMs in a massively multilingual large-scale setting remains largely unexplored.
There are only a few multimodal multilingual evaluation datasets (see Section~\ref{sec:m5b_bench_datasets} and~\ref{sec:limits_datasets}) and only two benchmarks: IGLUE~\cite{bugliarello22iglue} and MEGAVERSE.
However, IGLUE evaluates only non-autoregressive transformer-encoders, thus lacking state-of-the-art LLMs.
In MEGAVERSE, only five recent LMMs are evaluated on two datasets.
\section{The M5 Benchmark}
\label{sec:m5b}
This section describes the setup of the M5 Benchmark introduced by this work.
Details about the experimental setup, including prompts and hyperparameters, are reported in Appendix~\ref{appendix:experimental_setup_details}.
\subsection{Models}
\label{sec:m5b_bench_models}
We chose the LMMs included in this benchmark for the following reasons:
Firstly, we focussed on publicly available models released on \href{https://huggingface.co}{Hugging Face} except for GPT-4 Vision and Gemini Pro.
Secondly, we included LMMs well-performing on popular multimodal English-only benchmark
s such as MMMU~\citep{yue2023mmmu} and  MME~\citep{fu2023mme}.
Thirdly, we aimed to cover a mixture of different model families and a broad model size spectrum, including small models with $3$B to $9$B, medium models with $10$B to $19$B, and large models with $20$B to $40$B parameters.
For an overview of all models, including their number of parameters and other architectural details, see Table~\ref{tab:models_architecture_overview}.

\subsection{Datasets}
\label{sec:m5b_bench_datasets}
This section briefly introduces the existing datasets included in our benchmark.
In addition to these, we crafted two novel datasets described in Section~\ref{sec:m5b_datasets}.
For details about the languages covered by the datasets, please refer to Table~\ref{tab:dataset_details}.
\paragraph{xGQA}
\label{sec:m5b_bench_datasets_xgqa}
The xGQA dataset~\citep{pfeiffer2022xgqa} is a cross-lingual visual question-answering dataset.
%
% It extends the well-known English-only GQA dataset~\citep{hudson2019gqa} by manually translating the questions in the balanced test-dev set into seven topologically diverse languages comprising seven language families in five distinct scripts.
%
Each of the $9666$ questions is available in eight languages covering five scripts, while the answers are in English only.
The dataset holds $300$ unique images from Visual Genome~\citep{krishna2017visual_genome}.%, based on the popular MS COCO~\citep{lin2014coco} dataset.
%The dataset holds $300$ unique images from Visual Genome~\citep{krishna2017visual_genome}, based on the popular MS COCO~\citep{lin2014coco} dataset.
%
%%%%%%%%%%%%%
%
\paragraph{MaXM}
\label{sec:m5b_bench_datasets_maxm}
The MaXM dataset was introduced by~\cite{changpinyo2023maxm} and is a VQA dataset comprising seven languages in five scripts.
In MaXM, the questions and their respective answers are in the same language.
%
%Moreover, in MaXM, the images are a subset of the XM3600~\citep{thapliyal2022xm3600} dataset and are chosen to match a region where the language of the question-answer pair is spoken.
%
To increase cultural diversity, the images were selected to match the region where the target language is spoken.
%
% This also ensures cultural diversity in the images in addition to the language diversity in the question-answer texts.
%
%Moreover, the images are therefore out-of-domain, considering typical VQA benchmarks~\citep{ren2015vqa,krishna2017visual_genome,hudson2019gqa,marino2019okvqa} are primarily based on COCO images.
%
%%%%%%%%%%%%%%
%
\paragraph{XVNLI}
\label{sec:m5b_bench_datasets_xvnli}
The XVNLI dataset~\cite{bugliarello22iglue} introduces the task of Cross-lingual Visual Natural Language Inference where a model needs to predict whether a textual hypothesis \textit{entails}, \textit{contradicts}, or is \textit{neutral} concerning a visual premise.
XVNLI comprises five languages covering three scripts and $357$ unique images from Visual Genome.
%
%It is based on a combination of the text-only SNLI~\citep{bowman2015snli} dataset and its cross-lingual~\citep{agic2018cli} and cross-modal~\citep{xie2019visual_entailment} equivalents.
%
%%%%%%%%%%%%%%%
%
\paragraph{MaRVL}
\label{sec:m5b_bench_datasets_marvl}
The MaRVL dataset~\cite{liu2021marvl} aims to benchmark models on Multicultural Reasoning over Vision and Language.
A task sample comprises two images, a textual statement, and a binary true or false answer grounded in the images.
MaRVL comprises five languages covering three scripts and $4914$ culturally diverse images that match the respective languages.
%
%The unique characteristic of MaRVL is that the images in a sample are chosen to match the culture of the annotator who has written the textual statement in his or her native language.
%
%
%%%%%%%%%%%%%%%
%
\paragraph{XM3600}
\label{sec:m5b_bench_datasets_xm3600}
The XM3600 dataset~\cite{thapliyal2022xm3600} is a large multilingual image captioning dataset comprising $36$ languages with $261375$ captions for $100$ unique images per language.
The images are selected to match the language's cultural background, ensuring cultural and linguistic diversity.
%
%Notably, the captions are not automatically translated but were manually created by professional annotators who are native speakers of the respective language.
%
%Further, the captions are written with the respective script, covering $13$ different scripts.
%
%%%%%%%%%%%%%%%%
%
\paragraph{xFlickrCO}
\label{sec:m5b_bench_datasets_xflickco}
The xFlickrCO dataset~\citep{bugliarello22iglue} is an image captioning dataset and comprises $1000$ images from Flickr30k~\citep{young2014flickr30k} and $1000$ images from COCO~\citep{lin2014coco}.
Each image is captioned in eight languages, covering four different scripts.
%
% Notably, for all languages except English and German, the captions were manually crafted by crowdsourcing workers instead of translated from English to prevent bias and increase linguistic diversity.
%

%
\section{Novel M5 Datasets}
\label{sec:m5b_datasets}
%
% \begin{figure*}[ht!]
%      \centering
%      \begin{subfigure}[b]{.475\linewidth}
%         \includegraphics[width=1.\linewidth]{gfx/vgr_sample_zu_90_4bbacd9003aa4d0199939fa2fd80c276.png}
%         \caption{A Zulu sample of the novel M5-VGR dataset. \textbf{Hypothesis:} \textit{``Isithombe sokuqala nesithombe sesibili sibonisa iqanda elisehhokweni. (The first picture and the second picture show the egg on the head.)}'', \textbf{Label:} \textit{False}}
%         \label{fig:m5b_datasets_vgr_sample}
%      \end{subfigure}
%      \hspace{.5cm}
%      \begin{subfigure}{.475\linewidth}
%         \includegraphics[width=1.\linewidth]{gfx/vlod_sample_sw_87_9fbc54f2a62f4cbbbe216e84565cbdd6.jpg}
%         % \textbf{Topic:} ``\textit{soap for hands and body}'', 
%         \vspace{.55cm}
%         \caption{A Swahili sample of the novel M5-VLOD dataset. \textbf{Hypothesis:} ``\textit{Picha zote zinaonyesha sabuni inayotumika kwa mikono na mwili bila mtu yeyote. (All the images show soap applied to the hands and body without anyone.)}'', \textbf{Outlier:} $1$.}
%         \label{fig:m5b_datasets_vlod_sample}
%      \end{subfigure}
%      \caption{Examples for the two novel datasets, M5-VGR and M5-VLOD, introduced by this work.}
% \end{figure*}
%
In addition to the existing datasets introduced in the previous section, we crafted two novel multimodal and multilingual evaluation datasets.
The principal motivation behind this is to fill the gap in existing vision-language datasets concerning the lack of underrepresented languages, tasks, and cultural diversity.
Moreover, we aim to enable further examination of LMMs and their performance on non-English and non-Western data with a particular focus on African and Asian regions.
Details, statistics, and examples are reported in Appendix~\ref{appendix:dataset_details}.

%
%In the following, first, common characteristics and the general collection process are explained.
%
%Then, the two datasets are introduced individually.
%
%More details, statistics, and examples are reported in Appendix~\ref{appendix:dataset_details}.
%
%
\subsection*{Common Characteristics}
\label{sec:m5b_datasets_character}
\paragraph{Languages}
Both datasets comprise samples in $12$ languages covering seven scripts (see Table~\ref{tab:dataset_details}):
\textit{Amharic}, \textit{Berber}, \textit{Bengali}, \textit{German}, \textit{English}, \textit{Filipino}, \textit{Hausa}, \textit{Hindi}, \textit{Russian}, \textit{Swahili}, \textit{Thai}, \textit{Zulu}.
The languages were selected to enrich the set of languages covered by existing datasets, focusing on underrepresented languages from Asian and African countries or regions. 
To our knowledge, no other visio-linguistic evaluation dataset covers Amharic, Berber, Hausa, or Zulu.

\paragraph{Depicting Cultural Diversity}
The images in our datasets originate from the Dollar Street dataset~\citep{gaviria2022dollar}, comprising around $38K$ photos taken in $63$ different regions or countries around the globe.
These photos depict the lives of families, including their homes, neighborhoods, or everyday objects, in a culturally diverse way.
Further, each image in the original dataset is tagged with one or more ``topics'' that roughly describe its visual content.

\paragraph{Image Basis}
For our datasets, we sampled a subset of images from the dataset taken in regions where our $12$ target languages are spoken.
In this subset, which forms the visual basis for both of our datasets and is referred to as $\mathbb{B}$, each image $i_{l}^{t} \in \mathbb{B}$ is tagged with exactly one topic $t \in \mathbb{T} = \{t_0, \dots, t_{86}\}$ and was taken in a region $r_l$ where language $l \in \mathbb{L} = \{l_0, \dots, l_{11}\}$ is spoken.

\subsection{M5-VGR}
\label{sec:m5b_datasets_vgr}
\begin{figure}[!ht]
    \centering
    %\textbf{Topic:} ``\textit{street view}'', 
    \includegraphics[width=.9\linewidth]{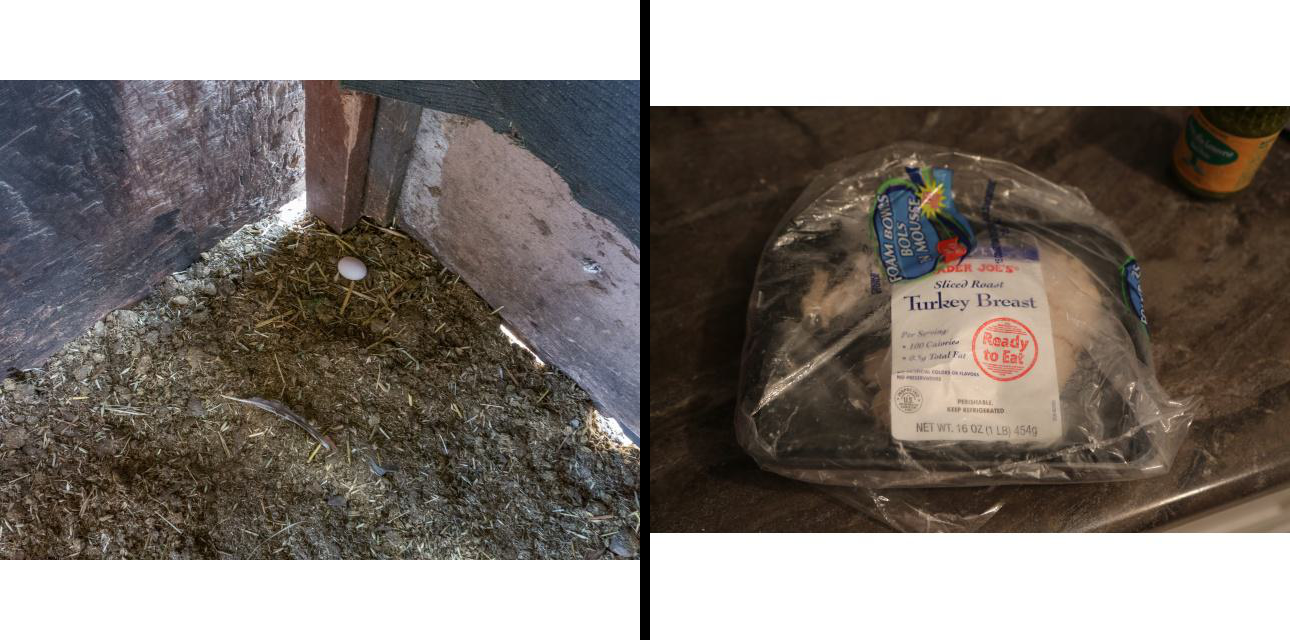}
     \caption{An Zulu example of the novel M5-VGR dataset. \textbf{Hypothesis:} \textit{``Isithombe sokuqala nesithombe sesibili sibonisa iqanda elisehhokweni. (The first picture and the second picture show the egg on the head.)}'', \textbf{Label:} \textit{False}}
    \label{fig:m5b_datasets_vgr_sample}
    \vspace{-.3cm}
\end{figure}
Inspired by MaRVL, the goal of the M5-VGR dataset is to provide a visually grounded reasoning (VGR) evaluation dataset that covers a wide range of topologically different languages and, at the same time, visually represents a diverse set of cultures in which the respective languages are spoken.
However, since the MaRVL dataset contains only five languages, we chose $11$ additional topologically diverse languages for our dataset.
To guarantee visual and linguistic diversity and high data quality in our dataset, we hired professional native-speaker annotators of the respective languages to annotate the data.
Moreover, we performed several rounds of data quality assessment in close collaboration with the annotators.

A task sample $s$ in M5-VGR contains two images $i_a$ and $i_b$, a textual visually grounded hypothesis $h$, and a binary label $c$ which is either true or false concerning the two visual premises (see Figure~\ref{fig:m5b_datasets_vgr_sample}).
More specifically, for each language $l \in \mathbb{L}$, we created $120$  tasks $s_l \in \mathbb{S}_l$ as follows:
In the first step, we sampled $120$ unique images $a_{l}^{t} \in \mathbb{B}$ from our image basis so that each topic $t \in \mathbb{T}$ occurs at least once across all $12$ languages.
Then, for each of the $120$ images, we randomly selected another image $b_{l_2}^{t} \in \mathbb{B}$ associated with another language $l_2 \ne l \in \mathbb{L}$ that shares the topic $t$.
In the third step, we asked the native-speaker annotators of the language $l$ to manually create a hypothesis $h$ and a label $c$ which is either true or false concerning the image premises $\left(a_{l}^{t}, b_{l_2}^{t}\right)$.
Further, the annotators were instructed to generate a hypothesis semantically related to the topic $t$ if possible.
%
%With this, we aim to enable a fine-granular topic-wise evaluation of LMMs (see Section~\ref{XXX}).
%

\subsection{M5-VLOD}
\label{sec:m5b_datasets_vlod}
\begin{figure}[!ht]
    \centering
    \includegraphics[width=1\linewidth]{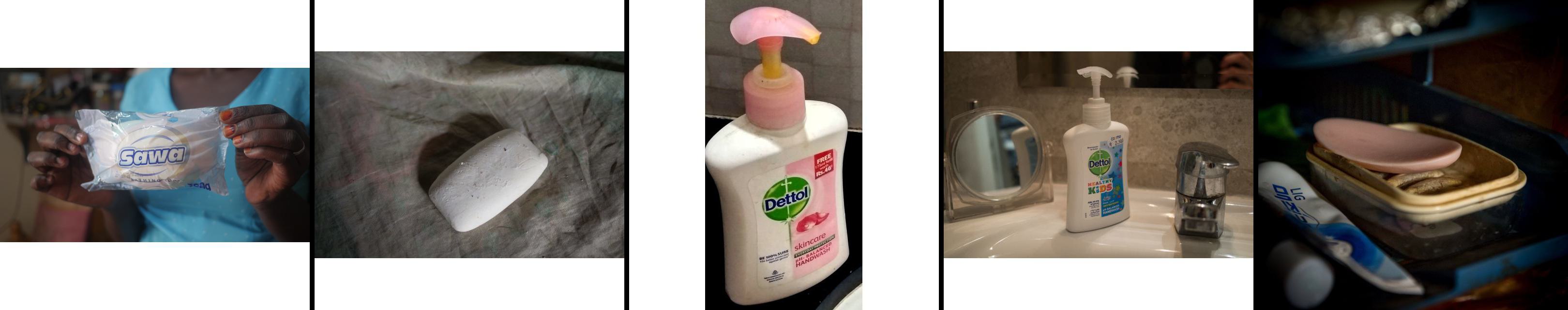}
    % \textbf{Topic:} ``\textit{soap for hands and body}'', 
     \caption{A Swahili example of the novel M5-VLOD dataset. \textbf{Hypothesis:} ``\textit{Picha zote zinaonyesha sabuni inayotumika kwa mikono na mwili bila mtu yeyote. (All the images show soap applied to the hands and body without anyone.)}'', \textbf{Outlier:} $1$.}
    \label{fig:m5b_datasets_vlod_sample}
    \vspace{-.3cm}
\end{figure}
With the M5-VLOD dataset, we introduce a novel multimodal task: Visio-Linguistic Outlier Detection.
The objective of the task is to detect an outlier image from a set of images considering a textual statement.
An example of the task is shown in Figure~\ref{fig:m5b_datasets_vlod_sample}, where five images related to the topic ``soap for hands and body'' are shown.
The machine-translated English statement is: ``All the images show soap applied to the hands and body without anyone.''.
Because only the first image shows a person, the statement is incorrect for the first image and, therefore, is considered the outlier image.
%
%Note that we can decrease the task's difficulty by keeping the outlier image but reducing the number of matching images.
%

The dataset was collected similarly to M5-VGR, as described in the previous section.
The major difference is that instead of sampling only one image in the second step, we sample four images so that a sample $s_{l_0} \in \mathbb{S}_{l_0}$ for language $l_0 \in \mathbb{L}$ comprises of five images: $\{a_{l_0}^{t}, b_{l_1}^{t}, c_{l_2}^{t}, d_{l_3}^{t}, e_{l_4}^{t},\}$ associated with five different languages $\{l_0, \dots, l_4 \in \mathbb{L}\}$ that share one topic $t \in \mathbb{T}$.
In the third step, we asked the native-speaker annotators of the language $l$ to manually create a textual statement $h$, valid for all but one of the images labeled as the outlier image.
\section{General Results Discussion}
\label{sec:results_general}
\begin{table*}[!ht]
\tiny
\centering
% \addtolength{\tabcolsep}{-0.125em}
\addtolength{\tabcolsep}{-0.466em}
\begin{tabular}{@{}lcc||cc||cc||cc||cc||cc||cc||cc||cc||c@{}}
\toprule
\textbf{Model} & \multicolumn{19}{c}{\textbf{Dataset}} \\ \midrule
 &
  \multicolumn{2}{c}{\textbf{xGQA}} &
  \multicolumn{2}{c}{\textbf{MaXM}} &
  \multicolumn{2}{c}{\textbf{XVNLI}} &
  \multicolumn{2}{c}{\textbf{MaRVL}} &
  \multicolumn{2}{c}{\textbf{M5-VLOD}} &
  \multicolumn{2}{c}{\textbf{M5-VGR}} &
  \multicolumn{2}{c}{\textbf{xFlickrCO}} &
  \multicolumn{2}{c}{\textbf{XM3600}} &
  \multicolumn{3}{c}{\textbf{ALL}} \\
 &
  \multicolumn{1}{c}{E} &
  \multicolumn{1}{c}{NE} &
  \multicolumn{1}{c}{E} &
  \multicolumn{1}{c}{NE} &
  \multicolumn{1}{c}{E} &
  \multicolumn{1}{c}{NE} &
  \multicolumn{1}{c}{E} &
  \multicolumn{1}{c}{NE} &
  \multicolumn{1}{c}{E} &
  \multicolumn{1}{c}{NE} &
  \multicolumn{1}{c}{E} &
  \multicolumn{1}{c}{NE} &
  \multicolumn{1}{c}{E} &
  \multicolumn{1}{c}{NE} &
  \multicolumn{1}{c}{E} &
  \multicolumn{1}{c}{NE} &
  \multicolumn{1}{c}{E} &
  \multicolumn{1}{c}{NE} &
  \multicolumn{1}{c}{$\Delta$} \\ \midrule
CogVLM &
  \cellcolor[HTML]{9ACE7F}$0.59$ &
  \cellcolor[HTML]{F86A6B}$0.30$ &
  \cellcolor[HTML]{FEEB84}$0.43$ &
  \cellcolor[HTML]{F8696B}$0.02$ &
  \cellcolor[HTML]{FCC07B}$0.47$ &
  \cellcolor[HTML]{F8696B}$0.29$ &
  \cellcolor[HTML]{FFEB84}$0.60$ &
  \cellcolor[HTML]{F8696B}$0.51$ &
  \cellcolor[HTML]{F97C6E}$0.10$ &
  \cellcolor[HTML]{F8696B}$0.08$ &
  \cellcolor[HTML]{F5E984}$0.68$ &
  \cellcolor[HTML]{FBAF78}$0.55$ &
  \cellcolor[HTML]{9BCE7F}$0.87$ &
  \cellcolor[HTML]{F8696B}$0.60$ &
  \cellcolor[HTML]{76C47D}$0.88$ &
  \cellcolor[HTML]{FDCD7E}$0.65$ &
  \cellcolor[HTML]{FDEB84}$0.58$ &
  \cellcolor[HTML]{F8696B}$0.38$ &
  \cellcolor[HTML]{F8786D}$-0.20$ \\
BakLLaVA &
  \cellcolor[HTML]{87C97E}$0.62$ &
  \cellcolor[HTML]{F8786E}$0.32$ &
  \cellcolor[HTML]{A6D27F}$0.53$ &
  \cellcolor[HTML]{F87A6E}$0.08$ &
  \cellcolor[HTML]{FCC57C}$0.48$ &
  \cellcolor[HTML]{F97F6F}$0.34$ &
  \cellcolor[HTML]{FEE382}$0.59$ &
  \cellcolor[HTML]{F97C6E}$0.53$ &
  \cellcolor[HTML]{FBAB77}$0.14$ &
  \cellcolor[HTML]{FFEB84}$0.20$ &
  \cellcolor[HTML]{E3E383}$0.71$ &
  \cellcolor[HTML]{F98871}$0.48$ &
  \cellcolor[HTML]{6EC27C}$0.91$ &
  \cellcolor[HTML]{F8786D}$0.63$ &
  \cellcolor[HTML]{76C47D}$0.88$ &
  \cellcolor[HTML]{FDC97D}$0.64$ &
  \cellcolor[HTML]{D8E082}$0.61$ &
  \cellcolor[HTML]{F8786E}$0.40$ &
  \cellcolor[HTML]{F8696B}$-0.21$ \\
LLaVA 1.6 7B &
  \cellcolor[HTML]{94CC7E}$0.60$ &
  \cellcolor[HTML]{F98971}$0.34$ &
  \cellcolor[HTML]{FDCE7E}$0.34$ &
  \cellcolor[HTML]{FA9473}$0.16$ &
  \cellcolor[HTML]{D7E082}$0.59$ &
  \cellcolor[HTML]{FCB479}$0.45$ &
  \cellcolor[HTML]{D0DE82}$0.62$ &
  \cellcolor[HTML]{F97D6E}$0.53$ &
  \cellcolor[HTML]{FBA776}$0.14$ &
  \cellcolor[HTML]{FEEB84}$0.21$ &
  \cellcolor[HTML]{FBAE78}$0.55$ &
  \cellcolor[HTML]{F86B6B}$0.42$ &
  \cellcolor[HTML]{8FCB7E}$0.88$ &
  \cellcolor[HTML]{F9806F}$0.64$ &
  \cellcolor[HTML]{76C47D}$0.88$ &
  \cellcolor[HTML]{FED880}$0.67$ &
  \cellcolor[HTML]{FEEA83}$0.57$ &
  \cellcolor[HTML]{F98971}$0.43$ &
  \cellcolor[HTML]{F4E884}$-0.15$ \\
LLaVA 1.5 7B &
  \cellcolor[HTML]{83C77D}$0.62$ &
  \cellcolor[HTML]{F8696B}$0.30$ &
  \cellcolor[HTML]{ACD380}$0.52$ &
  \cellcolor[HTML]{FA9172}$0.15$ &
  \cellcolor[HTML]{D1DE82}$0.60$ &
  \cellcolor[HTML]{FCC37C}$0.47$ &
  \cellcolor[HTML]{FDC97D}$0.57$ &
  \cellcolor[HTML]{F86C6B}$0.52$ &
  \cellcolor[HTML]{FCB87A}$0.15$ &
  \cellcolor[HTML]{FEEA83}$0.20$ &
  \cellcolor[HTML]{F98770}$0.48$ &
  \cellcolor[HTML]{F8696B}$0.42$ &
  \cellcolor[HTML]{65BF7C}$0.92$ &
  \cellcolor[HTML]{FA9C74}$0.68$ &
  \cellcolor[HTML]{72C37C}$0.89$ &
  \cellcolor[HTML]{FDD780}$0.67$ &
  \cellcolor[HTML]{E9E583}$0.59$ &
  \cellcolor[HTML]{F98971}$0.43$ &
  \cellcolor[HTML]{FDCD7E}$-0.17$ \\
Yi-VL 6B &
  \cellcolor[HTML]{A5D17F}$0.57$ &
  \cellcolor[HTML]{F8796E}$0.32$ &
  \cellcolor[HTML]{A6D27F}$0.53$ &
  \cellcolor[HTML]{FBA275}$0.20$ &
  \cellcolor[HTML]{FFEB84}$0.56$ &
  \cellcolor[HTML]{FA9373}$0.38$ &
  \cellcolor[HTML]{FEE883}$0.59$ &
  \cellcolor[HTML]{F98770}$0.53$ &
  \cellcolor[HTML]{FFEB84}$0.20$ &
  \cellcolor[HTML]{FEDF81}$0.19$ &
  \cellcolor[HTML]{D7E082}$0.73$ &
  \cellcolor[HTML]{FDCB7D}$0.61$ &
  \cellcolor[HTML]{6BC17C}$0.91$ &
  \cellcolor[HTML]{F97F6F}$0.64$ &
  \cellcolor[HTML]{64BF7C}$0.91$ &
  \cellcolor[HTML]{FDD57F}$0.66$ &
  \cellcolor[HTML]{C3DA81}$0.62$ &
  \cellcolor[HTML]{FA9373}$0.44$ &
  \cellcolor[HTML]{FBA476}$-0.18$ \\
MiniCPM-V &
  \cellcolor[HTML]{B9D780}$0.55$ &
  \cellcolor[HTML]{F8706C}$0.31$ &
  \cellcolor[HTML]{89C97E}$0.56$ &
  \cellcolor[HTML]{FA9F75}$0.19$ &
  \cellcolor[HTML]{80C77D}$0.66$ &
  \cellcolor[HTML]{FDCC7E}$0.49$ &
  \cellcolor[HTML]{E7E483}$0.61$ &
  \cellcolor[HTML]{F98971}$0.53$ &
  \cellcolor[HTML]{FEE282}$0.20$ &
  \cellcolor[HTML]{FEE282}$0.20$ &
  \cellcolor[HTML]{A0D07F}$0.80$ &
  \cellcolor[HTML]{FBB379}$0.56$ &
  \cellcolor[HTML]{6BC17C}$0.91$ &
  \cellcolor[HTML]{F98A71}$0.65$ &
  \cellcolor[HTML]{6CC17C}$0.90$ &
  \cellcolor[HTML]{FDD17F}$0.65$ &
  \cellcolor[HTML]{A7D27F}$0.65$ &
  \cellcolor[HTML]{FA9874}$0.45$ &
  \cellcolor[HTML]{F97D6E}$-0.20$ \\
LLaVA 1.5 13B &
  \cellcolor[HTML]{85C87D}$0.62$ &
  \cellcolor[HTML]{F98C71}$0.34$ &
  \cellcolor[HTML]{89C97E}$0.56$ &
  \cellcolor[HTML]{FA9D75}$0.19$ &
  \cellcolor[HTML]{DCE182}$0.59$ &
  \cellcolor[HTML]{FDCC7E}$0.49$ &
  \cellcolor[HTML]{FEEA83}$0.60$ &
  \cellcolor[HTML]{FA9573}$0.54$ &
  \cellcolor[HTML]{FCC07B}$0.16$ &
  \cellcolor[HTML]{FFEB84}$0.21$ &
  \cellcolor[HTML]{FCB679}$0.57$ &
  \cellcolor[HTML]{F97E6F}$0.46$ &
  \cellcolor[HTML]{69C07C}$0.91$ &
  \cellcolor[HTML]{FBA175}$0.69$ &
  \cellcolor[HTML]{6BC17C}$0.90$ &
  \cellcolor[HTML]{FEDF81}$0.69$ &
  \cellcolor[HTML]{D2DE82}$0.61$ &
  \cellcolor[HTML]{FA9974}$0.45$ &
  \cellcolor[HTML]{FEDA80}$-0.16$ \\
Qwen-VL &
  \cellcolor[HTML]{98CE7F}$0.59$ &
  \cellcolor[HTML]{F98570}$0.33$ &
  \cellcolor[HTML]{BBD881}$0.50$ &
  \cellcolor[HTML]{FBAC77}$0.23$ &
  \cellcolor[HTML]{B3D580}$0.62$ &
  \cellcolor[HTML]{FEE482}$0.54$ &
  \cellcolor[HTML]{F0E784}$0.60$ &
  \cellcolor[HTML]{F9826F}$0.53$ &
  \cellcolor[HTML]{FCBC7B}$0.16$ &
  \cellcolor[HTML]{FDEB84}$0.21$ &
  \cellcolor[HTML]{94CD7E}$0.82$ &
  \cellcolor[HTML]{FBA676}$0.54$ &
  \cellcolor[HTML]{88C97E}$0.89$ &
  \cellcolor[HTML]{F8746D}$0.62$ &
  \cellcolor[HTML]{66BF7C}$0.90$ &
  \cellcolor[HTML]{FDD07E}$0.65$ &
  \cellcolor[HTML]{B5D680}$0.64$ &
  \cellcolor[HTML]{FA9D75}$0.46$ &
  \cellcolor[HTML]{FBB078}$-0.18$ \\
Yi-VL 34B &
  \cellcolor[HTML]{A2D17F}$0.58$ &
  \cellcolor[HTML]{FCB579}$0.38$ &
  \cellcolor[HTML]{A1D07F}$0.53$ &
  \cellcolor[HTML]{FBA075}$0.20$ &
  \cellcolor[HTML]{D7E082}$0.59$ &
  \cellcolor[HTML]{FDD37F}$0.51$ &
  \cellcolor[HTML]{CCDD82}$0.62$ &
  \cellcolor[HTML]{FDD47F}$0.58$ &
  \cellcolor[HTML]{EEE784}$0.26$ &
  \cellcolor[HTML]{FEE082}$0.19$ &
  \cellcolor[HTML]{B8D780}$0.77$ &
  \cellcolor[HTML]{FA9C74}$0.52$ &
  \cellcolor[HTML]{6AC07C}$0.91$ &
  \cellcolor[HTML]{F9816F}$0.64$ &
  \cellcolor[HTML]{68C07C}$0.90$ &
  \cellcolor[HTML]{FDD47F}$0.66$ &
  \cellcolor[HTML]{AAD380}$0.65$ &
  \cellcolor[HTML]{FAA075}$0.46$ &
  \cellcolor[HTML]{FBA275}$-0.19$ \\
Gemini Pro V &
  \cellcolor[HTML]{F5E984}$0.46$ &
  \cellcolor[HTML]{F98C71}$0.34$ &
  \cellcolor[HTML]{CFDE82}$0.48$ &
  \cellcolor[HTML]{FBAB77}$0.23$ &
  \cellcolor[HTML]{FDC97D}$0.49$ &
  \cellcolor[HTML]{FDCB7D}$0.49$ &
  \cellcolor[HTML]{FA9B74}$0.55$ &
  \cellcolor[HTML]{FA9C74}$0.55$ &
  \cellcolor[HTML]{9BCF7F}$0.52$ &
  \cellcolor[HTML]{CEDD82}$0.36$ &
  \cellcolor[HTML]{A6D27F}$0.79$ &
  \cellcolor[HTML]{FEE582}$0.66$ &
  \cellcolor[HTML]{ABD380}$0.86$ &
  \cellcolor[HTML]{FA9673}$0.67$ &
  \cellcolor[HTML]{FDC87D}$0.63$ &
  \cellcolor[HTML]{F8696B}$0.41$ &
  \cellcolor[HTML]{E4E483}$0.60$ &
  \cellcolor[HTML]{FBA175}$0.46$ &
  \cellcolor[HTML]{DAE182}$-0.13$ \\
OmniLMM 12B &
  \cellcolor[HTML]{E1E383}$0.49$ &
  \cellcolor[HTML]{FA9F75}$0.36$ &
  \cellcolor[HTML]{D2DE82}$0.48$ &
  \cellcolor[HTML]{F98670}$0.11$ &
  \cellcolor[HTML]{98CE7F}$0.64$ &
  \cellcolor[HTML]{FEE082}$0.54$ &
  \cellcolor[HTML]{AAD380}$0.64$ &
  \cellcolor[HTML]{FBAA77}$0.56$ &
  \cellcolor[HTML]{FEDA80}$0.19$ &
  \cellcolor[HTML]{FCEB84}$0.21$ &
  \cellcolor[HTML]{ACD480}$0.78$ &
  \cellcolor[HTML]{FCC07B}$0.59$ &
  \cellcolor[HTML]{6FC27C}$0.91$ &
  \cellcolor[HTML]{FA8E72}$0.66$ &
  \cellcolor[HTML]{74C37C}$0.89$ &
  \cellcolor[HTML]{FEDC81}$0.68$ &
  \cellcolor[HTML]{C0D981}$0.63$ &
  \cellcolor[HTML]{FBA175}$0.46$ &
  \cellcolor[HTML]{FDD880}$-0.16$ \\
LLaVA 1.6 13B &
  \cellcolor[HTML]{71C37C}$0.65$ &
  \cellcolor[HTML]{FBAE78}$0.38$ &
  \cellcolor[HTML]{E0E383}$0.46$ &
  \cellcolor[HTML]{FBAF78}$0.24$ &
  \cellcolor[HTML]{C2DA81}$0.61$ &
  \cellcolor[HTML]{FEE683}$0.55$ &
  \cellcolor[HTML]{9CCF7F}$0.65$ &
  \cellcolor[HTML]{9ECF7F}$0.65$ &
  \cellcolor[HTML]{FBA776}$0.14$ &
  \cellcolor[HTML]{FEEB84}$0.21$ &
  \cellcolor[HTML]{B2D580}$0.78$ &
  \cellcolor[HTML]{FA9573}$0.50$ &
  \cellcolor[HTML]{7EC67D}$0.90$ &
  \cellcolor[HTML]{FA9773}$0.67$ &
  \cellcolor[HTML]{77C47D}$0.88$ &
  \cellcolor[HTML]{FEDE81}$0.68$ &
  \cellcolor[HTML]{BAD881}$0.63$ &
  \cellcolor[HTML]{FBB078}$0.48$ &
  \cellcolor[HTML]{EFE784}$-0.15$ \\
mBliP BloomZ &
  \cellcolor[HTML]{FEE983}$0.44$ &
  \cellcolor[HTML]{FCB87A}$0.39$ &
  \cellcolor[HTML]{95CD7E}$0.55$ &
  \cellcolor[HTML]{FCBF7B}$0.29$ &
  \cellcolor[HTML]{FAA075}$0.40$ &
  \cellcolor[HTML]{FBB279}$0.44$ &
  \cellcolor[HTML]{FBA276}$0.55$ &
  \cellcolor[HTML]{FCB77A}$0.56$ &
  \cellcolor[HTML]{FBAE78}$0.14$ &
  \cellcolor[HTML]{FDEB84}$0.21$ &
  \cellcolor[HTML]{EFE784}$0.69$ &
  \cellcolor[HTML]{FBB279}$0.56$ &
  \cellcolor[HTML]{65BF7C}$0.92$ &
  \cellcolor[HTML]{FCB87A}$0.72$ &
  \cellcolor[HTML]{64BF7C}$0.91$ &
  \cellcolor[HTML]{FEE883}$0.71$ &
  \cellcolor[HTML]{FFEB84}$0.58$ &
  \cellcolor[HTML]{FBB078}$0.49$ &
  \cellcolor[HTML]{8FCB7E}$-0.09$ \\
InternVL V1.1 &
  \cellcolor[HTML]{7EC67D}$0.63$ &
  \cellcolor[HTML]{E9E583}$0.48$ &
  \cellcolor[HTML]{78C47D}$0.58$ &
  \cellcolor[HTML]{FDD17F}$0.34$ &
  \cellcolor[HTML]{BFD981}$0.61$ &
  \cellcolor[HTML]{FEEA83}$0.56$ &
  \cellcolor[HTML]{B5D680}$0.63$ &
  \cellcolor[HTML]{EFE784}$0.60$ &
  \cellcolor[HTML]{FA9A74}$0.13$ &
  \cellcolor[HTML]{FEEB84}$0.21$ &
  \cellcolor[HTML]{D1DE82}$0.73$ &
  \cellcolor[HTML]{FDD17F}$0.62$ &
  \cellcolor[HTML]{63BE7B}$0.92$ &
  \cellcolor[HTML]{FA9172}$0.66$ &
  \cellcolor[HTML]{65BF7C}$0.91$ &
  \cellcolor[HTML]{FEDC81}$0.68$ &
  \cellcolor[HTML]{AED480}$0.64$ &
  \cellcolor[HTML]{FDC67C}$0.52$ &
  \cellcolor[HTML]{C7DB81}$-0.12$ \\
LLaVA 1.6 34B &
  \cellcolor[HTML]{6DC17C}$0.65$ &
  \cellcolor[HTML]{F3E884}$0.46$ &
  \cellcolor[HTML]{75C47D}$0.58$ &
  \cellcolor[HTML]{FDCA7D}$0.32$ &
  \cellcolor[HTML]{B4D680}$0.62$ &
  \cellcolor[HTML]{E1E383}$0.58$ &
  \cellcolor[HTML]{ADD480}$0.64$ &
  \cellcolor[HTML]{7EC67D}$0.66$ &
  \cellcolor[HTML]{EEE784}$0.26$ &
  \cellcolor[HTML]{FBEA84}$0.22$ &
  \cellcolor[HTML]{70C27C}$0.87$ &
  \cellcolor[HTML]{FEDC81}$0.64$ &
  \cellcolor[HTML]{83C87D}$0.89$ &
  \cellcolor[HTML]{FA9E75}$0.68$ &
  \cellcolor[HTML]{78C47D}$0.88$ &
  \cellcolor[HTML]{FEE282}$0.70$ &
  \cellcolor[HTML]{86C97E}$0.67$ &
  \cellcolor[HTML]{FDCF7E}$0.53$ &
  \cellcolor[HTML]{E6E483}$-0.14$ \\
mBliP mT0 &
  \cellcolor[HTML]{FEE883}$0.44$ &
  \cellcolor[HTML]{FCC47C}$0.40$ &
  \cellcolor[HTML]{BED881}$0.50$ &
  \cellcolor[HTML]{FEE983}$0.42$ &
  \cellcolor[HTML]{D2DE82}$0.59$ &
  \cellcolor[HTML]{F3E884}$0.57$ &
  \cellcolor[HTML]{F1E784}$0.60$ &
  \cellcolor[HTML]{BFD981}$0.63$ &
  \cellcolor[HTML]{FA9673}$0.12$ &
  \cellcolor[HTML]{FDC77D}$0.17$ &
  \cellcolor[HTML]{CBDC81}$0.74$ &
  \cellcolor[HTML]{F2E884}$0.69$ &
  \cellcolor[HTML]{64BF7C}$0.92$ &
  \cellcolor[HTML]{FCBC7B}$0.73$ &
  \cellcolor[HTML]{63BE7B}$0.91$ &
  \cellcolor[HTML]{FEE783}$0.71$ &
  \cellcolor[HTML]{DDE283}$0.60$ &
  \cellcolor[HTML]{FDD27F}$0.54$ &
  \cellcolor[HTML]{63BE7B}$-0.07$ \\
InternVL V1.2+ &
  \cellcolor[HTML]{63BE7B}$0.67$ &
  \cellcolor[HTML]{FEE081}$0.43$ &
  \cellcolor[HTML]{63BE7B}$0.60$ &
  \cellcolor[HTML]{FEEA83}$0.42$ &
  \cellcolor[HTML]{AAD380}$0.63$ &
  \cellcolor[HTML]{E8E583}$0.58$ &
  \cellcolor[HTML]{63BE7B}$0.68$ &
  \cellcolor[HTML]{DCE182}$0.61$ &
  \cellcolor[HTML]{E8E583}$0.28$ &
  \cellcolor[HTML]{F7E984}$0.23$ &
  \cellcolor[HTML]{76C47D}$0.86$ &
  \cellcolor[HTML]{F7E984}$0.68$ &
  \cellcolor[HTML]{64BF7C}$0.92$ &
  \cellcolor[HTML]{FBB078}$0.71$ &
  \cellcolor[HTML]{68C07C}$0.90$ &
  \cellcolor[HTML]{FEE482}$0.70$ &
  \cellcolor[HTML]{71C37C}$0.69$ &
  \cellcolor[HTML]{FDD780}$0.55$ &
  \cellcolor[HTML]{EEE683}$-0.15$ \\
GPT 4V &
  \cellcolor[HTML]{FEEB84}$0.45$ &
  \cellcolor[HTML]{FDCF7E}$0.41$ &
  \cellcolor[HTML]{C9DC81}$0.49$ &
  \cellcolor[HTML]{A6D27F}$0.53$ &
  \cellcolor[HTML]{63BE7B}$0.69$ &
  \cellcolor[HTML]{71C27C}$0.68$ &
  \cellcolor[HTML]{B2D580}$0.64$ &
  \cellcolor[HTML]{7BC57D}$0.66$ &
  \cellcolor[HTML]{63BE7B}$0.70$ &
  \cellcolor[HTML]{BAD881}$0.42$ &
  \cellcolor[HTML]{63BE7B}$0.88$ &
  \cellcolor[HTML]{9ACE7F}$0.81$ &
  \cellcolor[HTML]{79C57D}$0.90$ &
  \cellcolor[HTML]{FBAD78}$0.70$ &
  \cellcolor[HTML]{76C47D}$0.89$ &
  \cellcolor[HTML]{FBEA84}$0.72$ &
  \cellcolor[HTML]{63BE7B}$0.70$ &
  \cellcolor[HTML]{CCDD82}$0.62$ &
  \cellcolor[HTML]{87C97E}$-0.09$ \\ \midrule
Average &
 $0.57$ &
 $0.37$ &
 $0.51$ &
 $0.24$ &
 $0.58$ &
 $0.50$ &
 $0.61$ &
 $0.57$ &
 $0.22$ &
 $0.22$ &
 $0.73$ &
 $0.57$ &
 $0.90$ &
 $0.67$ &
 $0.88$ &
 $0.66$ &
 $0.63$ &
 $0.47$ &
 $-0.15$ \\ \bottomrule
\end{tabular}
\caption{Average performance in English (E) and non-English languages (NE) on all datasets for all models. For each dataset and the $\Delta$ column, the heatmaps are created individually, indicated by the column gutter. The column ``ALL'' represents the average across all datasets. For xFlickrCO and XM3600, we report BertScore F1 and for the rest of the datasets, we report the relaxed accuracy.}
\label{tab:big_res_tab}
\vspace{-.4cm}
\end{table*}

This section discusses the models' performance on the datasets considered in our benchmark.
Table~\ref{tab:big_res_tab} provides an overview of the performance in English compared to non-English languages for all models and datasets.
Note that we use friendly names for the models for better readability (see Table~\ref{tab:models_architecture_overview}).
Detailed results for each dataset and all their respective languages are provided in Appendix~\ref{appendix:result_details}.

\subsection{Summary of Findings}
\label{sec:results_general_summary}
Table~\ref{tab:big_res_tab} shows a clear pattern: Generally, LMMs perform significantly worse in non-English languages across all tasks.
More specifically, the average performance across all models and datasets in English is $0.63$ versus $0.47$ in non-English languages.
Most models have an average performance difference from English to non-English larger or equal to $0.12$.
However, for \texttt{GPT 4V} and despite their much smaller size also for \texttt{mBlip BloomZ}, and \texttt{mBlip T0}, the difference is smaller than $0.1$.
For the two \texttt{mBLIP} models, the authors explicitly stated in their paper the language distribution in the training data, which covers $96$ languages.
Hence, it can be assumed that this is the reason for this slight absolute performance difference, and, further, this might indicate that \texttt{GPT 4V} was also trained in a multilingual fashion.
Due to the difference in size and the architecture\footnote{While the architecture of \texttt{GPT 4V} is not known, it is likely different from the {mBlip} models' architecture, which employs Q-Formers, rarely used in state-of-the-art LMMs.} of the \texttt{mBlip} models and \texttt{GPT 4V}, applying this multilingual training strategy for LMMs would generally lead to more robust multilingual performance.

The average performance difference of the models is most significant on the MaXM, XM3600, and xFlickrCo datasets, for which the models are required to generate non-English text.

Interestingly, for the M5-VLOD dataset, the models that performed worse than the random baseline of $0.2$ in English performed better in non-English languages.
An explanation for this could be false assumptions drawn from the English text.
This finding also explains why the average English versus non-English performance disparity across all models is equal for the dataset and lies around the random baseline, indicating the challenge introduced by our dataset.

\subsubsection{Dataset-Specific Discussion}
\label{sec:results_general_dataset_specific}
%
%In this section, we discuss the dataset-specific results of the different models in more detail.
%
Note that due to brevity constraints, we report exact numbers and diagrams of the language-specific results for each dataset in Appendix~\ref{appendix:result_details}.

\paragraph{xGQA}
\label{sec:results_general_xgqa}
All models perform best in English mostly, with a significant gap in accuracy to the second-best language from up to $0.62$ in English to $0.36$ in Russian for \texttt{LLaVA 1.6 7B}.
In Bengali, where the models have the lowest average accuracy of $0.19$, all models besides \texttt{GPT 4V}, which achieves $0.44$, perform worst by far.
%
%Interestingly, the differences between the languages between the two \texttt{MBlip} models and \texttt{GPT 4V} are comparably minor.
%
%While we do not know the details of the \texttt{GPT 4V} training data, the reason behind this phenomenon, considering the MBlip models, is probably due to the massively multilingual training data.
%
The best-performing model in English and the best-performing model on average over all languages are the \texttt{InternVL v1.2} and \texttt{InternVL v1.1} models.
Notably, despite their (estimated) much larger size, \texttt{GPT 4V} and \texttt{Gemini Pro V} are among the worst-performing models in English.
After manually inspecting the results, we found the reason for this to be that the models did not respond in a single word but with a brief sentence, which is considered a false answer according to the applied metric (see Appendix~\ref{appendix:experimental_setup_details_relaxed_acc_metric} and Section~\ref{sec:limits_metrics_vqa}).
%

%All results for xGQA are reported in Table~\ref{tab:result_details_xgqa_tab}.

\paragraph{MaXM}
\label{sec:results_general_maxm}
The average accuracy of the models for Hindi ($0.22$), Hebrew ($0.19$), Romanian ($0.27$), Thai ($0.25$), and Chinese ($0.24$) is much lower than for English ($0.51$) and French ($0.35$).
%
%Noteworthy, \texttt{MBlip-mT0}, \texttt{GeminiProV}, and \texttt{GPT4 V} perform best in Thai, and the English performance is relatively poor compared to other models.
%
%We find this to be an unexpected result, which might be due to test data contamination.
%
%However, this is only a hypothesis and requires further investigation as done similarly in~\citet{ahuja2023mega} and \citet{ahuja2023megaverse}.
%
It is also worth pointing out that most models, regardless of their size, perform remarkably worse in languages other than English (and French).
In contrast, on xGQA, which is also a VQA dataset, the differences between the languages are much more minor.
This is likely due to the difference between the two datasets, i.e.,  that xGQA has multilingual questions but only English answers, while MaXM has multilingual questions and expects the answers in the respective language, too.
We further underline this in our language fidelity analysis in Section~\ref{sec:results_language_fidelity_analyses}.
%

%All results for MaXM are reported in Table~\ref{tab:result_details_maxm_tab}.

\paragraph{XVNLI}
\label{sec:results_general_xnlvi}
English accuracy is the best for most models, with an average of $0.58$, whereas Arabic accuracy is the worst, with an average of $0.43$.
The performance drop from English to the other languages, i.e., Spanish ($0.51$), French ($0.52$), and Russian, with average accuracy scores of $0.51$, $0.52$, and $0.52$, is less substantial. 
Note that XVNLI is an NLI dataset, i.e., the random baseline is at $\frac{1}{3}$.
All models surpass this baseline in all languages, except for \texttt{CogVLM} in Arabic ($0.26$) and French ($0.27$).
The best-performing model is \texttt{GPT4 V} with an average accuracy across all languages of $0.68$, followed by \texttt{LLaVA 1.6 34B} and \texttt{InternVL V1.2+} with average scores of $0.59$ and $0.58$, respectively.
%

%All results are reported in Table~\ref{tab:result_details_xvnli_tab}.
%

\paragraph{MaRVL}
\label{sec:results_general_marvl}
The dataset's random baseline is $0.5$, which is often only slightly surpassed by most models, especially for Swahili and Tamil languages, with an average accuracy of $0.53$ and $0.54$, respectively.
Notably, only $8$ of $18$ models perform best in English, with an average accuracy of $0.61$.
For the other models, the English performance is surpassed by Chinese, Indonesian, or Turkish, with an average accuracy of $0.60$, $0.60$, and $0.59$, respectively.
\texttt{GPT-4V} is on par with \texttt{LLaVA 1.6 34B} despite the latter having much fewer parameters.

%All results are reported in Table~\ref{tab:result_details_marvl_tab}.
%

\paragraph{M5-VGR}
\label{sec:results_general_m5b_vgr}
As with MaRVL, this dataset's random baseline is at $0.5$.
Only one of $18$ models, i.e., \texttt{InternVL V1.2+}, could surpass or reach this baseline in all languages.
As expected, most models performed best in English, German, or Russian, with average accuracies of $0.73$, $0.68$, and $0.69$, respectively.
They performed worst in low-resource languages such as Amharic, Berber,  Bengali, Hausa, or Zulu, with an average accuracy of $0.53$, $0.49$, $0.55$, and $0.52$, respectively.
Only three models, i.e., \texttt{Gemini Pro V}, \texttt{mBlip mT0}, and \texttt{GPT 4V}, consistently and significantly surpass the random baseline in all languages except for Berber.
The only languages where the average performance is significantly higher than the $0.5$ random baseline are English ($0.73$), German ($0.68$), Russian ($0.69$), and Thai ($0.62$).
The average scores of the other languages range from $0.49$ in Berber to $0.57$ in Hindi.
%

%All results are reported in Table~\ref{tab:result_details_m5b_vgr_tab}.
%

\paragraph{M5-VLOD}
\label{sec:results_general_m5b_vlod}
The dataset's random baseline is $0.2$ since the models need to find the outlier within five images.
Only \texttt{GPT 4V} and \texttt{Gemini Pro V} significantly surpassed that baseline in all languages, with an average accuracy of $0.42$ and $0.36$, respectively.
They achieve the best scores in English with an average accuracy of $0.70$ (\texttt{GPT 4V}) and $0.52$ (\texttt{Gemini Pro V}.
However, in Berber, both models only achieve scores around the random baseline.
All other models do not surpass the random baseline in all languages, including English, by more than $0.1$, with average scores between $0.08$ (\texttt{CogVLM}) and $0.23$ (\texttt{InternVL V1.2+})
This highlights the challenge introduced by our dataset and the performance gap between proprietary and open-source models.
%

%All results are reported in Table~\ref{tab:result_details_m5b_vlod_tab}.
%

\paragraph{xFlickrCO}
\label{sec:results_general_xflickrco}
The majority of models perform best in English, often with a significant margin in average chrF++, i.e., $24.93$ in English and $12.49$ in non-English languages.
Other languages where the modes perform comparably well are German and Spanish, with average chrF++ scores of $19.95$ and $19.55$, respectively.
Interestingly, all models perform worse in non-Latin script languages, i.e., Russian ($9.70$), Chinese ($4.53$), and Japanese ($4.05$).
Unexpectedly, the proprietary models \texttt{GPT 4V} and \texttt{Gemini Pro V} are surpassed by \texttt{mBliP BloomZ}, \texttt{mBliP mT0}, and \texttt{InternVL V1.2+}, which are much smaller open-source models.
Even in English, most open-source models outperform the proprietary models.

%All results are reported in Table~\ref{tab:result_details_xflickrco_tab}.
%

\paragraph{XM3600}
\label{sec:results_general_xm3600}
Note that due to limited resources, we evaluated \texttt{GPT 4V} only on a subset of $12$ of $36$ languages.
Most models perform best in English ($27.14$ average chrF++) by a large margin, followed by other Latin scripts in high-resource languages such as French ($23.65$), Spanish ($23.52$), or Dutch ($21.01$).
On average, the models perform worst on non-Latin script languages like Korean ($3.50$), Telugu ($4.79$), and Bengali ($5.11$).
However, although the chrF++ metric claims to be script and language-independent, the low scores in high-resource languages like Chinese ($3.95$) and Japanese ($5.13$) make the metric questionable.
While detailed analysis is out of the scope of this work, in future work, we will investigate this issue further (see Section~\ref{sec:limits_metrics_ic}).
%

%All results are reported in Table~\ref{tab:result_details_xm3600_tab}.
%
%
\section{Aggregated Result Analyses}
%
%This section presents and discusses analyses performed by aggregating the raw experiment results (see Appendix~\ref{appendix:result_details}) across different dimensions.
%

\subsection{Performance per Language}
\label{sec:results_per_language}
\begin{figure}[ht!]
     \centering
     \begin{subfigure}{1.\linewidth}
        \includegraphics[width=1.\linewidth]{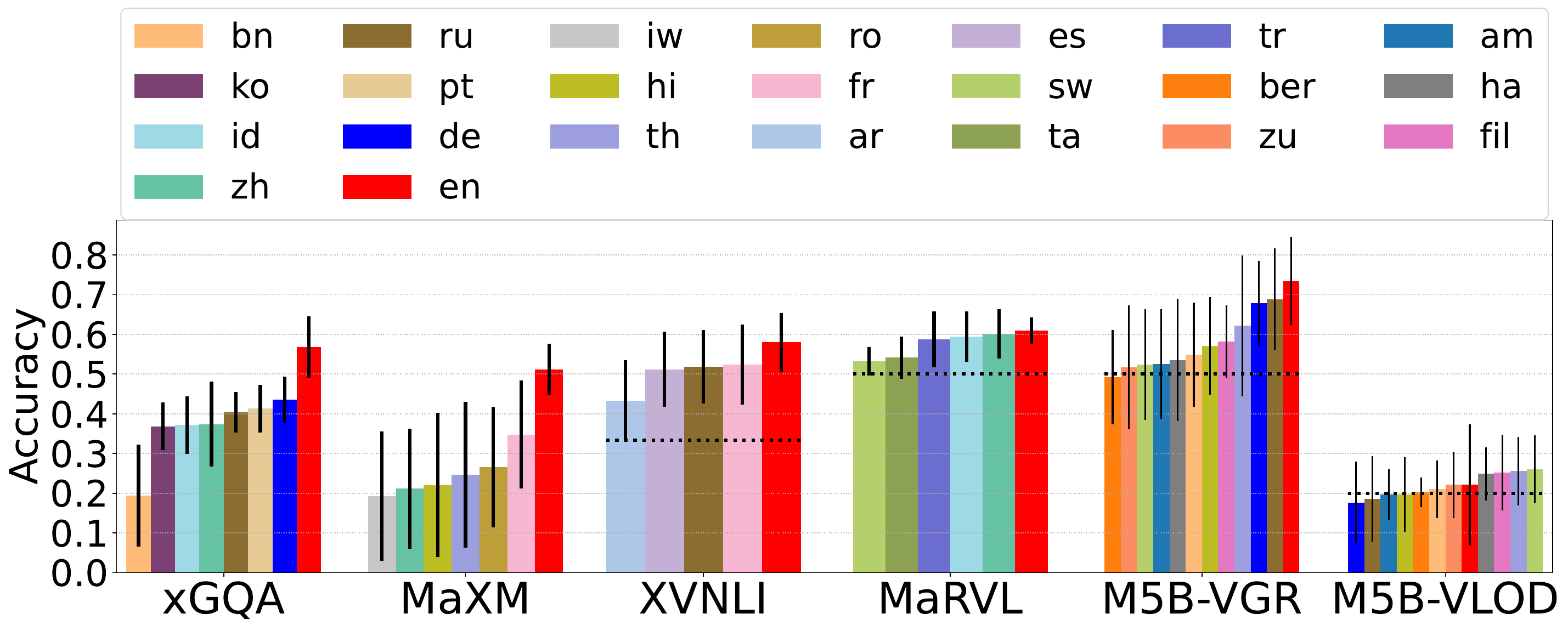}
        \caption{Performance on VQA, VGR, and VNLI datasets aggregated by language.}
        \label{fig:results_per_language_no_ic}
     \end{subfigure}

     \begin{subfigure}{1.\linewidth}
        \includegraphics[width=1.\linewidth]{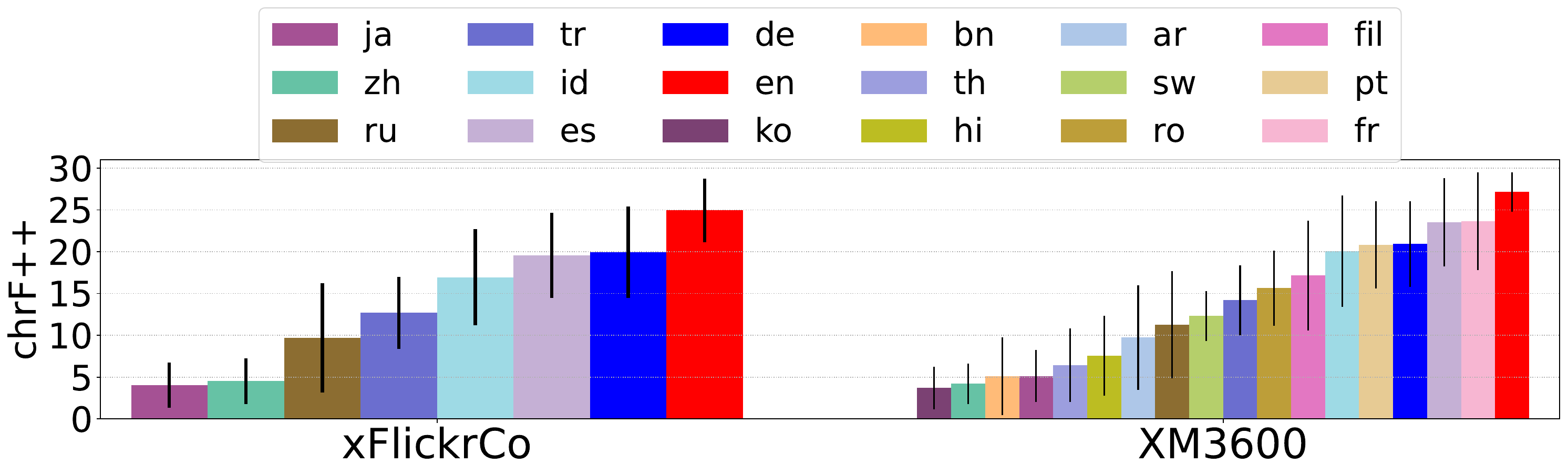}
        \vspace{-.6cm}
        \caption{Performance on image captioning datasets aggregated by language.}
        \label{fig:results_per_language_ic}
     \end{subfigure}

     \begin{subfigure}{1.\linewidth}
        \includegraphics[width=1.\linewidth]{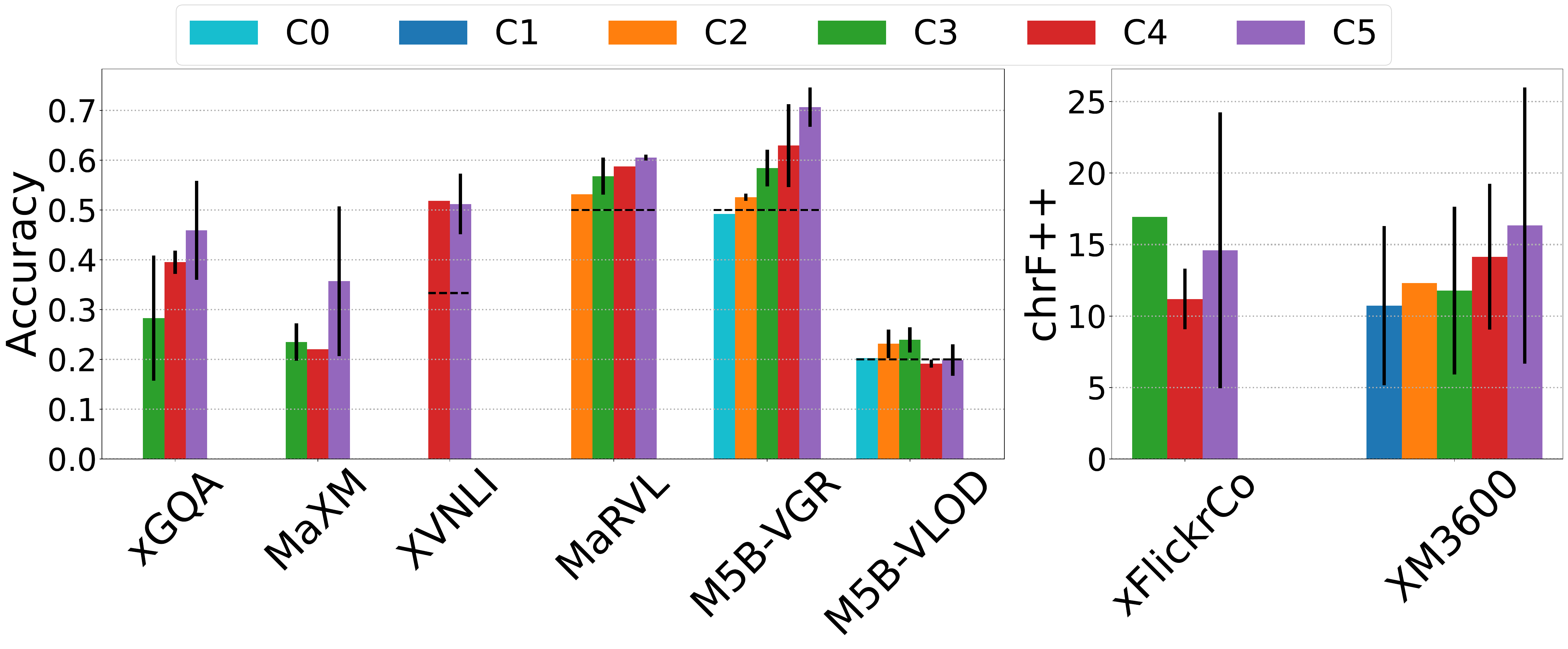}
        \vspace{-.6cm}
        \caption{Performance on datasets aggregated by language taxonomy class as introduced by~\citet{joshi2020langtax}.}
        \label{fig:results_per_tax}
     \end{subfigure}
        \caption{Models' performances on all datasets aggregated by language or language taxonomy classes.}
        %The legends show the ISO 639 language codes.}
        \label{fig:results_per_language}
        \vspace{-0.5cm}
\end{figure}

Figure~\ref{fig:results_per_language} shows the average performances aggregated by language\footnote{We do not show all $36$ languages of XM3600 for better readability.} or language taxonomy classes~\cite{joshi2020langtax}.
These taxonomy classes indicated how well a respective language is represented and considered within the research field of NLP based on papers published at \*CL conferences.
High-resource languages such as English or German are in Class 5, whereas low-resource languages such as Berber are in Class 0.
For details about the languages and their taxonomy classes, please refer to Table~\ref{tab:language_details}.

As can be observed from Figure~\ref{fig:results_per_language_no_ic} and Figure~\ref{fig:results_per_language_ic}, the models perform best in English, followed by other European languages across all datasets.
Our newly presented M5-VLOD dataset is an exception, where the average performance for all languages is around the random baseline, indicating the challenge it implies.
As expected, the models consistently perform worse on low-resource languages than on high-resource languages on all datasets.
This is also displayed in Figure~\ref{fig:results_per_tax}, where it can be observed that the average performance decreases with the language taxonomy class.
Note that this is not precisely true for xFlickrCO and XVNLI because the average on Class-5 languages is lowered by outliers, as indicated by the large error bars.
In contrast, the models performed comparably well in only one Class 3 or 4 language, respectively.

\subsection{Performance vs. Model Parameters}
\label{sec:results_performance_vs_params}
\begin{figure}[ht!]
     \centering
     \begin{subfigure}{1.\linewidth}
        \includegraphics[width=1.\linewidth]{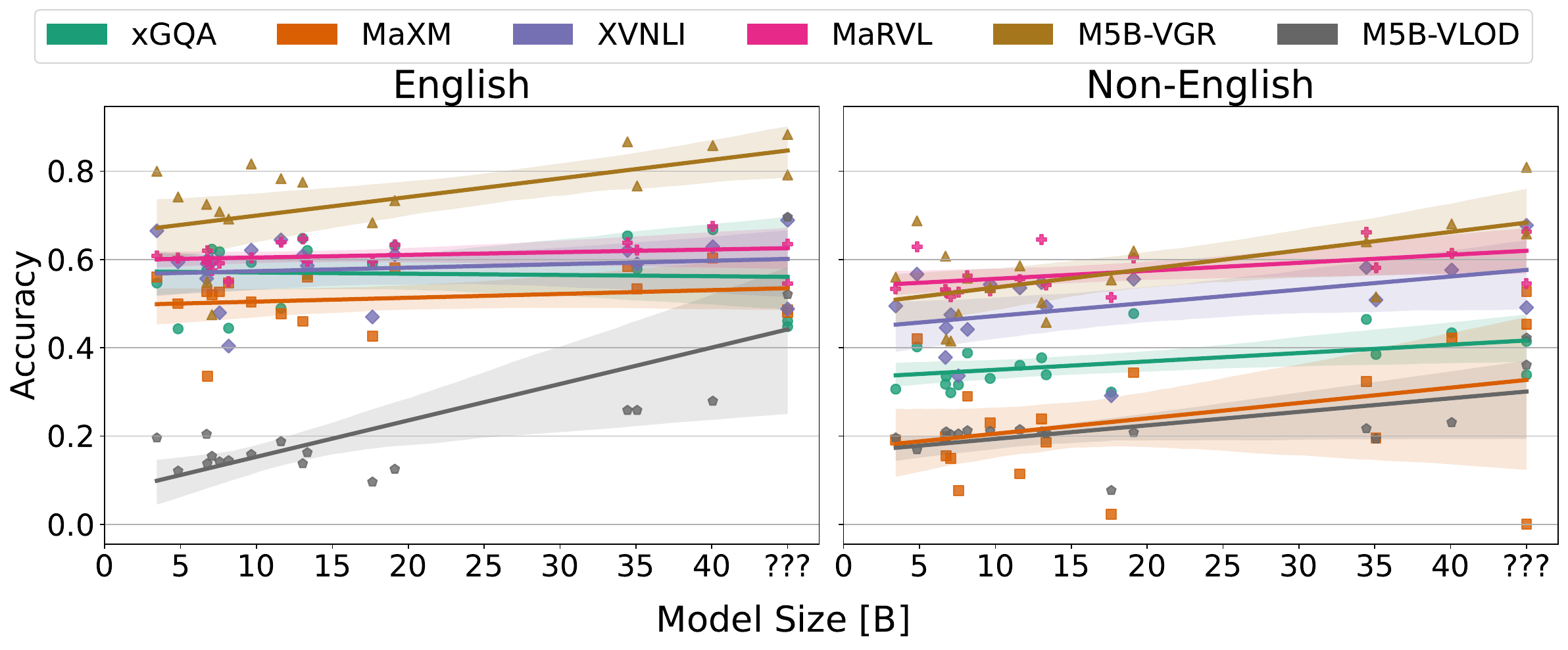}
        % \caption{VQA, VNLI, VGR, and VLOD datasets.}
        % \label{fig:results_performance_vs_params_no_ic}
     \end{subfigure}
     
     \vspace{-.25cm}
     \begin{subfigure}{1.\linewidth}
        \includegraphics[width=1.\linewidth]{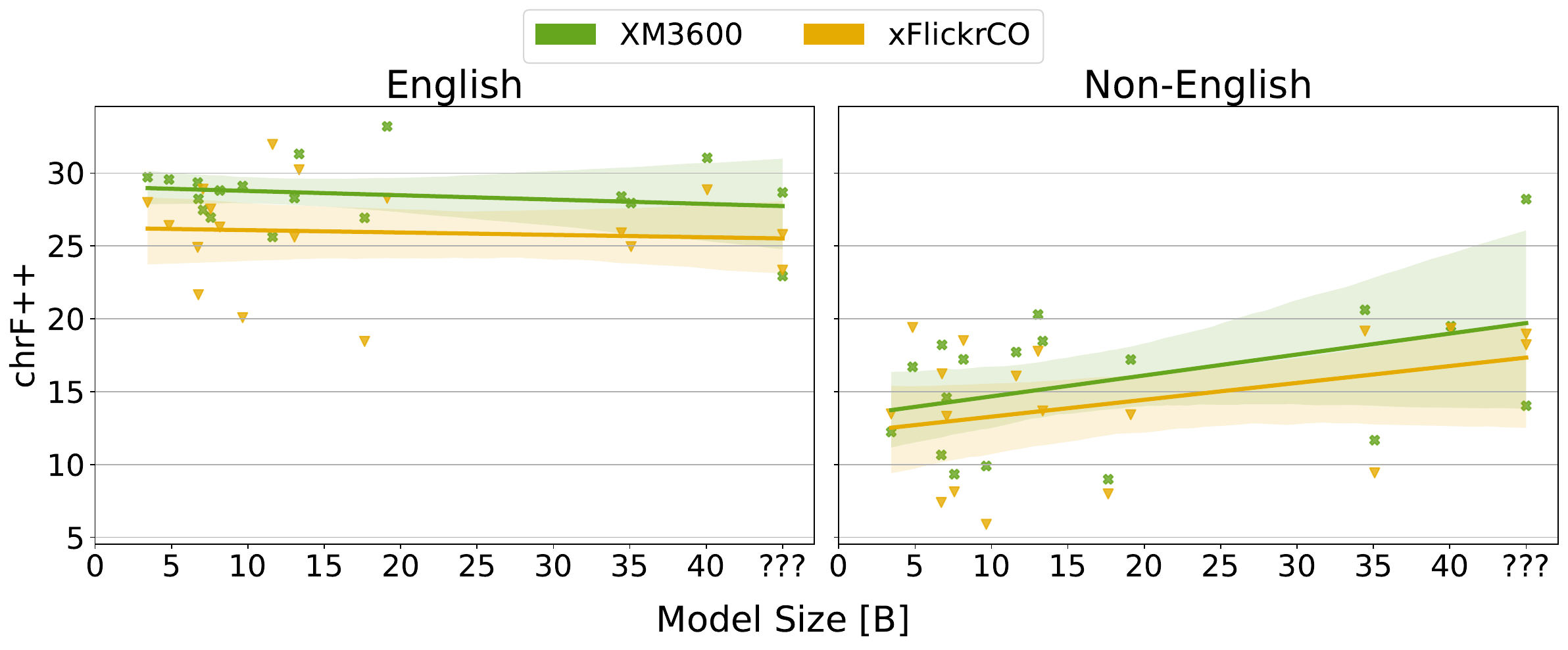}
        % \label{fig:results_performance_vs_params_ic}
        % \caption{Image captioning datasets.}
     \end{subfigure}
    \caption{Regression plots showing the English and average non-English performance versus model size on different datasets. On the x-axis, we indicated the unknown sizes of \texttt{GPT 4V} and \texttt{Gemini Pro V} by ``???''.}
    \label{fig:results_performance_vs_params}
    \vspace{-.6cm}
\end{figure}

In Figure~\ref{fig:results_performance_vs_params}, we plot the English and non-English average performance on the employed datasets versus the models' sizes in multiple regression plots.
Note that, on the x-axes, we indicated the unknown sizes of \texttt{GPT 4V} and \texttt{Gemini Pro V} by ``\textit{???}'', which are estimated to be of magnitudes larger than all other models evaluated in this benchmark hence should be much further right.
However, we did not do so to improve the readability of the plots.

In the figures, we can make several observations:
Firstly, the average English performance is higher than the non-English performance for all models on all datasets.
Secondly, the markers, which represent the average performance of a specific model on a dataset, show that the largest model does not always perform best and that the difference between smaller and larger models is often neglectable.
The same finding is shown by the relatively flat slope of the regression lines.
However, for the M5-VLOD and VGR datasets, the regression line for the average English scores is steeper, meaning that larger models perform considerably better than the smaller models.
Since this work introduces the datasets and M5-VLOD even introduces a novel task, it can be concluded that larger models can better generalize to unseen data.

\subsection{Language Fidelity Analysis}
\label{sec:results_language_fidelity_analyses}
Inspired by~\citet{holtermann2024multiq}, we report the results of a language fidelity analysis, which assesses how often a model responds in the requested language on average.
For this, we used GlotLIDv3~\cite{kargaran2023glotlid} to predict the language based on the output text of the respective models.
Since it is hard to predict the language of a word or a multi-word expression due to ambiguity, we selected the xFlickrCO dataset, where the expected response of a model is an image caption, i.e., a sentence, in one of eight languages.
As it can be observed from Table~\ref{tab:results_language_fidelity_analyses}, all models achieve (almost) perfect fidelity in English where, whereas for Japanese, Russian, and Turkish, the average fidelity drops to two-thirds.
%
%Only seven of $18$ models have an average fidelity above $90\%$ across all languages.
%, and only three have a fidelity above $95\%$ across all languages.
%
Interestingly, the small-sized \texttt{mBLIP} models have almost perfect fidelity in all languages, (slightly) surpassing larger models like \texttt{InternVL V1.2+} and \texttt{GPT 4V}.
\begin{table}[!ht]
\tiny
\centering
\caption{Language fidelity results on the xFlickrCO dataset.}
\label{tab:results_language_fidelity_analyses}
\addtolength{\tabcolsep}{-0.2em}
\begin{tabular}{@{}l|cccccccc|c}
\toprule
\multicolumn{1}{l}{\textbf{Model}} & 
\multicolumn{9}{c}{\textbf{Language}} \\ \midrule
\multicolumn{1}{l}{\textbf{}} &
  \textbf{zh} &
  \textbf{en} &
  \textbf{de} &
  \textbf{id} &
  \textbf{ja} &
  \textbf{ru} &
  \textbf{es} &
  \textbf{tr} &
  \textbf{Avg.} \\ \midrule
BakLLaVA &
  \cellcolor[HTML]{F8696B}$.00$ &
  \cellcolor[HTML]{65BF7C}$1.0$ &
  \cellcolor[HTML]{FAA075}$.39$ &
  \cellcolor[HTML]{F8716C}$.06$ &
  \cellcolor[HTML]{F8696B}$.00$ &
  \cellcolor[HTML]{F8696B}$.00$ &
  \cellcolor[HTML]{FBA777}$.44$ &
  \cellcolor[HTML]{F8696B}$.00$ &
  \cellcolor[HTML]{F98A71}$.24$ \\
Yi-VL 6B &
  \cellcolor[HTML]{F97D6E}$.14$ &
  \cellcolor[HTML]{65BF7C}$1.0$ &
  \cellcolor[HTML]{F98470}$.20$ &
  \cellcolor[HTML]{F8696B}$.00$ &
  \cellcolor[HTML]{F98570}$.20$ &
  \cellcolor[HTML]{F86A6B}$.01$ &
  \cellcolor[HTML]{FCBA7A}$.57$ &
  \cellcolor[HTML]{F8696B}$.00$ &
  \cellcolor[HTML]{FA9072}$.28$ \\
Qwen-VL &
  \cellcolor[HTML]{C6DB81}$.95$ &
  \cellcolor[HTML]{79C57D}$.99$ &
  \cellcolor[HTML]{F9816F}$.18$ &
  \cellcolor[HTML]{F8786D}$.11$ &
  \cellcolor[HTML]{F97E6F}$.15$ &
  \cellcolor[HTML]{F8746D}$.08$ &
  \cellcolor[HTML]{F97E6F}$.15$ &
  \cellcolor[HTML]{F8726C}$.07$ &
  \cellcolor[HTML]{FA9874}$.33$ \\
Yi-VL 34B &
  \cellcolor[HTML]{FBA676}$.43$ &
  \cellcolor[HTML]{64BF7C}$1.0$ &
  \cellcolor[HTML]{FED980}$.79$ &
  \cellcolor[HTML]{FBA977}$.45$ &
  \cellcolor[HTML]{FCBB7A}$.58$ &
  \cellcolor[HTML]{F98770}$.22$ &
  \cellcolor[HTML]{F98C71}$.25$ &
  \cellcolor[HTML]{FA9773}$.33$ &
  \cellcolor[HTML]{FBB078}$.51$ \\
CogVLM &
  \cellcolor[HTML]{FBA776}$.44$ &
  \cellcolor[HTML]{BBD881}$.95$ &
  \cellcolor[HTML]{FDD17F}$.74$ &
  \cellcolor[HTML]{FDD57F}$.76$ &
  \cellcolor[HTML]{FA9F75}$.38$ &
  \cellcolor[HTML]{FBA576}$.43$ &
  \cellcolor[HTML]{FEDD81}$.82$ &
  \cellcolor[HTML]{FCB579}$.54$ &
  \cellcolor[HTML]{FCC27C}$.63$ \\
LLaVA 1.5 13B &
  \cellcolor[HTML]{FEE583}$.88$ &
  \cellcolor[HTML]{64BF7C}$1.0$ &
  \cellcolor[HTML]{FDD37F}$.75$ &
  \cellcolor[HTML]{FCB679}$.55$ &
  \cellcolor[HTML]{FEE983}$.90$ &
  \cellcolor[HTML]{F98D72}$.26$ &
  \cellcolor[HTML]{FDD47F}$.75$ &
  \cellcolor[HTML]{FBA175}$.40$ &
  \cellcolor[HTML]{FDCA7D}$.69$ \\
LLaVA 1.5 7B &
  \cellcolor[HTML]{FEDF81}$.83$ &
  \cellcolor[HTML]{65BF7C}$1.0$ &
  \cellcolor[HTML]{A6D27F}$.96$ &
  \cellcolor[HTML]{FEDE81}$.83$ &
  \cellcolor[HTML]{F8756D}$.09$ &
  \cellcolor[HTML]{F98770}$.22$ &
  \cellcolor[HTML]{97CD7E}$.97$ &
  \cellcolor[HTML]{FDC87D}$.67$ &
  \cellcolor[HTML]{FDCC7E}$.70$ \\
MiniCPM-V &
  \cellcolor[HTML]{F98670}$.21$ &
  \cellcolor[HTML]{65BF7C}$1.0$ &
  \cellcolor[HTML]{E2E383}$.93$ &
  \cellcolor[HTML]{FED980}$.79$ &
  \cellcolor[HTML]{FEE783}$.89$ &
  \cellcolor[HTML]{ACD480}$.96$ &
  \cellcolor[HTML]{FFEB84}$.91$ &
  \cellcolor[HTML]{FDCA7D}$.68$ &
  \cellcolor[HTML]{FEDA80}$.80$ \\
LLaVA 1.6 7B &
  \cellcolor[HTML]{79C57D}$.99$ &
  \cellcolor[HTML]{72C37C}$.99$ &
  \cellcolor[HTML]{FDC77D}$.66$ &
  \cellcolor[HTML]{FEEA83}$.91$ &
  \cellcolor[HTML]{FCBC7B}$.59$ &
  \cellcolor[HTML]{FEE683}$.88$ &
  \cellcolor[HTML]{FEEA83}$.91$ &
  \cellcolor[HTML]{FEE783}$.89$ &
  \cellcolor[HTML]{FEE282}$.85$ \\
InternVL V1.1 &
  \cellcolor[HTML]{A6D27F}$.96$ &
  \cellcolor[HTML]{65BF7C}$1.0$ &
  \cellcolor[HTML]{E6E483}$.93$ &
  \cellcolor[HTML]{FDD880}$.78$ &
  \cellcolor[HTML]{FEE683}$.88$ &
  \cellcolor[HTML]{FEE883}$.89$ &
  \cellcolor[HTML]{93CC7E}$.97$ &
  \cellcolor[HTML]{FDC77D}$.66$ &
  \cellcolor[HTML]{FEE683}$.89$ \\
OmniLMM 12B &
  \cellcolor[HTML]{FCC37C}$.63$ &
  \cellcolor[HTML]{63BE7B}$1.0$ &
  \cellcolor[HTML]{C0D981}$.95$ &
  \cellcolor[HTML]{F2E884}$.92$ &
  \cellcolor[HTML]{FEDE81}$.83$ &
  \cellcolor[HTML]{FBEA84}$.92$ &
  \cellcolor[HTML]{88C97E}$.98$ &
  \cellcolor[HTML]{FEE683}$.88$ &
  \cellcolor[HTML]{FEE783}$.89$ \\
Gemini Pro &
  \cellcolor[HTML]{C2DA81}$.95$ &
  \cellcolor[HTML]{BBD881}$.95$ &
  \cellcolor[HTML]{BCD881}$.95$ &
  \cellcolor[HTML]{FEE683}$.88$ &
  \cellcolor[HTML]{FEEB84}$.91$ &
  \cellcolor[HTML]{A6D27F}$.96$ &
  \cellcolor[HTML]{A1D07F}$.97$ &
  \cellcolor[HTML]{A9D27F}$.96$ &
  \cellcolor[HTML]{CCDD82}$.94$ \\
LLaVA 1.6 13B &
  \cellcolor[HTML]{68C07C}$1.0$ &
  \cellcolor[HTML]{67C07C}$1.0$ &
  \cellcolor[HTML]{FEE983}$.90$ &
  \cellcolor[HTML]{ABD380}$.96$ &
  \cellcolor[HTML]{FEE983}$.91$ &
  \cellcolor[HTML]{FEE582}$.87$ &
  \cellcolor[HTML]{9ACE7F}$.97$ &
  \cellcolor[HTML]{DBE182}$.93$ &
  \cellcolor[HTML]{CBDC81}$.94$ \\
LLaVA 1.6 34B &
  \cellcolor[HTML]{FEE683}$.88$ &
  \cellcolor[HTML]{64BF7C}$1.0$ &
  \cellcolor[HTML]{7EC67D}$.99$ &
  \cellcolor[HTML]{7EC67D}$.99$ &
  \cellcolor[HTML]{FEE282}$.86$ &
  \cellcolor[HTML]{6EC27C}$.99$ &
  \cellcolor[HTML]{75C47D}$.99$ &
  \cellcolor[HTML]{75C37C}$.99$ &
  \cellcolor[HTML]{A9D380}$.96$ \\
GPT 4V &
  \cellcolor[HTML]{99CE7F}$.97$ &
  \cellcolor[HTML]{63BE7B}$1.0$ &
  \cellcolor[HTML]{6BC17C}$1.0$ &
  \cellcolor[HTML]{8BCA7E}$.98$ &
  \cellcolor[HTML]{FEE582}$.88$ &
  \cellcolor[HTML]{6EC27C}$.99$ &
  \cellcolor[HTML]{72C37C}$.99$ &
  \cellcolor[HTML]{63BE7B}$1.0$ &
  \cellcolor[HTML]{8FCB7E}$.98$ \\
InternVL V1.2+ &
  \cellcolor[HTML]{74C37C}$.99$ &
  \cellcolor[HTML]{65BF7C}$1.0$ &
  \cellcolor[HTML]{6AC07C}$1.0$ &
  \cellcolor[HTML]{C5DB81}$.95$ &
  \cellcolor[HTML]{9ACE7F}$.97$ &
  \cellcolor[HTML]{79C57D}$.99$ &
  \cellcolor[HTML]{76C47D}$.99$ &
  \cellcolor[HTML]{AED480}$.96$ &
  \cellcolor[HTML]{87C97E}$.98$ \\
mBliP BloomZ &
  \cellcolor[HTML]{B1D580}$.96$ &
  \cellcolor[HTML]{63BE7B}$1.0$ &
  \cellcolor[HTML]{63BE7B}$1.0$ &
  \cellcolor[HTML]{70C27C}$.99$ &
  \cellcolor[HTML]{70C27C}$.99$ &
  \cellcolor[HTML]{68C07C}$1.0$ &
  \cellcolor[HTML]{66BF7C}$1.0$ &
  \cellcolor[HTML]{6DC17C}$.99$ &
  \cellcolor[HTML]{72C37C}$.99$ \\
mBliP mT0 &
  \cellcolor[HTML]{ACD480}$.96$ &
  \cellcolor[HTML]{63BE7B}$1.0$ &
  \cellcolor[HTML]{63BE7B}$1.0$ &
  \cellcolor[HTML]{6EC27C}$.99$ &
  \cellcolor[HTML]{7EC67D}$.99$ &
  \cellcolor[HTML]{64BF7C}$1.0$ &
  \cellcolor[HTML]{64BF7C}$1.0$ &
  \cellcolor[HTML]{69C07C}$1.0$ &
  \cellcolor[HTML]{72C37C}$.99$ \\ \hline
Avg. &
  \cellcolor[HTML]{FFEB84}$.73$ &
  \cellcolor[HTML]{63BE7B}$.99$ &
  \cellcolor[HTML]{DAE182}$.79$ &
  \cellcolor[HTML]{FDD37F}$.72$ &
  \cellcolor[HTML]{F98370}$.67$ &
  \cellcolor[HTML]{F8696B}$.65$ &
  \cellcolor[HTML]{CFDD82}$.81$ &
  \cellcolor[HTML]{F9826F}$.66$ &
  \cellcolor[HTML]{F2E884}$.75$ \\ \bottomrule
\end{tabular}
\end{table}

While the language fidelity of a model focuses on the generated text, we argue that the fidelity is also an indicator of the model's general language capabilities.
To prove this hypothesis, we computed Pearson correlation coefficients between the reported fidelity and the models' performance on the datasets for the xFlickrCO languages.
As shown in Table~\ref{tab:results_language_fidelity_analyses_correlations}, there is a positive moderate or high correlation between the average fidelity and the average score for most datasets.
However, for xGQA and M5-VLOD, there is only a minor positive average correlation.%, which would require deeper analysis to explain and is outside the scope of this work.
\section{Conclusion}
\label{sec:conclusion}
%
%\footnote{We will release all code and data upon acceptance.}
We introduced M5, a diverse benchmark in which we evaluated $18$ Large Multimodal Models (LMMs) with varying sizes across five visio-linguistic tasks in eight datasets comprising $41$ unique languages.
Further, we presented two novel datasets -- M5-VGR and M5-VLOD -- which focus on underrepresented languages and depict culturally diverse scenes.
With M5-VLOD, we introduce a new visio-linguistic outlier detection task in which only proprietary models achieve reasonable scores.
Our experiments revealed that model size does not always correlate with better performance, especially in non-English languages, underscoring the importance of diverse, multilingual training data and robust architectures.
Performance disparities were prominent between high-resource languages like English and low-resource languages across all datasets and models, highlighting ongoing challenges in achieving globally equitable multilingual AI.
With M5, we aim to impel the development of more inclusive models suitable for diverse languages and cultures.
%
%Future work will expand on tasks, datasets, and the new generation of LMMs.
%
%
\section{Limitations}
\label{sec:limits}

This section outlines several limitations of our current study that will be addressed in future work.
\subsection{Metrics for Multilingual Image Captioning}
\label{sec:limits_metrics_ic}
Our benchmark and current research generally lack robust metrics for evaluating multilingual image captioning, especially for non-Latin script languages.
The issue, which is the same for machine translation tasks, arises because of the nature of most metrics, such as chrF~\cite{popovic2017chrf}, CIDEr~\cite{vedantam2015cider}, ROUGE~\cite{lin2004rouge}, BLUE~\cite{papineni2002bleu}, or METEOR~\cite{banerjee2005meteor}, which are based on comparing word or character n-grams between the source and target sequence.
For non-Latin scripts, tokenization or segmentation can be challenging because it might not contain spaces or punctuation, or the characters are logographic.
Hence, their usability or effectiveness is doubtful in such scenarios because the metrics rely on tokenization.

Other metrics, such as BERTScore~\cite{zhang2020bertscore}, CLIPScore~\cite{hessel2021clipscore}, or COMET~\cite{rei2020comet}, do not rely on the captions' surface forms but on their token or sentence embeddings.
However, they suffer from other issues: They require strong multilingual or cross-lingual encoder models capable of computing embeddings for many languages, which itself is a challenging task.
Further, the scores computed with these metrics are often not calibrated across languages and thus not directly comparable between different languages.

A promising currently popular solution might be the use of robust multilingual state-of-the-art LLMs such as GPT 4o\footnote{\url{https://openai.com/index/hello-gpt-4o/}}, Claude 3 Opus\footnote{\url{https://www.anthropic.com/news/claude-3-family}}, or Gemini 1.5 Ultra\footnote{\url{https://blog.google/technology/ai/google-gemini-next-generation-model-february-2024/}} as a judge~\cite{zheng2024llmjudge}.
However, this would require more computational and financial resources and, most importantly, more investigation.

\subsection{VQA Metrics for Generative Models}
\label{sec:limits_metrics_vqa}
The problem when employing and evaluating generative language models on question-answering tasks is that the models can generally output arbitrary token sequences.
However, the gold label answers are limited and often comprise only a short phrase, a single word, or even a binary label.
Hence, mapping the predicted answers to their gold labels is not straightforward, and the difficulty drastically increases in multilingual scenarios.
The relaxed accuracy metric employed in this study (see Section~\ref{appendix:experimental_setup_metrics}) has been found to occasionally incorrectly classify correct answers, leading to false negatives, especially in open vocabulary visual question answering (VQA).
One way to address this issue is to leverage strong state-of-the-art LLMs as judges, as described above, to enhance the accuracy of the evaluations.

\subsection{Influence of Prompting}
\label{sec:limits_prompts}
Another limitation of this and most, if not all, other current studies is grounded in the model prompting.
Since different models might react differently to specific prompting styles, and we only employ a single prompt per dataset for all models\footnote{We do apply the model-specific prompt or chat templates, though.} (see Figure~\ref{fig:setup_prompts}), the results might not be optimal.
This issue has been partially addressed by~\citet{ahuja2023mega} but is out of the scope of this work.

\subsection{``Outdated'' Models}
\label{sec:limits_models}
Since the pace of current research in NLP, CV, and multimodal machine learning is swift, the models employed in our benchmarking exercise might be considered slightly outdated.
Note that we considered models released until March 2024.
Since then, numerous improved LMMs based on state-of-the-art LLMs, such as Llama3\footnote{\url{https://ai.meta.com/blog/meta-llama-3}} and novel image encoders techniques such as NaVIT~\cite{dehghani2024navit}, have been publicly released.
Because this was foreseeable, we designed our benchmark to be easily extendable with newer models, which we will include in future work.

\subsection{Small M5 Datasets}
\label{sec:limits_small_m5b_ds}
This work introduced two datasets, M5-VGR and M5-VLOD, which comprise about $115$ samples for each of the $12$ languages.
Compared to other datasets, they can be considered small.
We will increase their sizes in future work to obtain more robust and generalizable results.

\subsection{Missing multimodal and Multilingual Datasets}
\label{sec:limits_datasets}
Currently, the M5 Benchmark comprises $5$ text-image tasks, i.e., VQA, VGR, VNLI, and image captioning, thus missing other suitable tasks like multimodal and multilingual summarization.
Further, other multimodal multilingual VQA and VGR datasets have emerged while writing this paper.
We will include both new tasks and new datasets in future versions of the M5.
\bibliography{custom}
\appendix
\onecolumn
\section{Experimental Setup Details}
\label{appendix:experimental_setup_details}
This section details the employed metrics, prompts, and generation hyperparameters.

Note that we ran all experiments on A6000 ($50$GB) and A100 ($80$GB) GPUs. The largest evaluated model ($40$B) fits on an A100.

\subsection{Metrics}
\label{appendix:experimental_setup_metrics}
Following~\citet{geigle2023mblip}, we report a relaxed accuracy metric for the xGQA, MaXM, XVNLI, and MaRVL datasets due to the generative nature of the considered models.
More specifically, we post-process the generated answers by, e.g., lowercasing, stripping, or removing punctuation.
We then consider the processed generated answer correct if it matches the gold answer or starts or ends with the gold answer.
Further, we allow synonyms for boolean and numerical values.
Examples can be found in Table~\ref{appendix:experimental_setup_details_relaxed_acc_metric}.

Inspired by~\citet{ahuja2023megaverse}, we report the chrF++~\cite{popovic2017chrf} metric for the xFlickrCo and XM3600 datasets.
\subsection{Relaxed Accuracy Metric}
\label{appendix:experimental_setup_details_relaxed_acc_metric}
\begin{table}[!ht]
\centering
\caption{Examples of generated answers considered correct or incorrect in the relaxed accuracy metric used to measure the performance on the xGQA, MaXM, MaRVL, XVNLI, M5-VGR, and M5-VLOD datasets. For more details, please refer to our GitHub repository.}
\begin{tabular}{llc}
\toprule
\textbf{Generated Answer}           & \textbf{Gold Answer} & \textbf{Considered Correct}  \\
\midrule
\{Yes, 1, True\}                    & true                 & \yes                         \\
\{No, 0, False\}                    & false                & \yes                         \\
A car.                              & car                  & \yes                         \\
Yes, it is correct.                 & yes                  & \yes                         \\
It is not correct, no.              & no                   & \yes                         \\
The color of the leaf is green.     & green                & \yes                         \\
There are three birds.              & three birds          & \yes                         \\
Five                                & 5                    & \yes                         \\
\{yes, true\}                       & entailment           & \yes                       \\
\{no, false\}                       & contradiction        & \yes                       \\
maybe                               & neutral              & \yes                       \\
There are three birds in the image. & three birds          & \no                        \\
There are three birds.              & 3                    & \no                        \\
three birds                         & 3                    & \no                        \\
three birds                         & 3 birds              & \no                        \\
\bottomrule
\end{tabular}
\end{table}
\subsection{Prompts}
\label{appendix:experimental_setup_details_prompts}
\begin{figure}[ht!]\tiny
    \begin{promptbox}{xGQA}
    Question: \{QUESTION\} Short answer in English:
    \end{promptbox}
    \begin{promptbox}{MaXM}
    Question: \{QUESTION\} Short answer in \{LANGUAGE\}:
    \end{promptbox}
    \begin{promptbox}{MaRVL}
    Based on the two images, is it correct to say ``\{HYPOTHESIS\}''? Yes or no? One word answer in English:
    \end{promptbox}
    \begin{promptbox}{XVNLI}
    Is it guaranteed true that ``\{HYPOTHESIS\}''? Yes, no, or maybe? One word answer in English:
    \end{promptbox}
    \begin{promptbox}{M5-VGR}
    Based on the two images, is it correct to say ``\{HYPOTHESIS\}''? Yes or no? One word answer in English:
    \end{promptbox}
    \begin{promptbox}{M5-VLOD}
    Based on the 5 images ordered from top-left to bottom-right, which image does not match the hypothesis ``\{HYPOTHESIS\}''? Choose one from [A, B, C, D, E] and only output a single letter:
    \end{promptbox}
    \begin{promptbox}{xFlickrCo}
    Brief caption in \{LANGUAGE\}:
    \end{promptbox}
    \begin{promptbox}{XM3600}
    Brief caption in \{LANGUAGE\}:
    \end{promptbox}
    \caption{Prompts employed for the different datasets.}
    \label{fig:setup_prompts}
\end{figure}
Figure~\ref{fig:setup_prompts} presents the dataset-specific textual prompts we used for all models in this benchmark.
Note that this does not include model-specific prompt templates, image placeholders, special tags, or symbols, only the ''raw'' textual prompt, which is then embedded in the template as required by the respective model.
The placeholders \texttt{\{QUESTION\}}, \texttt{\{LANGUAGE\}}, or \texttt{\{HYPOTHESIS\}} are replaced by the sample specific text.
The prompts are partially inspired by \citet{geigle2023mblip} or \citet{bugliarello22iglue}.
\subsection{Hyperparameters}
\label{appendix:experimental_setup_details_hyperparameters}
This section briefly reports hyperparameters used within our experiments for better reproducibility.
\subsubsection{Generation Parameters}
We used the same generation hyperparameters to generate responses with all the employed open-source models on all datasets (see Table~\ref{tab:generation_kwargs}).
Those are inspired by the default parameters in the ``transformers'' library\footnote{\url{https://huggingface.co/docs/transformers/en/main_classes/text_generation}}.
Because for CogVLM, beam search is not supported, we set ``num\_beams'' to $1$.
For GPT 4V and Gemini Pro V, we use the default parameters of the respective Python clients.
\begin{table}[!ht]
\centering
\caption{Generation hyperparameters to generate responses with all the employed models on all datasets.}
\label{tab:generation_kwargs}
\begin{tabular}{ll}
\toprule
Parameter        & Value  \\
\midrule
num\_beams       & $2$    \\
do\_sample       & True   \\
max\_new\_tokens & $50$   \\
temperature      & $1.0$  \\
top\_k           & $50$   \\
top\_p           & $0.95$ \\
\bottomrule
\end{tabular}
\end{table}

\subsubsection{Image Order for Multi-Image Datasets}
Most models employed in our dataset only support a single image per prompt.
For datasets where a sample comprises more than one image, i.e., for MaRVL, M5-VGR, and M5-VLOD, we use the following strategy:
We first stack the images horizontally with a gutter of $10$ pixels, provide them as a single image in the prompt, and generate the response.
Then, we do the same again but stack the images vertically.
For M5-VLOD, we also create a stacked image with two columns and three rows.
The reported scores are the average of all variants.
\newpage
\onecolumn
\section{Dataset Details}
\label{appendix:dataset_details}

\subsection{M5-VGR and M5-VLOD Details}
\label{appendix:dataset_details_m5b}
\subsubsection{M5-VGR Examples}
\label{appendix:dataset_details_m5b_examples_vgr}
\begin{figure*}[!htb]
    \centering
    \includegraphics[width=1.\linewidth]{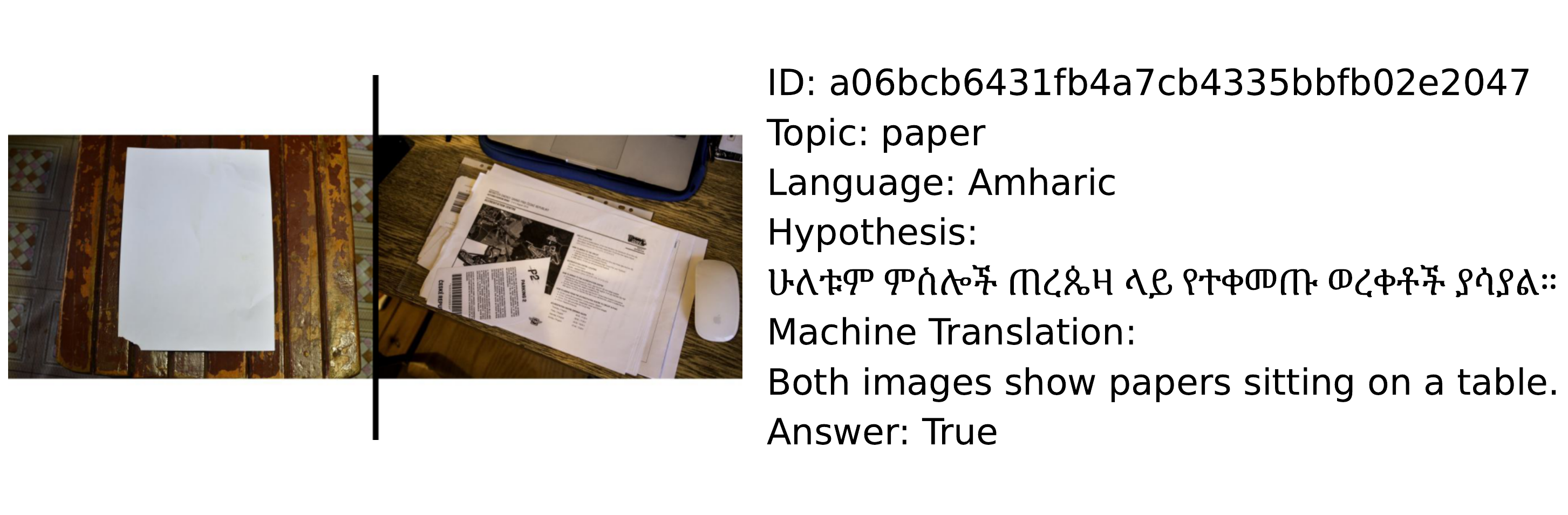}
    \caption{Amharic M5-VGR Sample.}
\end{figure*}
\begin{figure*}[!htb]
    \centering
    \includegraphics[width=1.\linewidth]{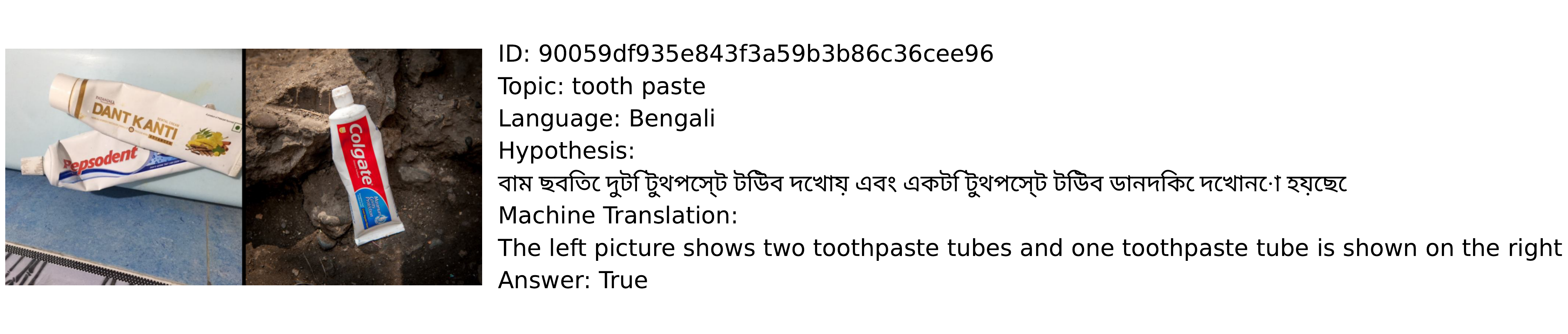}
    \caption{Bengali M5-VGR Sample.}
\end{figure*}

\begin{figure*}[!htb]
    \centering
    \includegraphics[width=1.\linewidth]{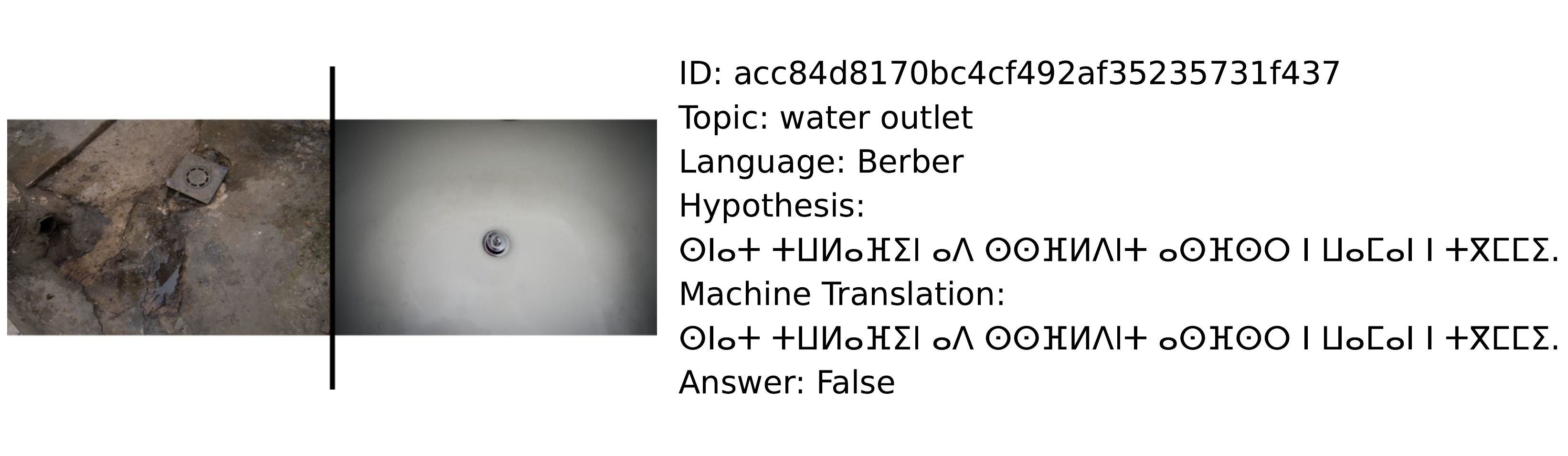}
    \caption{Berber M5-VGR Sample.}
\end{figure*}

\begin{figure*}[!htb]
    \centering
    \includegraphics[width=1.\linewidth]{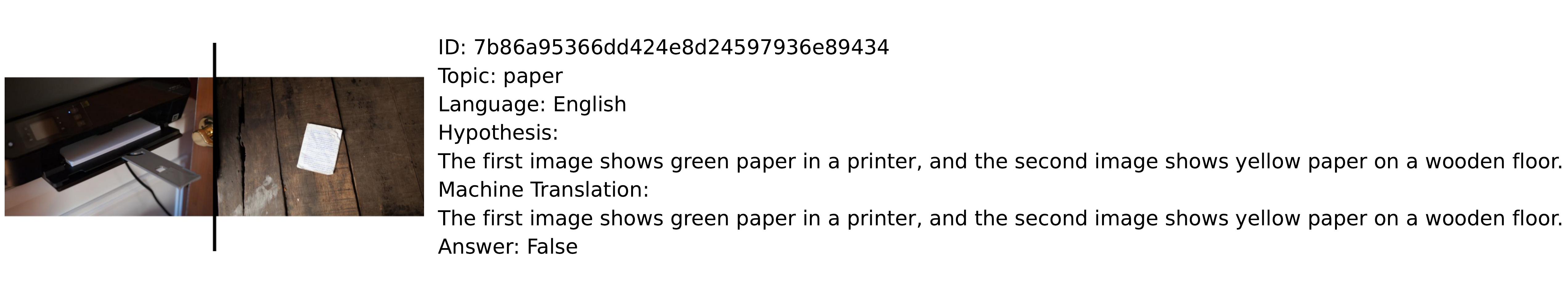}
    \caption{English M5-VGR Sample.}
\end{figure*}

\begin{figure*}[!htb]
    \centering
    \includegraphics[width=1.\linewidth]{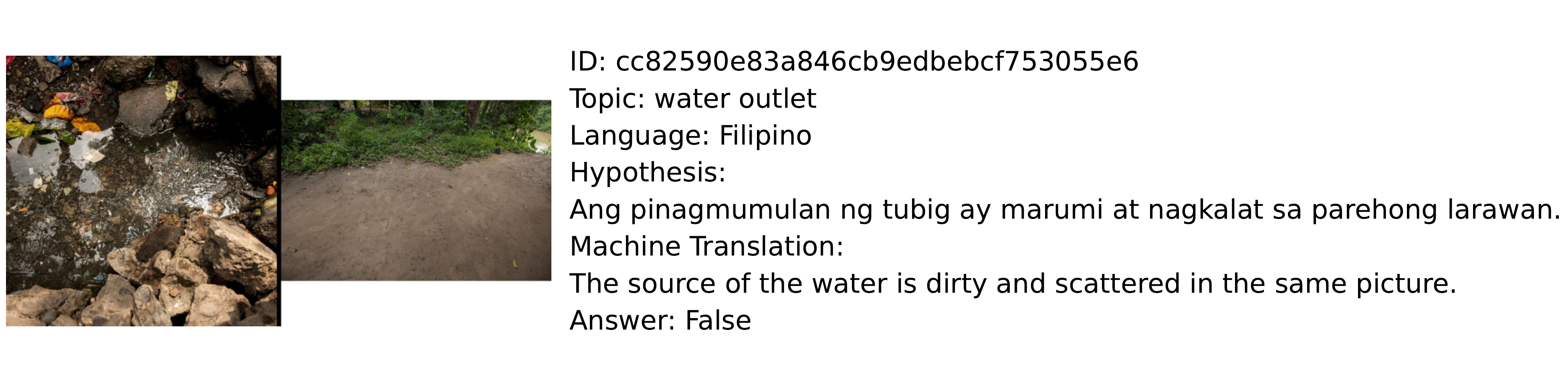}
    \caption{Filipino M5-VGR Sample.}
\end{figure*}

\begin{figure*}[!htb]
    \centering
    \includegraphics[width=1.\linewidth]{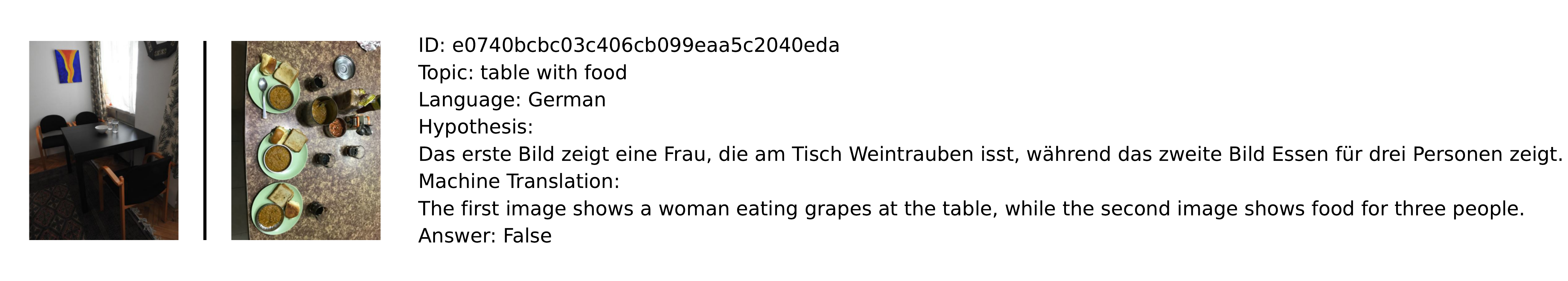}
    \caption{German M5-VGR Sample.}
\end{figure*}

\begin{figure*}[!htb]
    \centering
    \includegraphics[width=1.\linewidth]{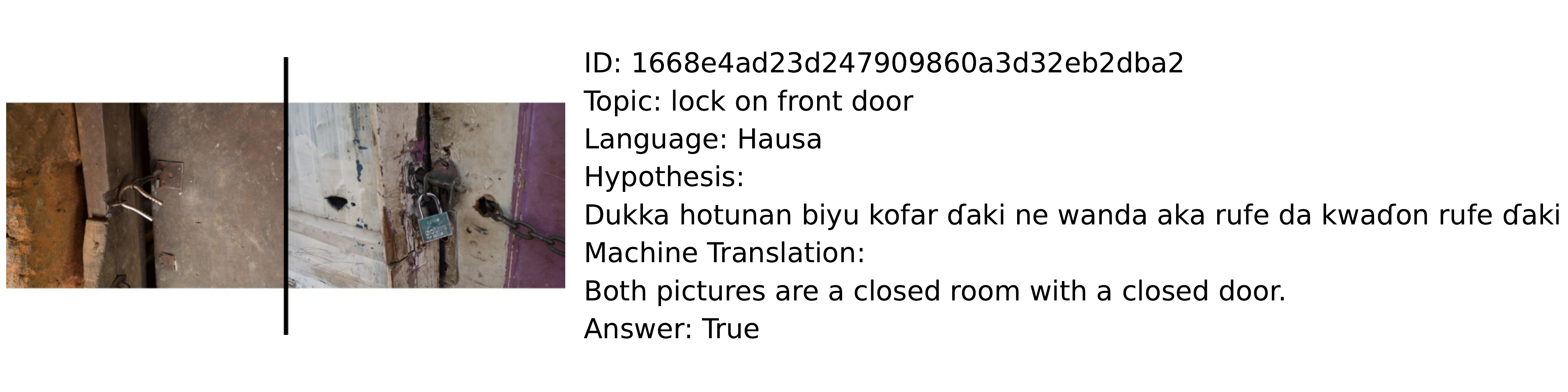}
    \caption{Hausa M5-VGR Sample.}
\end{figure*}

\begin{figure*}[!htb]
    \centering
    \includegraphics[width=1.\linewidth]{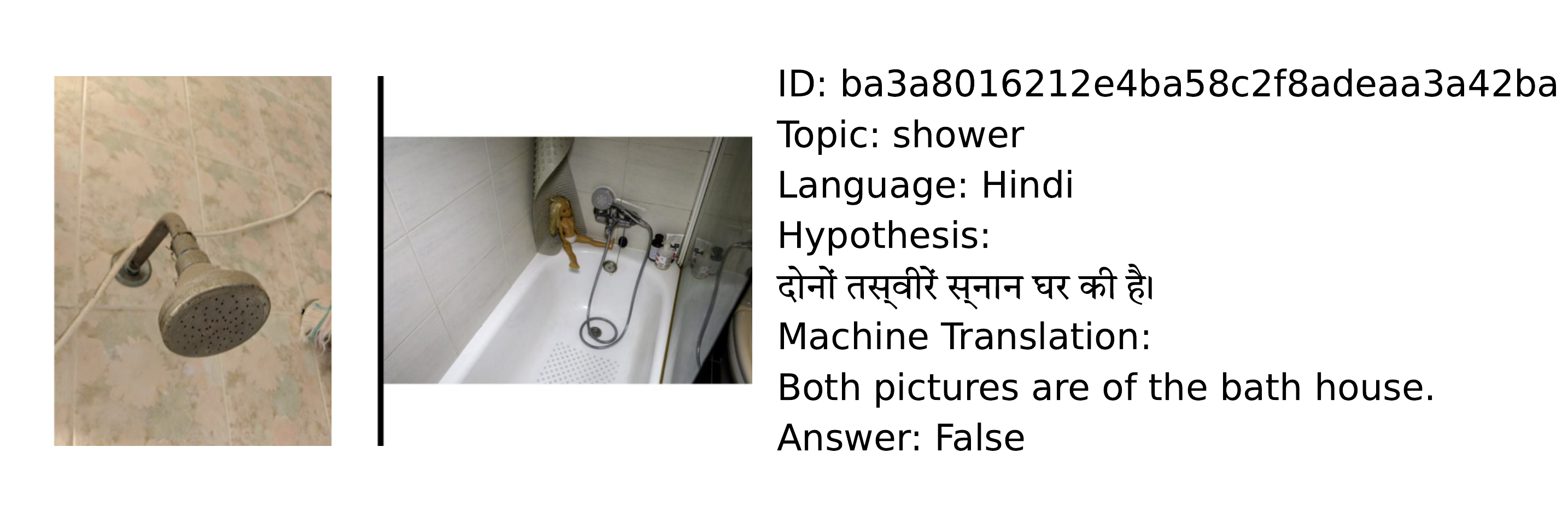}
    \caption{Hindi M5-VGR Sample.}
\end{figure*}

\begin{figure*}[!htb]
    \centering
    \includegraphics[width=1.\linewidth]{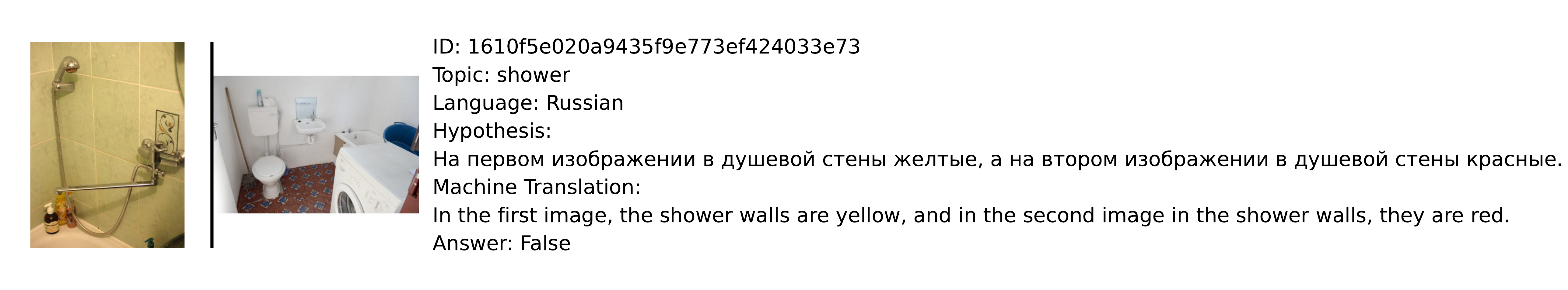}
    \caption{Russian M5-VGR Sample.}
\end{figure*}

\begin{figure*}[!htb]
    \centering
    \includegraphics[width=1.\linewidth]{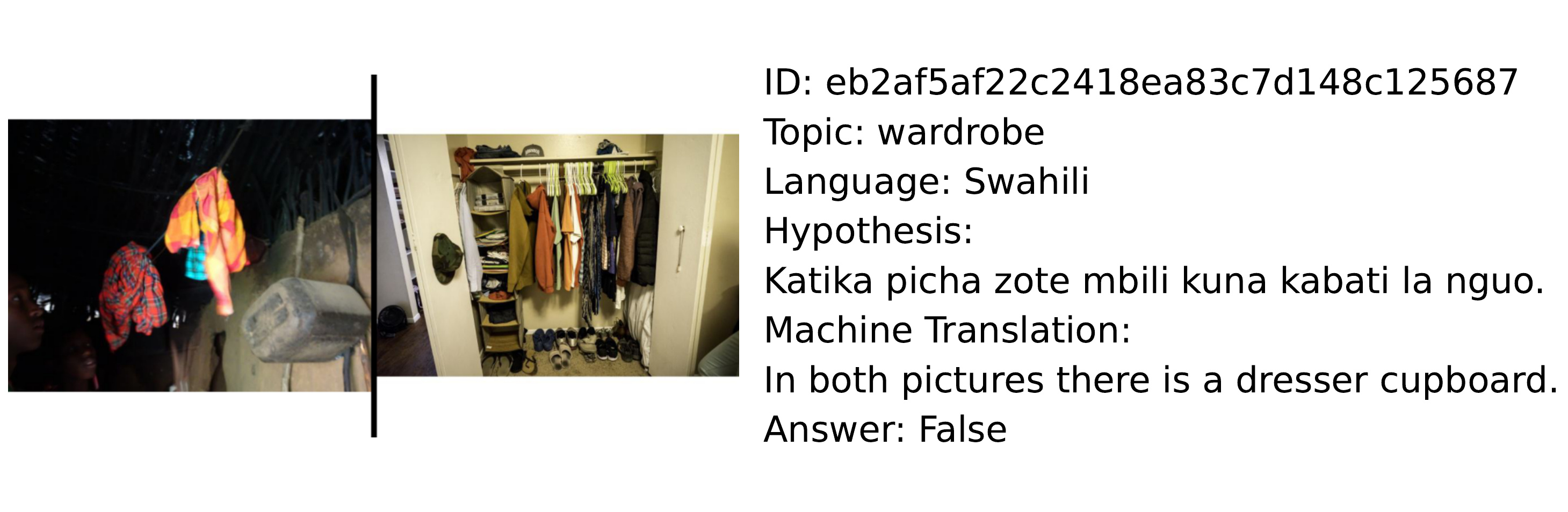}
    \caption{Swahili M5-VGR Sample.}
\end{figure*}

\begin{figure*}[!htb]
    \centering
    \includegraphics[width=1.\linewidth]{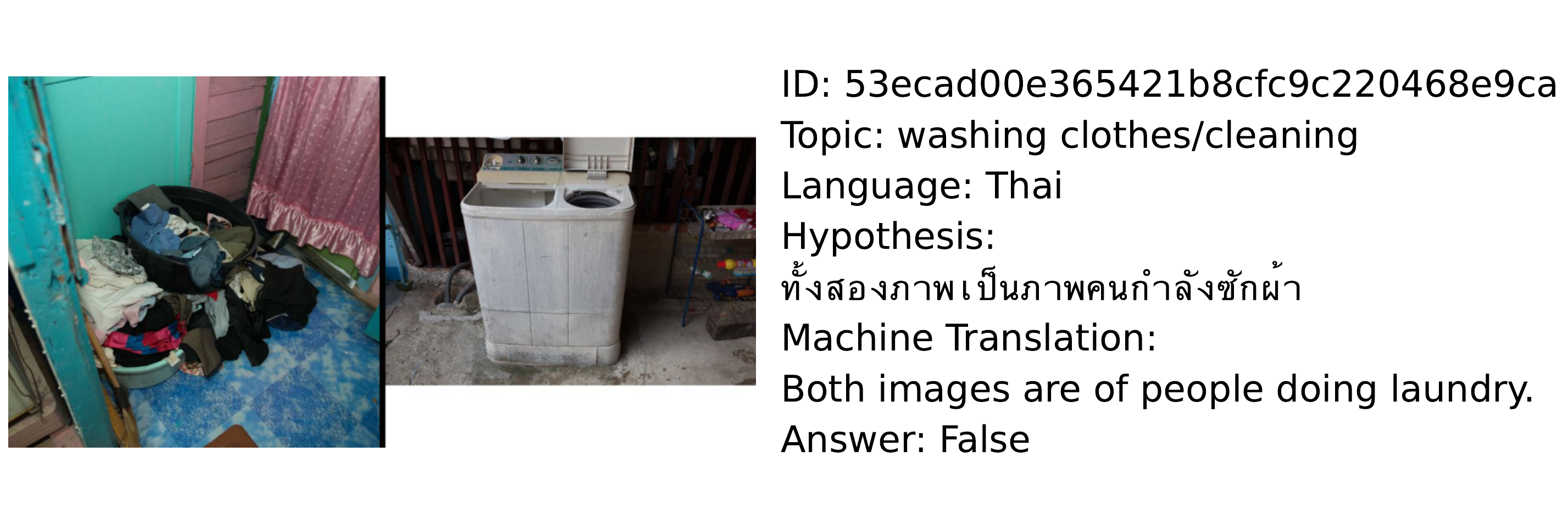}
    \caption{Thai M5-VGR Sample.}
\end{figure*}

\begin{figure*}[!htb]
    \centering
    \includegraphics[width=1.\linewidth]{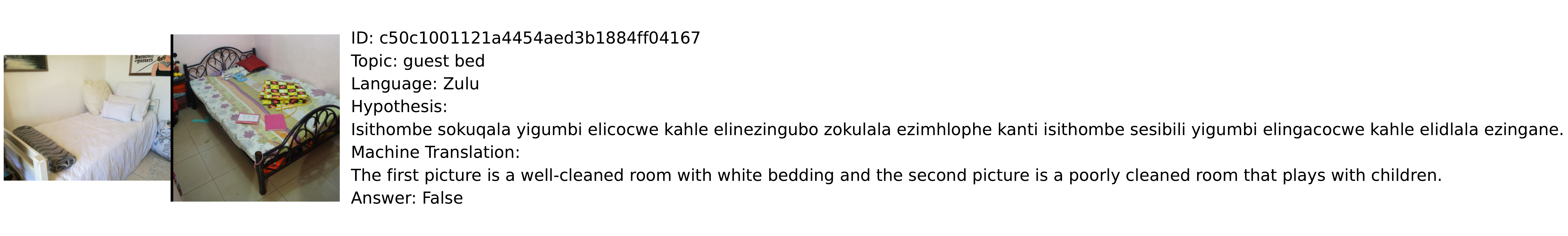}
    \caption{Zulu M5-VGR Sample.}
\end{figure*}

\FloatBarrier
\subsubsection{M5-VLOD Examples}
\label{appendix:dataset_details_m5b_examples_vlod}

\begin{figure*}[!htb]
    \centering
    \includegraphics[width=1.\linewidth]{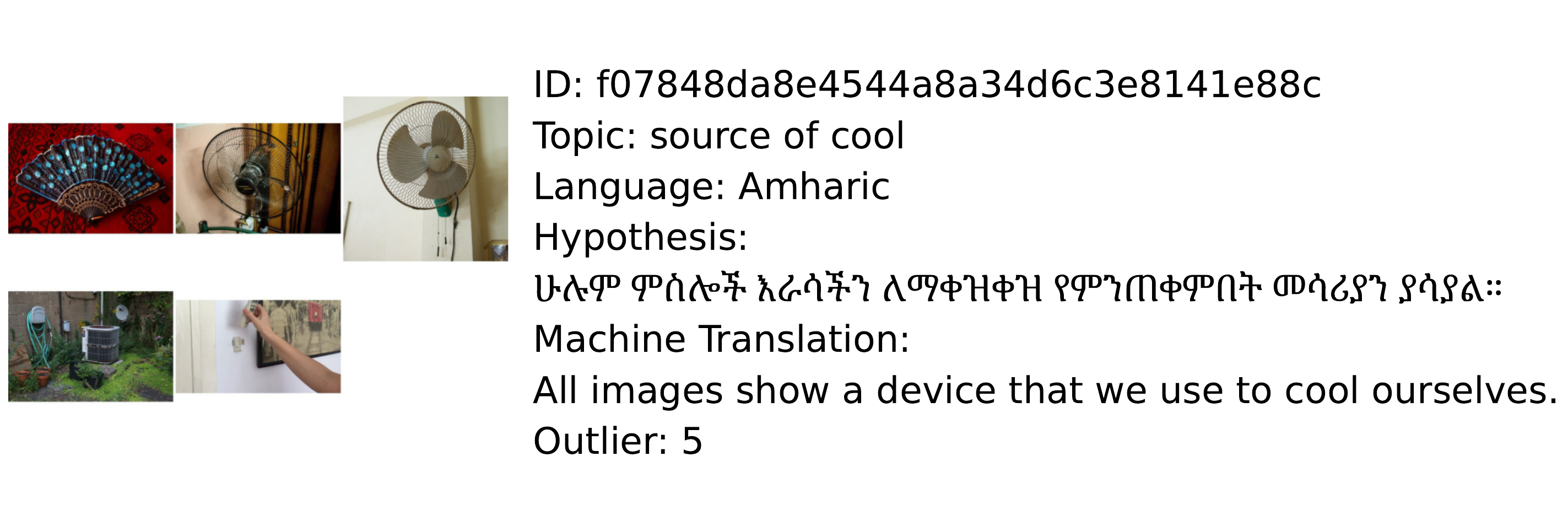}
    \caption{Amharic M5-VLOD Sample. The images are ordered from top-left to bottom-right.}
\end{figure*}

\begin{figure*}[!htb]
    \centering
    \includegraphics[width=1.\linewidth]{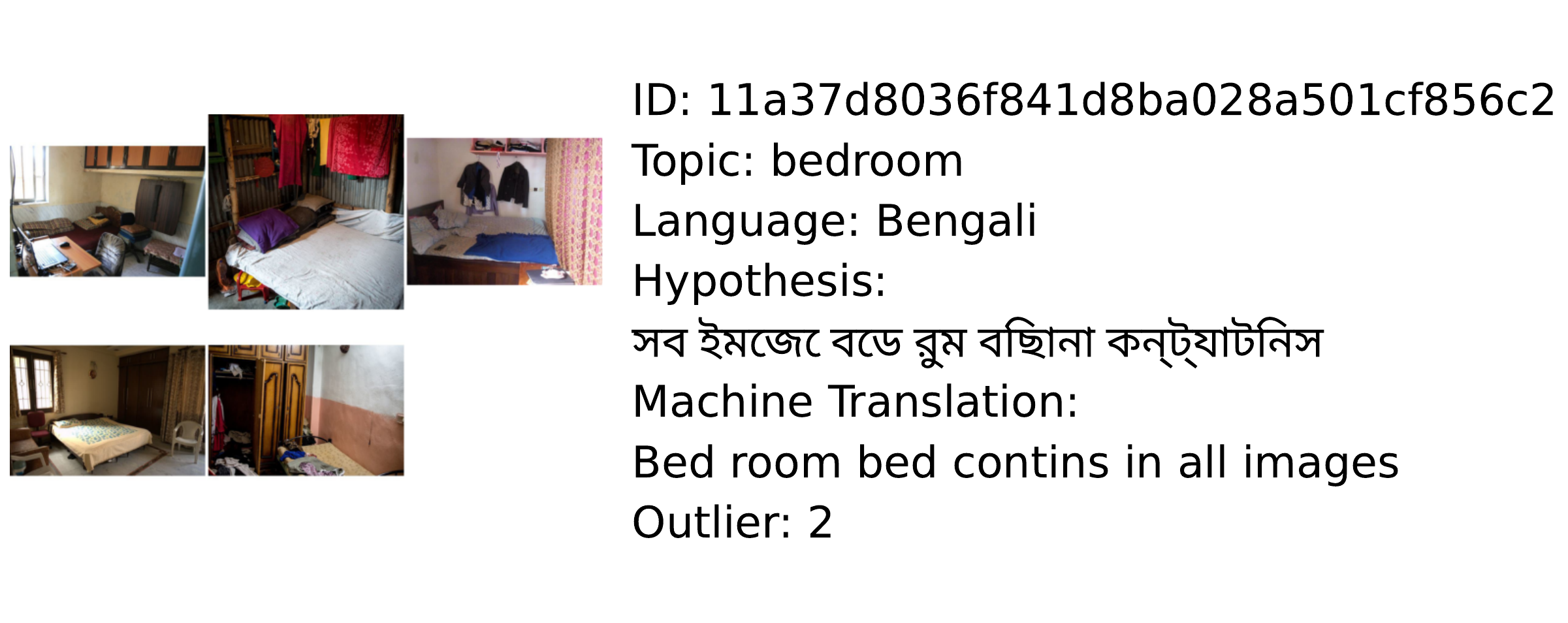}
    \caption{Bengali M5-VLOD Sample. The images are ordered from top-left to bottom-right.}
\end{figure*}

\begin{figure*}[!htb]
    \centering
    \includegraphics[width=1.\linewidth]{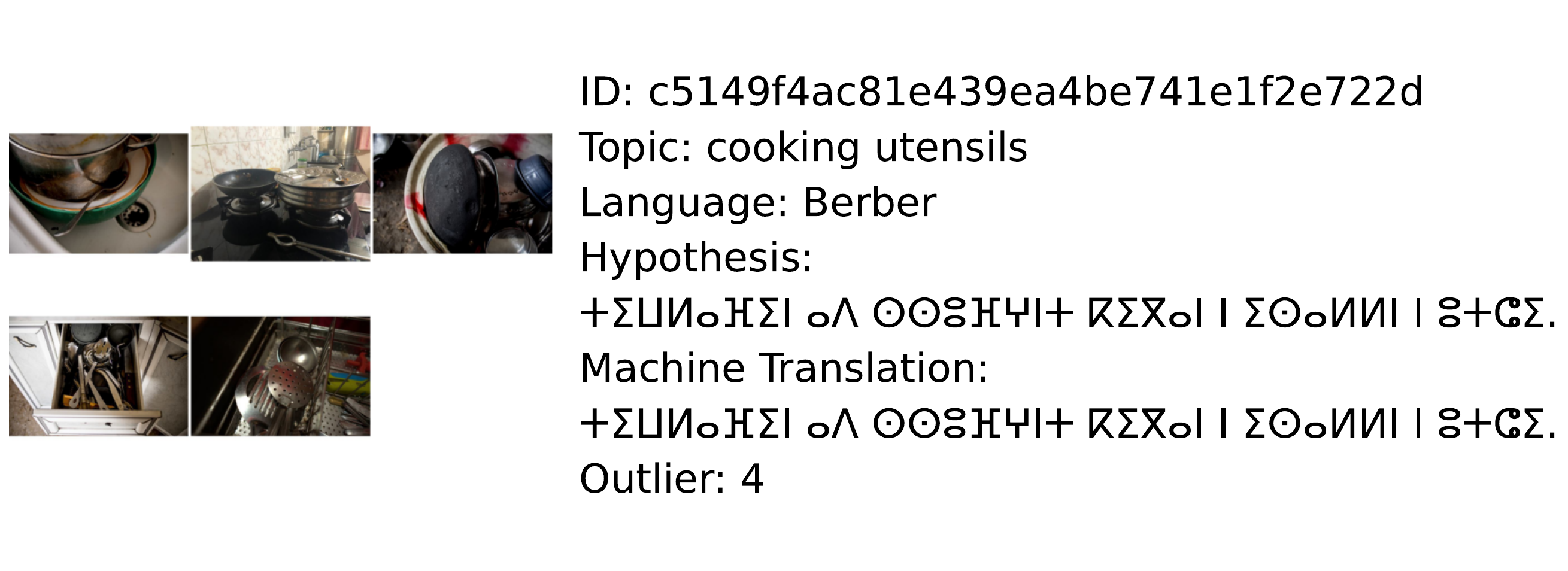}
    \caption{Berber M5-VLOD Sample. The images are ordered from top-left to bottom-right.}
\end{figure*}

\begin{figure*}[!htb]
    \centering
    \includegraphics[width=1.\linewidth]{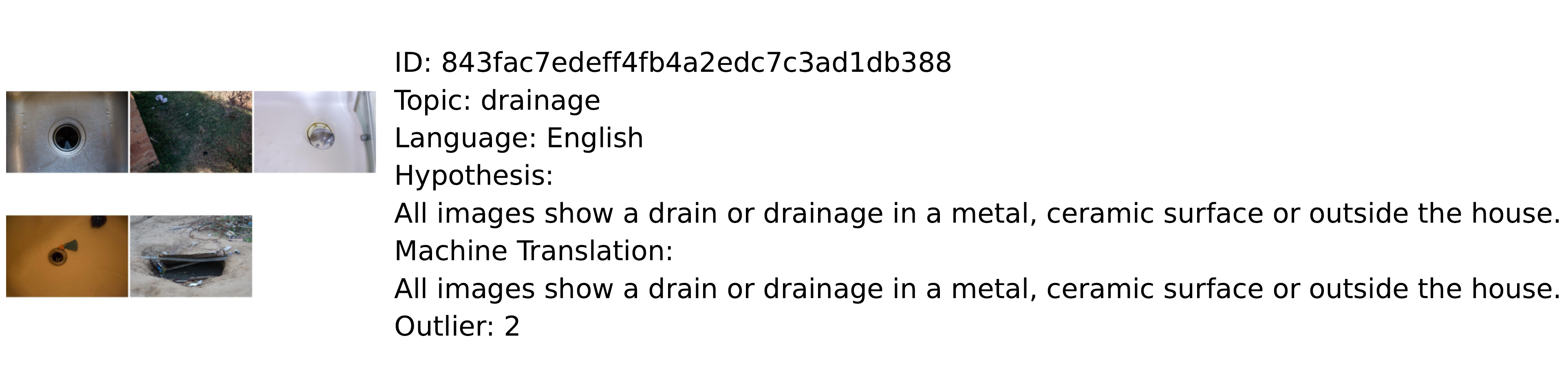}
    \caption{English M5-VLOD Sample. The images are ordered from top-left to bottom-right.}
\end{figure*}

\begin{figure*}[!htb]
    \centering
    \includegraphics[width=1.\linewidth]{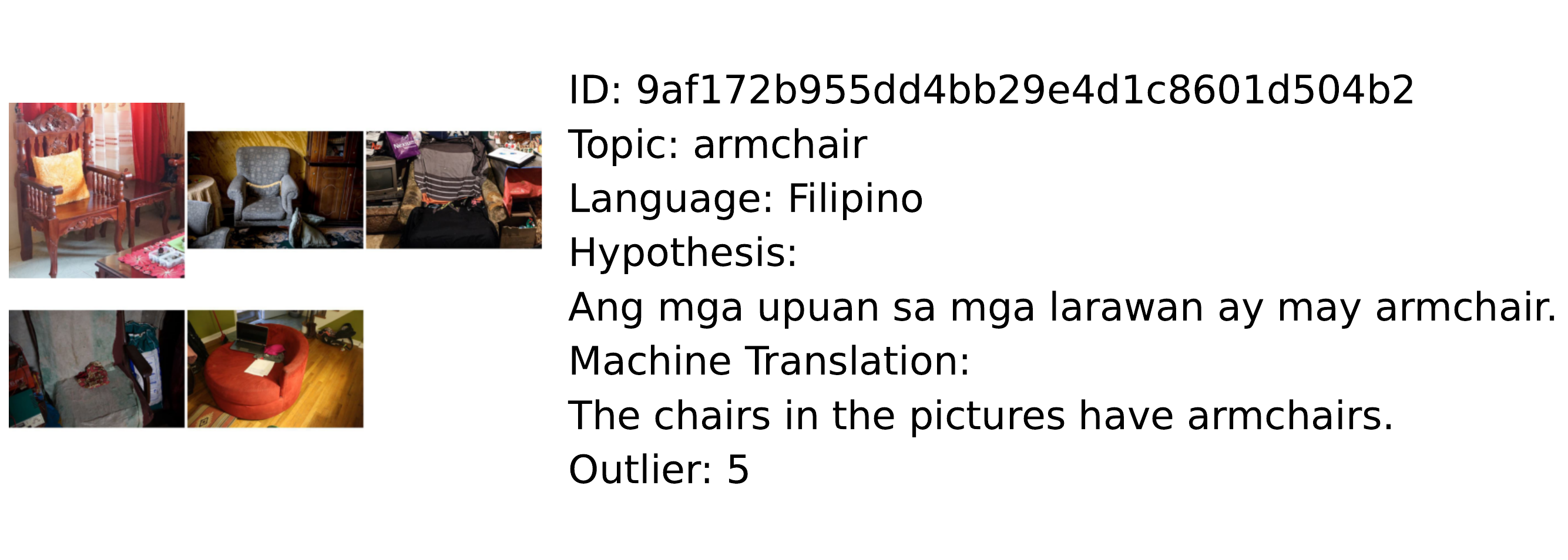}
    \caption{Filipino M5-VLOD Sample. The images are ordered from top-left to bottom-right.}
\end{figure*}

\begin{figure*}[!htb]
    \centering
    \includegraphics[width=1.\linewidth]{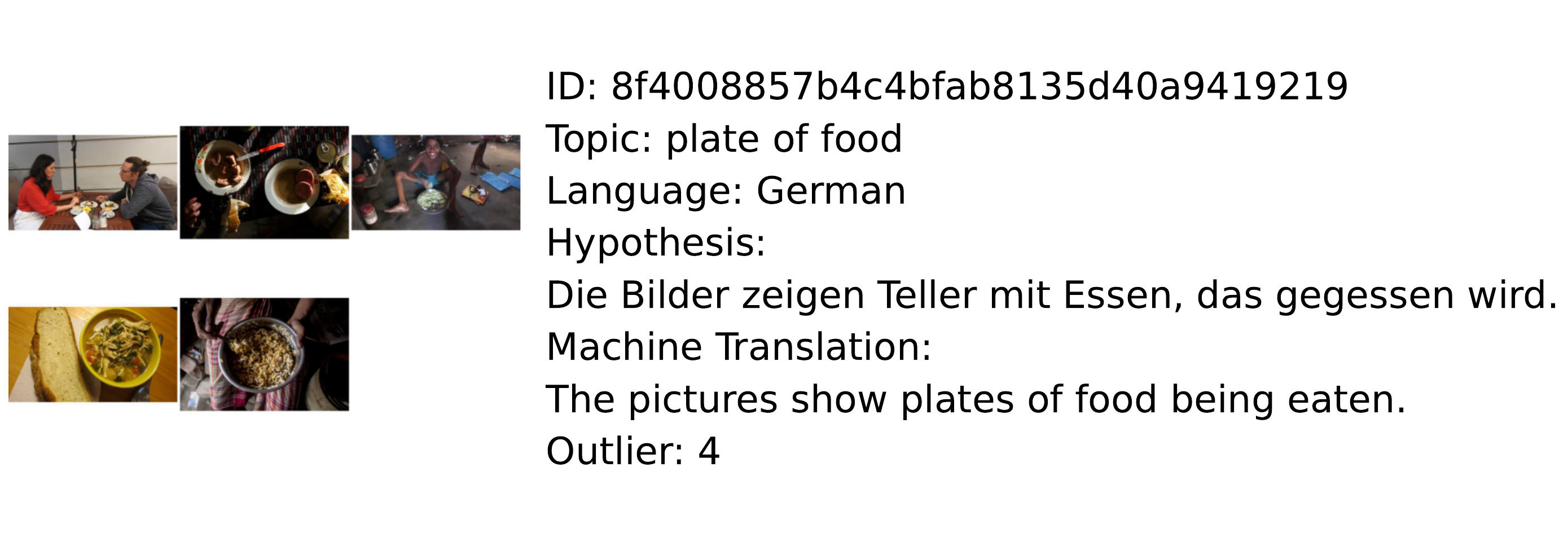}
    \caption{German M5-VLOD Sample. The images are ordered from top-left to bottom-right.}
\end{figure*}

\begin{figure*}[!htb]
    \centering
    \includegraphics[width=1.\linewidth]{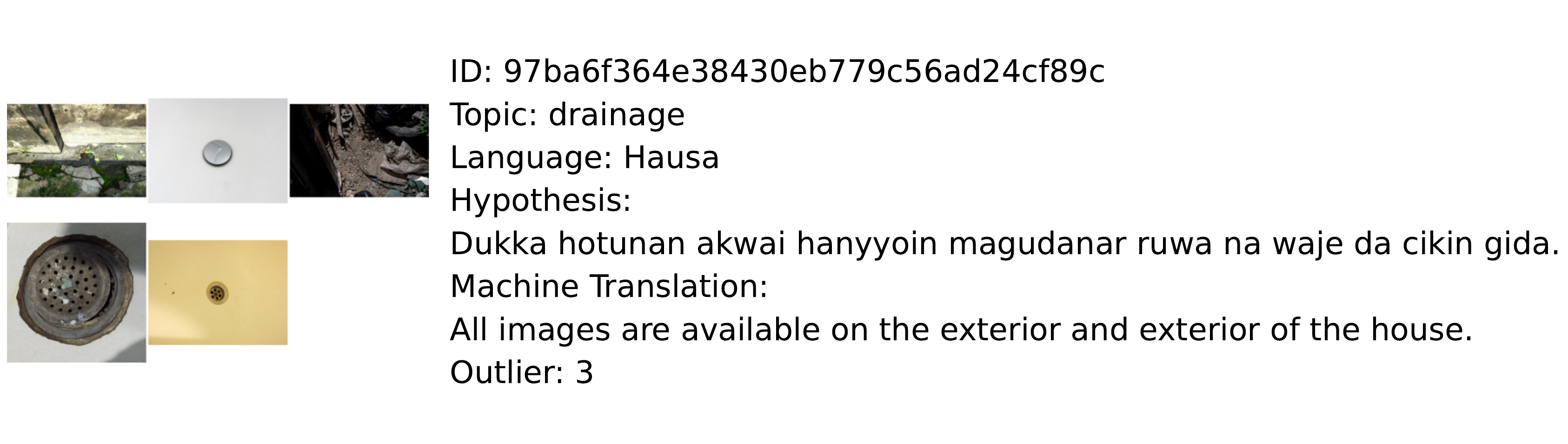}
    \caption{Hausa M5-VLOD Sample. The images are ordered from top-left to bottom-right.}
\end{figure*}

\begin{figure*}[!htb]
    \centering
    \includegraphics[width=1.\linewidth]{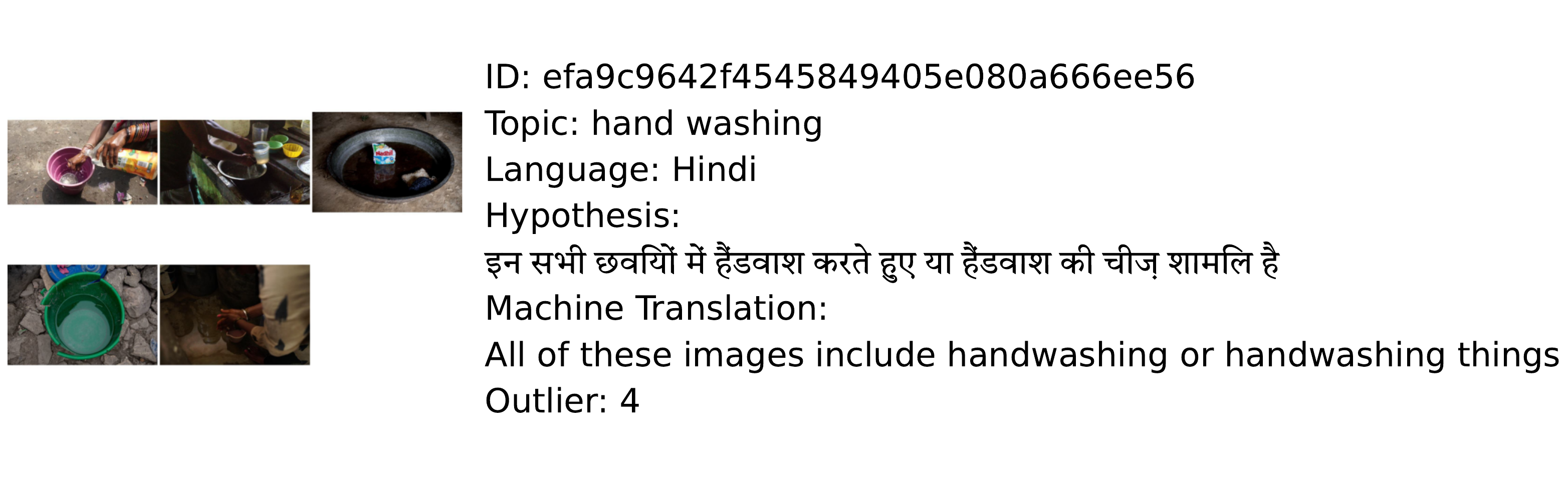}
    \caption{Hindi M5-VLOD Sample. The images are ordered from top-left to bottom-right.}
\end{figure*}

\begin{figure*}[!htb]
    \centering
    \includegraphics[width=1.\linewidth]{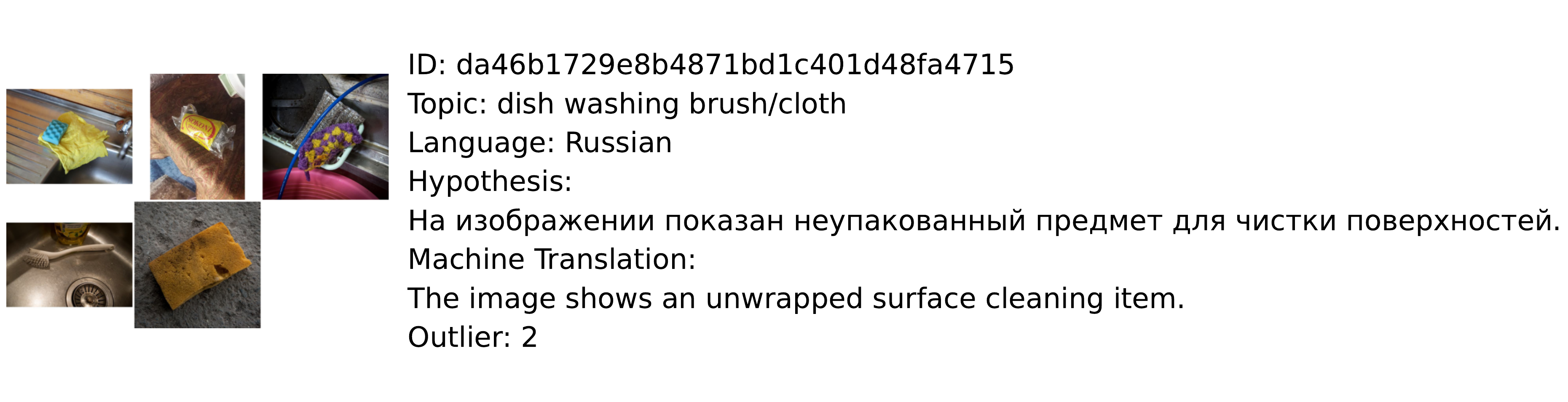}
    \caption{Russian M5-VLOD Sample. The images are ordered from top-left to bottom-right.}
\end{figure*}

\begin{figure*}[!htb]
    \centering
    \includegraphics[width=1.\linewidth]{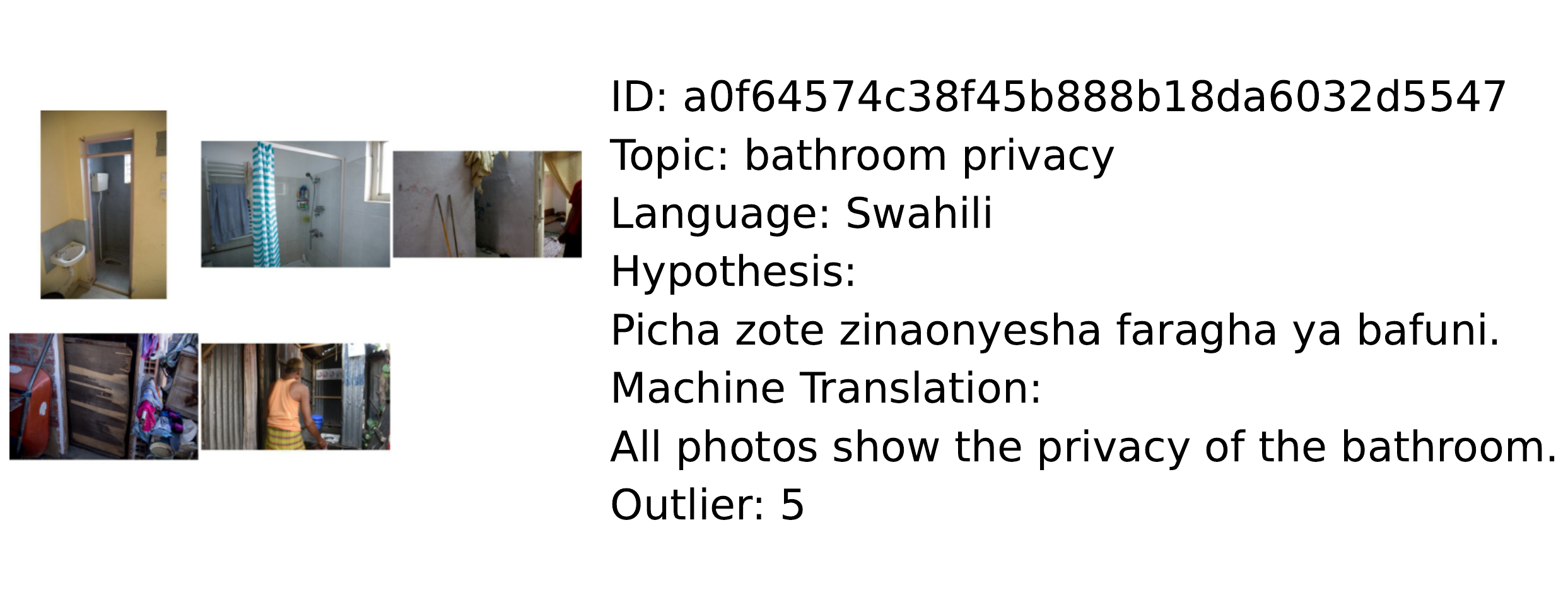}
    \caption{Swahili M5-VLOD Sample. The images are ordered from top-left to bottom-right.}
\end{figure*}

\begin{figure*}[!htb]
    \centering
    \includegraphics[width=1.\linewidth]{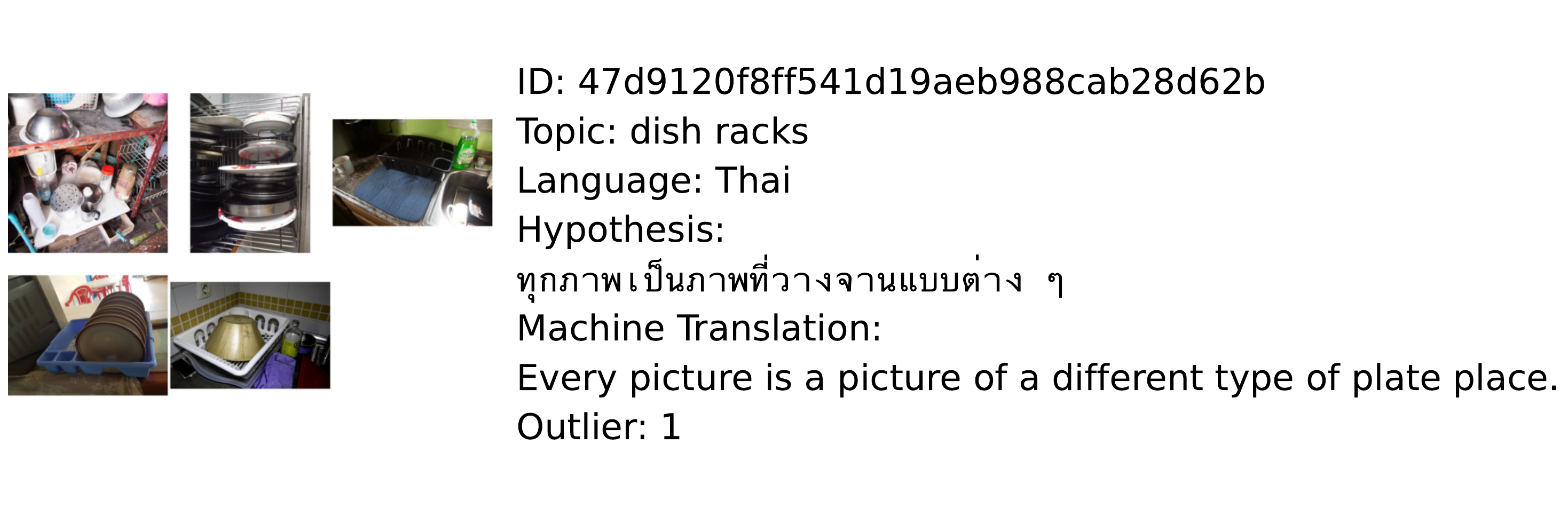}
    \caption{Thai M5-VLOD Sample. The images are ordered from top-left to bottom-right.}
\end{figure*}

\begin{figure*}[!htb]
    \centering
    \includegraphics[width=1.\linewidth]{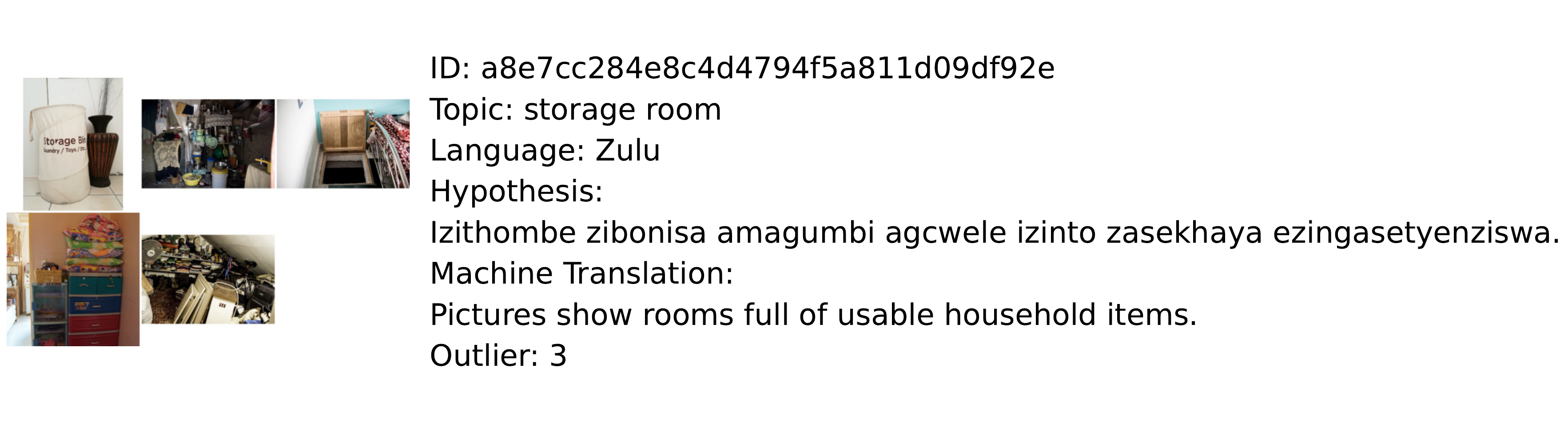}
    \caption{Zulu M5-VLOD Sample. The images are ordered from top-left to bottom-right.}
\end{figure*}

\FloatBarrier
\subsubsection{Topics}
\label{appendix:dataset_details_m5b_topics}
\begin{table}[!htb]
    \centering
    \tiny
    \caption{Number of images tagged with a certain topic in the M5-VGR (A) and M5-VLOD (B) datasets.}
    \label{tab:m5b_dataset_topics}
    \addtolength{\tabcolsep}{-0.3em}
    \begin{tabular}{@{}lcccccccccccccccccccccccc@{}}
    \toprule
    \multicolumn{1}{c}{\textbf{Topic}} &
    \multicolumn{24}{c}{\textbf{Language}} \\
     & 
     \multicolumn{2}{c}{Amharic} & 
     \multicolumn{2}{c}{Berber} &
     \multicolumn{2}{c}{Bengali} & 
     \multicolumn{2}{c}{German} & 
     \multicolumn{2}{c}{English} & 
     \multicolumn{2}{c}{Filipino} &
     \multicolumn{2}{c}{Hausa} &
     \multicolumn{2}{c}{Hindi} &
     \multicolumn{2}{c}{Russian} & 
     \multicolumn{2}{c}{Swahili} & 
     \multicolumn{2}{c}{Thai} &
     \multicolumn{2}{c}{Zulu} \\
     & 
     \multicolumn{1}{c}{A} & 
     \multicolumn{1}{c}{B} &
     \multicolumn{1}{c}{A} & 
     \multicolumn{1}{c}{B} &
     \multicolumn{1}{c}{A} & 
     \multicolumn{1}{c}{B} &
     \multicolumn{1}{c}{A} & 
     \multicolumn{1}{c}{B} &
     \multicolumn{1}{c}{A} & 
     \multicolumn{1}{c}{B} &
     \multicolumn{1}{c}{A} & 
     \multicolumn{1}{c}{B} &
     \multicolumn{1}{c}{A} & 
     \multicolumn{1}{c}{B} &
     \multicolumn{1}{c}{A} & 
     \multicolumn{1}{c}{B} &
     \multicolumn{1}{c}{A} & 
     \multicolumn{1}{c}{B} &
     \multicolumn{1}{c}{A} & 
     \multicolumn{1}{c}{B} &
     \multicolumn{1}{c}{A} & 
     \multicolumn{1}{c}{B} &
     \multicolumn{1}{c}{A} & 
     \multicolumn{1}{c}{B} \\
    \midrule
    armchair & $1$ & $2$ & $1$ & $1$ & $1$ & $1$ & $1$ & $1$ & $1$ & $2$ & $3$ & $1$ & $1$ & $1$ & $1$ & $1$ & $2$ & $1$ & $3$ & $1$ & $1$ & $1$ & $1$ & $1$ \\
    backyard & $1$ & $1$ & $1$ & $2$ & $1$ & $1$ & $1$ & $1$ & $1$ & $1$ & $1$ & $1$ & $1$ & $1$ & $1$ & $1$ & $1$ & $1$ & $1$ & $1$ & $1$ & $1$ & $3$ & $1$ \\
    bathroom privacy & $1$ & $1$ & $3$ & $3$ & $1$ & $1$ & $2$ & $1$ & $1$ & $1$ & $3$ & $4$ & $1$ & $1$ & $1$ & $1$ & $1$ & $1$ & $2$ & $1$ & $1$ & $1$ & $1$ & $1$ \\
    bathroom/toilet & $1$ & $2$ & $3$ & $1$ & $1$ & $2$ & $1$ & $3$ & $2$ & $1$ & $1$ & $1$ & $2$ & $1$ & $1$ & $1$ & $1$ & $1$ & $3$ & $3$ & $1$ & $1$ & $1$ & $1$ \\
    bed & $1$ & $1$ & $1$ & $2$ & $2$ & $1$ & $1$ & $1$ & $1$ & $3$ & $1$ & $2$ & $4$ & $1$ & $1$ & $1$ & $2$ & $2$ & $1$ & $1$ & $4$ & $1$ & $1$ & $1$ \\
    bedroom & $2$ & $4$ & $1$ & $2$ & $2$ & $2$ & $1$ & $1$ & $1$ & $2$ & $1$ & $1$ & $3$ & $1$ & $1$ & $2$ & $1$ & $1$ & $2$ & $1$ & $1$ & $1$ & $1$ & $1$ \\
    books & $2$ & $2$ & $1$ & $1$ & $1$ & $2$ & $1$ & $1$ & $1$ & $1$ & $2$ & $1$ & $1$ & $1$ & $1$ & $1$ & $2$ & $1$ & $1$ & $1$ & $1$ & $1$ & $1$ & $1$ \\
    ceiling & $1$ & $2$ & $1$ & $1$ & $2$ & $1$ & $1$ & $1$ & $2$ & $2$ & $1$ & $1$ & $1$ & $4$ & $2$ & $1$ & $2$ & $1$ & $2$ & $2$ & $2$ & $1$ & $1$ & $2$ \\
    children room & $1$ & $1$ & $2$ & $1$ & $1$ & $1$ & $1$ & $1$ & $1$ & $1$ & $1$ & $1$ & $1$ & $1$ & $1$ & $1$ & $2$ & $2$ & $1$ & $1$ & $2$ & $1$ & $1$ & $1$ \\
    cleaning equipment & $1$ & $1$ & $1$ & $1$ & $1$ & $1$ & $1$ & $1$ & $1$ & $1$ & $1$ & $1$ & $2$ & $1$ & $1$ & $1$ & $1$ & $1$ & $1$ & $1$ & $1$ & $1$ & $1$ & $1$ \\
    cooking pots & $1$ & $1$ & $2$ & $2$ & $2$ & $1$ & $1$ & $1$ & $1$ & $1$ & $2$ & $1$ & $2$ & $2$ & $1$ & $1$ & $2$ & $1$ & $1$ & $2$ & $1$ & $2$ & $1$ & $1$ \\
    cooking utensils & $1$ & $1$ & $3$ & $2$ & $1$ & $3$ & $1$ & $1$ & $1$ & $1$ & $1$ & $1$ & $1$ & $3$ & $2$ & $1$ & $1$ & $1$ & $1$ & $1$ & $2$ & $1$ & $1$ & $1$ \\
    couch & $1$ & $1$ & $1$ & $1$ & $1$ & $1$ & $1$ & $2$ & $2$ & $1$ & $3$ & $3$ & $3$ & $1$ & $1$ & $1$ & $2$ & $1$ & $2$ & $1$ & $1$ & $1$ & $3$ & $1$ \\
    cups/mugs/glasses & $1$ & $1$ & $1$ & $1$ & $1$ & $1$ & $1$ & $1$ & $3$ & $1$ & $1$ & $1$ & $1$ & $2$ & $1$ & $1$ & $1$ & $1$ & $1$ & $1$ & $1$ & $1$ & $1$ & $1$ \\
    cutlery & $1$ & $1$ & $1$ & $1$ & $1$ & $1$ & $1$ & $1$ & $2$ & $1$ & $3$ & $1$ & $1$ & $1$ & $1$ & $2$ & $1$ & $1$ & $1$ & $1$ & $1$ & $2$ & $1$ & $1$ \\
    dish racks & $1$ & $1$ & $1$ & $2$ & $1$ & $1$ & $2$ & $1$ & $1$ & $1$ & $1$ & $1$ & $1$ & $1$ & $1$ & $1$ & $1$ & $1$ & $1$ & $1$ & $3$ & $2$ & $1$ & $1$ \\
    dish washing brush/cloth & $2$ & $1$ & $1$ & $1$ & $3$ & $1$ & $1$ & $1$ & $1$ & $3$ & $1$ & $1$ & $1$ & $1$ & $3$ & $2$ & $1$ & $2$ & $1$ & $1$ & $1$ & $1$ & $1$ & $1$ \\
    dish washing soap & $1$ & $1$ & $1$ & $1$ & $1$ & $1$ & $1$ & $1$ & $3$ & $1$ & $1$ & $1$ & $1$ & $3$ & $1$ & $2$ & $2$ & $2$ & $2$ & $2$ & $1$ & $2$ & $1$ & $1$ \\
    drainage & $1$ & $2$ & $1$ & $1$ & $1$ & $1$ & $1$ & $1$ & $1$ & $2$ & $1$ & $2$ & $1$ & $1$ & $1$ & $1$ & $1$ & $1$ & $1$ & $1$ & $1$ & $1$ & $1$ & $1$ \\
    drinking water & $3$ & $4$ & $2$ & $2$ & $1$ & $2$ & $1$ & $1$ & $1$ & $1$ & $1$ & $1$ & $2$ & $1$ & $1$ & $1$ & $1$ & $1$ & $1$ & $1$ & $1$ & $1$ & $4$ & $2$ \\
    drying & $3$ & $1$ & $1$ & $1$ & $5$ & $1$ & $1$ & $1$ & $1$ & $1$ & $2$ & $2$ & $1$ & $2$ & $3$ & $1$ & $1$ & $1$ & $1$ & $1$ & $1$ & $1$ & $1$ & $1$ \\
    everyday shoes & $1$ & $2$ & $1$ & $2$ & $2$ & $1$ & $3$ & $1$ & $1$ & $1$ & $2$ & $1$ & $2$ & $3$ & $1$ & $1$ & $1$ & $2$ & $1$ & $2$ & $2$ & $1$ & $2$ & $2$ \\
    family & $2$ & $2$ & $4$ & $1$ & $2$ & $1$ & $3$ & $2$ & $1$ & $1$ & $2$ & $1$ & $3$ & $3$ & $1$ & $1$ & $1$ & $2$ & $1$ & $1$ & $2$ & $2$ & $2$ & $2$ \\
    floor & $1$ & $1$ & $3$ & $1$ & $2$ & $1$ & $1$ & $1$ & $1$ & $1$ & $1$ & $1$ & $1$ & $1$ & $1$ & $1$ & $2$ & $1$ & $1$ & $2$ & $1$ & $1$ & $1$ & $1$ \\
    front door & $2$ & $1$ & $4$ & $1$ & $1$ & $1$ & $1$ & $1$ & $1$ & $1$ & $1$ & $1$ & $1$ & $3$ & $2$ & $1$ & $1$ & $1$ & $4$ & $1$ & $3$ & $2$ & $1$ & $1$ \\
    grains & $2$ & $1$ & $1$ & $1$ & $2$ & $1$ & $2$ & $1$ & $1$ & $1$ & $1$ & $1$ & $1$ & $1$ & $1$ & $2$ & $2$ & $2$ & $2$ & $1$ & $1$ & $1$ & $1$ & $1$ \\
    guest bed & $3$ & $1$ & $1$ & $1$ & $1$ & $1$ & $1$ & $1$ & $1$ & $1$ & $2$ & $1$ & $1$ & $1$ & $1$ & $1$ & $2$ & $2$ & $1$ & $1$ & $1$ & $1$ & $1$ & $1$ \\
    hair brush/comb & $1$ & $1$ & $1$ & $1$ & $3$ & $1$ & $2$ & $3$ & $1$ & $1$ & $1$ & $1$ & $2$ & $2$ & $1$ & $1$ & $2$ & $1$ & $1$ & $2$ & $1$ & $1$ & $3$ & $1$ \\
    hand back & $1$ & $2$ & $1$ & $1$ & $1$ & $1$ & $1$ & $1$ & $1$ & $1$ & $2$ & $2$ & $3$ & $1$ & $2$ & $1$ & $2$ & $1$ & $1$ & $1$ & $1$ & $2$ & $1$ & $2$ \\
    hand palm & $1$ & $1$ & $3$ & $2$ & $2$ & $1$ & $1$ & $1$ & $1$ & $1$ & $1$ & $1$ & $1$ & $1$ & $1$ & $1$ & $1$ & $1$ & $2$ & $1$ & $2$ & $2$ & $1$ & $1$ \\
    hand washing & $2$ & $1$ & $3$ & $2$ & $1$ & $1$ & $1$ & $5$ & $1$ & $1$ & $1$ & $4$ & $2$ & $1$ & $1$ & $3$ & $2$ & $2$ & $2$ & $1$ & $2$ & $1$ & $2$ & $3$ \\
    home & $1$ & $1$ & $3$ & $2$ & $2$ & $1$ & $2$ & $1$ & $1$ & $2$ & $1$ & $4$ & $1$ & $1$ & $5$ & $1$ & $1$ & $2$ & $1$ & $2$ & $2$ & $2$ & $1$ & $2$ \\
    jewelry & $1$ & $1$ & $1$ & $1$ & $1$ & $2$ & $1$ & $1$ & $1$ & $2$ & $1$ & $1$ & $1$ & $2$ & $1$ & $1$ & $1$ & $1$ & $2$ & $1$ & $1$ & $1$ & $1$ & $1$ \\
    kitchen & $2$ & $1$ & $1$ & $2$ & $1$ & $1$ & $1$ & $1$ & $4$ & $1$ & $2$ & $2$ & $1$ & $1$ & $1$ & $1$ & $2$ & $2$ & $1$ & $2$ & $2$ & $2$ & $1$ & $1$ \\
    kitchen sink & $1$ & $2$ & $2$ & $2$ & $1$ & $1$ & $4$ & $2$ & $1$ & $2$ & $2$ & $1$ & $1$ & $2$ & $1$ & $2$ & $1$ & $1$ & $1$ & $3$ & $1$ & $1$ & $3$ & $3$ \\
    light source in kitchen & $1$ & $1$ & $2$ & $1$ & $2$ & $3$ & $2$ & $2$ & $3$ & $2$ & $1$ & $1$ & $1$ & $1$ & $3$ & $1$ & $1$ & $1$ & $2$ & $1$ & $1$ & $1$ & $1$ & $1$ \\
    light source in livingroom & $1$ & $2$ & $2$ & $2$ & $1$ & $1$ & $1$ & $1$ & $2$ & $1$ & $1$ & $2$ & $1$ & $1$ & $1$ & $1$ & $2$ & $1$ & $1$ & $1$ & $1$ & $1$ & $1$ & $1$ \\
    living room & $1$ & $1$ & $1$ & $1$ & $1$ & $1$ & $1$ & $2$ & $2$ & $1$ & $3$ & $1$ & $1$ & $1$ & $1$ & $1$ & $2$ & $2$ & $1$ & $1$ & $2$ & $1$ & $1$ & $1$ \\
    lock on front door & $1$ & $1$ & $1$ & $1$ & $1$ & $4$ & $1$ & $1$ & $3$ & $1$ & $1$ & $1$ & $2$ & $1$ & $1$ & $3$ & $1$ & $2$ & $1$ & $1$ & $2$ & $3$ & $1$ & $1$ \\
    make up & $1$ & $1$ & $1$ & $2$ & $1$ & $1$ & $2$ & $1$ & $1$ & $1$ & $1$ & $1$ & $1$ & $1$ & $1$ & $1$ & $1$ & $1$ & $1$ & $1$ & $2$ & $1$ & $2$ & $1$ \\
    meat or fish & $1$ & $1$ & $1$ & $1$ & $1$ & $1$ & $1$ & $1$ & $1$ & $1$ & $1$ & $1$ & $1$ & $2$ & $1$ & $1$ & $1$ & $1$ & $1$ & $1$ & $1$ & $1$ & $1$ & $1$ \\
    medication & $1$ & $1$ & $1$ & $1$ & $1$ & $2$ & $1$ & $1$ & $1$ & $2$ & $1$ & $1$ & $2$ & $1$ & $1$ & $1$ & $2$ & $1$ & $1$ & $1$ & $1$ & $1$ & $1$ & $1$ \\
    most loved item & $1$ & $1$ & $1$ & $1$ & $1$ & $2$ & $3$ & $1$ & $2$ & $2$ & $3$ & $3$ & $2$ & $1$ & $2$ & $2$ & $2$ & $2$ & $2$ & $0$ & $1$ & $1$ & $4$ & $4$ \\
    most loved toy & $1$ & $1$ & $1$ & $1$ & $1$ & $2$ & $1$ & $1$ & $1$ & $1$ & $1$ & $1$ & $1$ & $1$ & $1$ & $1$ & $1$ & $2$ & $1$ & $1$ & $1$ & $2$ & $1$ & $1$ \\
    nicest shoes & $1$ & $1$ & $1$ & $1$ & $1$ & $2$ & $2$ & $2$ & $2$ & $1$ & $1$ & $1$ & $1$ & $1$ & $2$ & $1$ & $1$ & $1$ & $1$ & $2$ & $2$ & $2$ & $1$ & $1$ \\
    oven & $1$ & $1$ & $1$ & $1$ & $1$ & $1$ & $1$ & $1$ & $1$ & $1$ & $1$ & $1$ & $1$ & $2$ & $1$ & $1$ & $2$ & $2$ & $1$ & $1$ & $1$ & $1$ & $1$ & $1$ \\
    paper & $2$ & $1$ & $1$ & $2$ & $1$ & $2$ & $2$ & $1$ & $2$ & $1$ & $1$ & $1$ & $1$ & $1$ & $1$ & $1$ & $1$ & $1$ & $1$ & $1$ & $1$ & $1$ & $1$ & $1$ \\
    pen/pencils & $1$ & $1$ & $1$ & $2$ & $1$ & $2$ & $1$ & $1$ & $1$ & $1$ & $1$ & $1$ & $1$ & $1$ & $1$ & $1$ & $1$ & $2$ & $3$ & $1$ & $1$ & $1$ & $1$ & $1$ \\
    phone & $2$ & $2$ & $1$ & $1$ & $2$ & $1$ & $1$ & $1$ & $2$ & $3$ & $1$ & $3$ & $2$ & $1$ & $2$ & $1$ & $2$ & $1$ & $1$ & $3$ & $1$ & $1$ & $2$ & $1$ \\
    place where eating dinner & $1$ & $2$ & $1$ & $1$ & $1$ & $1$ & $1$ & $1$ & $1$ & $1$ & $1$ & $1$ & $1$ & $1$ & $2$ & $1$ & $1$ & $2$ & $1$ & $1$ & $2$ & $2$ & $1$ & $1$ \\
    plate of food & $2$ & $1$ & $1$ & $4$ & $1$ & $1$ & $2$ & $3$ & $1$ & $1$ & $2$ & $1$ & $3$ & $1$ & $1$ & $1$ & $1$ & $2$ & $1$ & $1$ & $1$ & $1$ & $1$ & $1$ \\
    plates & $2$ & $1$ & $1$ & $1$ & $2$ & $2$ & $1$ & $1$ & $1$ & $1$ & $1$ & $3$ & $1$ & $1$ & $1$ & $1$ & $1$ & $2$ & $1$ & $1$ & $2$ & $1$ & $1$ & $1$ \\
    play area & $1$ & $1$ & $2$ & $2$ & $1$ & $1$ & $1$ & $1$ & $1$ & $1$ & $2$ & $1$ & $1$ & $2$ & $1$ & $1$ & $1$ & $1$ & $1$ & $1$ & $1$ & $1$ & $2$ & $1$ \\
    power outlet & $1$ & $2$ & $1$ & $1$ & $1$ & $1$ & $3$ & $4$ & $2$ & $1$ & $1$ & $1$ & $1$ & $1$ & $3$ & $1$ & $1$ & $1$ & $1$ & $1$ & $1$ & $4$ & $2$ & $1$ \\
    refrigerator & $1$ & $1$ & $1$ & $1$ & $2$ & $1$ & $4$ & $4$ & $1$ & $3$ & $1$ & $3$ & $1$ & $1$ & $1$ & $3$ & $2$ & $1$ & $1$ & $1$ & $1$ & $1$ & $1$ & $1$ \\
    roof & $2$ & $1$ & $1$ & $1$ & $1$ & $3$ & $2$ & $2$ & $1$ & $1$ & $1$ & $1$ & $1$ & $1$ & $1$ & $1$ & $1$ & $1$ & $1$ & $1$ & $2$ & $1$ & $1$ & $1$ \\
    shampoo & $1$ & $2$ & $1$ & $1$ & $1$ & $1$ & $1$ & $1$ & $1$ & $1$ & $1$ & $1$ & $1$ & $1$ & $2$ & $1$ & $1$ & $2$ & $2$ & $1$ & $1$ & $2$ & $1$ & $1$ \\
    shower & $1$ & $1$ & $1$ & $2$ & $1$ & $1$ & $1$ & $1$ & $1$ & $2$ & $1$ & $2$ & $1$ & $1$ & $1$ & $1$ & $2$ & $2$ & $1$ & $1$ & $1$ & $1$ & $1$ & $1$ \\
    sitting area & $1$ & $1$ & $1$ & $2$ & $1$ & $1$ & $1$ & $1$ & $1$ & $1$ & $1$ & $1$ & $1$ & $1$ & $1$ & $1$ & $1$ & $1$ & $1$ & $1$ & $1$ & $1$ & $1$ & $1$ \\
    soap for hands and body & $1$ & $1$ & $2$ & $2$ & $2$ & $2$ & $1$ & $2$ & $1$ & $1$ & $1$ & $1$ & $2$ & $1$ & $2$ & $1$ & $2$ & $1$ & $1$ & $1$ & $2$ & $2$ & $1$ & $1$ \\
    social drink & $1$ & $1$ & $1$ & $1$ & $1$ & $1$ & $2$ & $2$ & $1$ & $1$ & $1$ & $1$ & $1$ & $1$ & $1$ & $1$ & $1$ & $1$ & $1$ & $1$ & $1$ & $1$ & $1$ & $2$ \\
    sofa & $1$ & $1$ & $1$ & $1$ & $1$ & $1$ & $1$ & $1$ & $1$ & $2$ & $1$ & $1$ & $2$ & $4$ & $1$ & $1$ & $1$ & $1$ & $1$ & $1$ & $2$ & $1$ & $1$ & $1$ \\
    source of cool & $1$ & $1$ & $1$ & $1$ & $2$ & $1$ & $1$ & $1$ & $2$ & $1$ & $1$ & $1$ & $2$ & $1$ & $2$ & $2$ & $1$ & $1$ & $1$ & $1$ & $1$ & $1$ & $1$ & $2$ \\
    spices & $1$ & $2$ & $1$ & $1$ & $2$ & $1$ & $1$ & $3$ & $3$ & $3$ & $2$ & $1$ & $1$ & $1$ & $2$ & $1$ & $2$ & $2$ & $1$ & $2$ & $1$ & $2$ & $1$ & $3$ \\
    storage room & $1$ & $2$ & $1$ & $1$ & $1$ & $2$ & $1$ & $1$ & $1$ & $1$ & $5$ & $1$ & $2$ & $1$ & $1$ & $2$ & $1$ & $2$ & $1$ & $1$ & $1$ & $1$ & $1$ & $1$ \\
    stove/hob & $2$ & $1$ & $1$ & $2$ & $1$ & $1$ & $1$ & $1$ & $1$ & $3$ & $1$ & $1$ & $1$ & $5$ & $1$ & $2$ & $1$ & $1$ & $2$ & $3$ & $2$ & $1$ & $1$ & $4$ \\
    street detail & $4$ & $1$ & $1$ & $1$ & $1$ & $1$ & $1$ & $1$ & $1$ & $3$ & $2$ & $1$ & $1$ & $2$ & $1$ & $2$ & $2$ & $1$ & $2$ & $1$ & $2$ & $1$ & $1$ & $1$ \\
    street view & $1$ & $1$ & $1$ & $2$ & $1$ & $4$ & $2$ & $1$ & $1$ & $1$ & $1$ & $1$ & $1$ & $2$ & $1$ & $3$ & $1$ & $1$ & $2$ & $1$ & $2$ & $2$ & $1$ & $2$ \\
    switch on/off & $2$ & $1$ & $1$ & $1$ & $1$ & $1$ & $1$ & $1$ & $1$ & $2$ & $1$ & $1$ & $1$ & $1$ & $1$ & $1$ & $1$ & $1$ & $1$ & $1$ & $2$ & $3$ & $1$ & $1$ \\
    table with food & $2$ & $4$ & $1$ & $1$ & $1$ & $1$ & $2$ & $1$ & $2$ & $1$ & $1$ & $1$ & $1$ & $1$ & $1$ & $1$ & $1$ & $1$ & $1$ & $1$ & $1$ & $1$ & $1$ & $1$ \\
    teeth & $1$ & $1$ & $1$ & $1$ & $1$ & $1$ & $1$ & $1$ & $1$ & $2$ & $1$ & $2$ & $3$ & $1$ & $1$ & $1$ & $2$ & $2$ & $2$ & $2$ & $1$ & $2$ & $1$ & $1$ \\
    toilet & $1$ & $2$ & $1$ & $2$ & $1$ & $1$ & $1$ & $1$ & $1$ & $1$ & $2$ & $1$ & $1$ & $1$ & $1$ & $1$ & $2$ & $1$ & $1$ & $2$ & $1$ & $1$ & $1$ & $1$ \\
    toilet paper & $3$ & $2$ & $1$ & $1$ & $1$ & $1$ & $1$ & $1$ & $1$ & $1$ & $1$ & $1$ & $1$ & $1$ & $1$ & $1$ & $1$ & $2$ & $1$ & $1$ & $1$ & $1$ & $3$ & $2$ \\
    tooth paste & $2$ & $1$ & $1$ & $1$ & $1$ & $2$ & $2$ & $3$ & $2$ & $2$ & $2$ & $1$ & $1$ & $1$ & $4$ & $1$ & $2$ & $2$ & $3$ & $1$ & $1$ & $1$ & $2$ & $3$ \\
    toothbrush & $1$ & $2$ & $1$ & $1$ & $1$ & $1$ & $3$ & $2$ & $1$ & $1$ & $2$ & $1$ & $1$ & $2$ & $1$ & $1$ & $1$ & $2$ & $1$ & $3$ & $1$ & $1$ & $3$ & $3$ \\
    toys & $2$ & $1$ & $2$ & $1$ & $3$ & $5$ & $1$ & $1$ & $1$ & $3$ & $1$ & $2$ & $2$ & $1$ & $1$ & $4$ & $2$ & $3$ & $1$ & $1$ & $2$ & $3$ & $1$ & $1$ \\
    trash/waste & $1$ & $1$ & $1$ & $1$ & $1$ & $1$ & $1$ & $1$ & $3$ & $1$ & $1$ & $1$ & $1$ & $1$ & $1$ & $1$ & $1$ & $1$ & $1$ & $1$ & $2$ & $1$ & $1$ & $1$ \\
    tv & $1$ & $1$ & $1$ & $2$ & $3$ & $2$ & $1$ & $1$ & $2$ & $2$ & $1$ & $1$ & $1$ & $1$ & $1$ & $1$ & $2$ & $1$ & $1$ & $1$ & $1$ & $2$ & $4$ & $6$ \\
    vegetables & $1$ & $2$ & $2$ & $1$ & $2$ & $1$ & $1$ & $1$ & $3$ & $1$ & $1$ & $1$ & $1$ & $1$ & $1$ & $1$ & $2$ & $2$ & $1$ & $1$ & $1$ & $1$ & $1$ & $1$ \\
    wall & $2$ & $1$ & $1$ & $1$ & $1$ & $1$ & $1$ & $2$ & $1$ & $1$ & $1$ & $1$ & $1$ & $1$ & $1$ & $1$ & $1$ & $1$ & $1$ & $1$ & $1$ & $1$ & $1$ & $1$ \\
    wall clock & $2$ & $1$ & $1$ & $2$ & $1$ & $1$ & $1$ & $1$ & $1$ & $1$ & $1$ & $4$ & $1$ & $1$ & $1$ & $2$ & $2$ & $1$ & $1$ & $1$ & $1$ & $2$ & $1$ & $1$ \\
    wall decoration & $1$ & $2$ & $1$ & $1$ & $1$ & $1$ & $2$ & $1$ & $1$ & $1$ & $1$ & $1$ & $2$ & $2$ & $1$ & $1$ & $1$ & $2$ & $2$ & $0$ & $2$ & $1$ & $1$ & $1$ \\
    wall inside & $1$ & $2$ & $1$ & $1$ & $1$ & $1$ & $2$ & $1$ & $1$ & $1$ & $1$ & $2$ & $2$ & $1$ & $1$ & $1$ & $1$ & $2$ & $1$ & $1$ & $1$ & $1$ & $1$ & $1$ \\
    wardrobe & $1$ & $3$ & $2$ & $1$ & $2$ & $1$ & $1$ & $2$ & $1$ & $1$ & $2$ & $1$ & $1$ & $1$ & $2$ & $2$ & $1$ & $1$ & $2$ & $2$ & $2$ & $2$ & $1$ & $1$ \\
    washing clothes/cleaning & $1$ & $1$ & $1$ & $2$ & $1$ & $1$ & $1$ & $1$ & $1$ & $2$ & $1$ & $1$ & $1$ & $1$ & $3$ & $1$ & $1$ & $1$ & $4$ & $4$ & $1$ & $3$ & $1$ & $1$ \\
    washing detergent & $2$ & $1$ & $1$ & $1$ & $1$ & $2$ & $1$ & $1$ & $1$ & $1$ & $1$ & $2$ & $1$ & $1$ & $1$ & $2$ & $1$ & $1$ & $2$ & $2$ & $1$ & $2$ & $1$ & $1$ \\
    water outlet & $2$ & $1$ & $3$ & $2$ & $1$ & $1$ & $1$ & $2$ & $2$ & $1$ & $1$ & $1$ & $1$ & $1$ & $2$ & $1$ & $1$ & $1$ & $1$ & $1$ & $1$ & $1$ & $1$ & $1$ \\
    \bottomrule
    \end{tabular}
\end{table}

\FloatBarrier
\subsection{Dataset Language Details}
\label{appendix:dataset_details_languages}
\begin{table}[htb!]
\tiny
\centering
\caption{Language support of the datasets considered in this work. More details one the languages are reported in Table~\ref{tab:language_details}.}
\label{tab:dataset_details}
\rowcolors{1}{gray!10}{white}
\begin{tabular}{lllccccccc}
\toprule
\rowcolor{white}
\textbf{Language} & \textbf{Script} & \textbf{MaXM} & \textbf{xGQA} & \textbf{XNLVI} & \textbf{MaRVL} & \textbf{M5-VLOD} & \textbf{M5-VGR} & \textbf{xFlickrCO} & \textbf{XM3600} \\
\midrule
Amharic & Ethiopic & \no & \no & \no & \no & \yes & \yes & \no & \no \\
Arabic & Arabic & \no & \no & \yes & \no & \no & \no & \no & \yes \\
Bengali & Bengali & \no & \yes & \no & \no & \yes & \yes & \no & \yes \\
Berber & Tifinagh & \no & \no & \no & \no & \yes & \yes & \no & \no \\
Chinese & Hanzi & \yes & \yes & \no & \yes & \no & \no & \yes & \yes \\
Croatian & Latin & \no & \no & \no & \no & \no & \no & \no & \yes \\
Czech & Latin & \no & \no & \no & \no & \no & \no & \no & \yes \\
Danish & Latin & \no & \no & \no & \no & \no & \no & \no & \yes \\
Dutch & Latin & \no & \no & \no & \no & \no & \no & \no & \yes \\
English & Latin & \yes & \yes & \yes & \no & \yes & \yes & \yes & \yes \\
Filipino & Latin & \no & \no & \no & \no & \yes & \yes & \no & \yes \\
Finnish & Latin & \no & \no & \no & \no & \no & \no & \no & \yes \\
French & Latin & \yes & \no & \yes & \no & \no & \no & \no & \yes \\
German & Latin & \no & \yes & \no & \no & \yes & \yes & \yes & \yes \\
Greek & Greek & \no & \no & \no & \no & \no & \no & \no & \yes \\
Hausa & Latin & \no & \no & \no & \no & \yes & \yes & \no & \no \\
Hebrew & Hebrew & \yes & \no & \no & \no & \no & \no & \no & \yes \\
Hindi & Devanagari & \yes & \no & \no & \no & \yes & \yes & \no & \yes \\
Hungarian & Latin & \no & \no & \no & \no & \no & \no & \no & \yes \\
Indonesian & Latin & \no & \yes & \no & \yes & \no & \no & \yes & \yes \\
Italian & Latin & \no & \no & \no & \no & \no & \no & \no & \yes \\
Japanese & Japanese & \no & \no & \no & \no & \no & \no & \yes & \yes \\
Korean & Hangul & \no & \yes & \no & \no & \no & \no & \no & \yes \\
Maori & Latin & \no & \no & \no & \no & \no & \no & \no & \yes \\
Norwegian & Latin & \no & \no & \no & \no & \no & \no & \no & \yes \\
Persian & Perso-Arabic & \no & \no & \no & \no & \no & \no & \no & \yes \\
Polish & Latin & \no & \no & \no & \no & \no & \no & \no & \yes \\
Portuguese & Latin & \no & \yes & \no & \no & \no & \no & \no & \yes \\
Quechua & Latin & \no & \no & \no & \no & \no & \no & \no & \yes \\
Romanian & Latin & \yes & \no & \no & \no & \no & \no & \no & \yes \\
Russian & Cyrillic & \no & \yes & \yes & \no & \yes & \yes & \yes & \yes \\
Spanish & Latin & \no & \no & \yes & \no & \no & \no & \yes & \yes \\
Swahili & Latin & \no & \no & \no & \yes & \yes & \yes & \no & \yes \\
Swedish & Latin & \no & \no & \no & \no & \no & \no & \no & \yes \\
Tamil & Tamil & \no & \no & \no & \yes & \no & \no & \no & \no \\
Telugu & Telugu & \no & \no & \no & \no & \no & \no & \no & \yes \\
Thai & Thai & \yes & \no & \no & \no & \yes & \yes & \no & \yes \\
Turkish & Latin & \no & \no & \no & \yes & \no & \no & \yes & \yes \\
Ukrainian & Cyrillic & \no & \no & \no & \no & \no & \no & \no & \yes \\
Vietnamese & Latin & \no & \no & \no & \no & \no & \no & \no & \yes \\
Zulu & Latin & \no & \no & \no & \no & \yes & \yes & \no & \no \\
\midrule
\rowcolor{gray!10}
\multicolumn{2}{l}{\textbf{Unique Languages}} & 
\multicolumn{1}{c}{7} & 
\multicolumn{1}{c}{8} & 
\multicolumn{1}{c}{5} & 
\multicolumn{1}{c}{5} & 
\multicolumn{1}{c}{12} & 
\multicolumn{1}{c}{12} & 
\multicolumn{1}{c}{8} & 
\multicolumn{1}{c}{36} \\

% \rowcolor{gray!10}
% \multicolumn{3}{l}{\textbf{Unique Language Families}} & \multicolumn{1}{c}{2} & 
% \multicolumn{1}{c}{4} & 
% \multicolumn{1}{c}{2} & 
% \multicolumn{1}{c}{5} & 
% \multicolumn{1}{c}{5} & 
% \multicolumn{1}{c}{13} \\

\rowcolor{gray!10}
\multicolumn{2}{l}{\textbf{Unique Scripts}} & 
\multicolumn{1}{c}{4} & 
\multicolumn{1}{c}{5} & 
\multicolumn{1}{c}{3} & 
\multicolumn{1}{c}{3} & 
\multicolumn{1}{c}{7} & 
\multicolumn{1}{c}{7} & 
\multicolumn{1}{c}{4} & 
\multicolumn{1}{c}{12} \\
\bottomrule
\end{tabular}
\end{table}
\newpage
\subsection{Language Details}
\label{appendix:language_details}
\begin{landscape}
\begin{table}[!ht]
\scriptsize
\centering
\caption[Details and statistics of languages comprised in the datasets of this benchmark. The continent and subregion columns refer to the content or subregion where the respective language is mostly spoken. The number of speakers is an estimate of the number of L1 and L2 speakers based on different public sources such as Wikipedia, Ethnologue, and Statista]{Details and statistics of languages comprised in the datasets of this benchmark. The continent and subregion columns refer to the content or subregion where the respective language is mostly spoken. The number of speakers is an estimate of the number of L1 and L2 speakers based on different public sources such as Wikipedia\protect\footnotemark, Ethnologue~\protect\footnotemark, and Statista\protect\footnotemark. The ``Taxonomy'' column indicates the taxonomy class of the language based on~\citet{joshi2020langtax}.}
\label{tab:language_details}
\begin{tabular}{llllllrr}
\toprule
\textbf{Language} & \textbf{ISO 639} & \textbf{Lang. Family} & \textbf{Script} & \textbf{Continent} & \textbf{Subregion} & \textbf{Taxonomy} & \textbf{Speakers $\mathbf{/~10^6}$} \\
\midrule
Arabic & ar & Afro-Asiatic & Arabic & Afrika \& Asia & North Africa \& Middle East & 5 & 630.00 \\
Chinese & zh & Sino-Tibetan & Hanzi & Asia & Northeastern Asia & 5 & 1330.00 \\
English & en & Indo-European & Latin & America & North America & 5 & 1457.00 \\
French & fr & Indo-European & Latin & Europe & Western Europe & 5 & 310.00 \\
German & de & Indo-European & Latin & Europe & Western Europe & 5 & 175.00 \\
Japanese & ja & Japonic & Japanese & Asia & Northeastern Asia & 5 & 128.00 \\
Spanish & es & Indo-European & Latin & Europe & Southern Europe & 5 & 600.00 \\ \hline
Croatian & hr & Indo-European & Latin & Europe & Central \& Eastern Europe & 4 & 6.80 \\
Czech & cs & Indo-European & Latin & Europe & Central \& Eastern Europe & 4 & 11.00 \\
Dutch & nl & Indo-European & Latin & Europe & Western Europe & 4 & 30.00 \\
Finnish & fi & Uralic & Latin & Europe & Northern Europe & 4 & 5.80 \\
Hindi & hi & Indo-European & Devanagari & Asia & Central \& South Asia & 4 & 600.00 \\
Hungarian & hu & Uralic & Latin & Europe & Central \& Eastern Europe & 4 & 17.00 \\
Italian & it & Indo-European & Latin & Europe & Southern Europe & 4 & 68.00 \\
Korean & ko & Koreanic & Hangul & Asia & Northeastern Asia & 4 & 82.00 \\
Persian & fa & Indo-European & Perso-Arabic & Asia & Middle East & 4 & 130.00 \\
Polish & pl & Indo-European & Latin & Europe & Central \& Eastern Europe & 4 & 41.00 \\
Portuguese & pt & Indo-European & Latin & Europe \& America & Southern Europe \& South America & 4 & 360.00 \\
Russian & ru & Indo-European & Cyrillic & Asia & Central Asia & 4 & 260.00 \\
Swedish & sv & Indo-European & Latin & Europe & Northern Europe & 4 & 13.00 \\
Turkish & tr & Turkic & Latin & Asia & Middle East & 4 & 90.00 \\
Vietnamese & vi & Austroasiatic & Latin & Asia & Southeastern Asia & 4 & 85.00 \\\hline
Bengali & bn & Indo-European & Bengali & Asia & Central \& South Asia & 3 & 270.00 \\
Danish & da & Indo-European & Latin & Europe & Western Europe & 3 & 6.00 \\
Filipino & fil & Austronesian & Latin & Asia & Southeastern Asia & 3 & 83.00 \\
Greek & el & Indo-European & Greek & Europe & Central \& Eastern Europe & 3 & 13.50 \\
Hebrew & he \& iw & Afro-Asiatic & Hebrew & Asia & Middle East & 3 & 9.00 \\
Indonesian & id & Austronesian & Latin & Asia & Southeastern Asia & 3 & 300.00 \\
Romanian & ro & Indo-European & Latin & Europe & Central \& Eastern Europe & 3 & 28.50 \\
Tamil & ta & Dravidian & Tamil & Asia & Central \& South Asia & 3 & 86.00 \\
Thai & th & Kra-Dai & Thai & Asia & Southeastern Asia & 3 & 80.00 \\
Ukrainian & uk & Indo-European & Cyrillic & Europe & Central \& Eastern Europe & 3 & 32.80 \\\hline
Amharic & am & Afro-Asiatic & Ethiopic & Africa & Eastern Africa & 2 & 57.00 \\
Hausa & ha & Afro-Asiatic & Latin & Africa & Western Africa & 2 & 79.00 \\
Swahili & sw & Niger-Congo & Latin & Africa & Eastern Africa & 2 & 73.00 \\
Zulu & zu & Niger-Congo & Latin & Africa & Southern Africa & 2 & 28.00 \\ \hline
Maori & mi & Austronesian & Latin & Australia \& Oceania & Australia \& Oceania & 1 & 0.19 \\
Norwegian & no & Indo-European & Latin & Europe & Northern Europe & 1 & 4.32 \\
Quechua & quz & Quechuan & Latin & America & South America & 1 & 9.00 \\
Telugu & te & Dravidian & Telugu & Asia & Central \& South Asia & 1 & 96.00 \\ \hline
Berber & ber & Afro-Asiatic & Tifinagh & Africa & Northern Africa & 0 & 26.20 \\
\bottomrule
\end{tabular}
\end{table}
\footnotetext[3]{\url{https://en.wikipedia.org/wiki/List_of_languages_by_total_number_of_speakers}}
\footnotetext[4]{\url{https://www.ethnologue.com/}}
\footnotetext[5]{\url{https://www.statista.com/statistics/266808/the-most-spoken-languages-worldwide/}}
\end{landscape}
\newpage
\onecolumn
\section{Model Details}
\label{appendix:model_details}
\begin{table*}[ht!]
    \scriptsize
    \centering
    \caption{Architectural details of the LMMs evaluated in this study. The columns LM, VM, and ML are ``\textbf{L}anguage \textbf{M}odel'', ``\textbf{V}ision \textbf{M}odel'', and ``\textbf{M}apping \textbf{M}odules'', respectively, and show the number of parameters of the particular module. ``|Total|'' shows all parameters of the model. Note that we report friedly names of the models which are enriched with hyperlinks pointing to the respective Huggingface repositories (when viewed digitally). For Gemini Pro Vision and GPT-4 Vision, we used the \texttt{gemini-1.0-pro-vision} and \texttt{gpt-4-1106-vision-preview} variants, respectively.}
    \label{tab:models_architecture_overview}
    \begin{tabular}{@{}llll|rrrr@{}}
    \toprule
    \textbf{Model} & \textbf{LM} & \textbf{VM} & \textbf{MM} & \textbf{|Total|} & \textbf{|LM|} & \textbf{|VM|} & \textbf{|MM|} \\
    \midrule
    \href{https://huggingface.co/openbmb/MiniCPM-V}{MiniCPM-V}~[\citenum{hu2024minicpm, yu2023rlhf}] & \href{https://huggingface.co/openbmb/MiniCPM-2B-dpo-bf16}{MiniCPM-2B} & \href{https://huggingface.co/google/siglip-so400m-patch14-384}{SigLIP 400M} & MLP & $3.43\mathrm{B}$ & $3.01\mathrm{B}$ & $397.75\mathrm{M}$ & $29.51\mathrm{M}$ \\
    \href{https://huggingface.co/Gregor/mblip-mt0-xl}{mBliP mT0}~[\citenum{geigle2023mblip}] & \href{https://huggingface.co/google/flan-t5-xl}{Flan-T5-XL} & \href{https://huggingface.co/QuanSun/EVA-CLIP/blob/main/EVA01_g_psz14.pt}{EVA01 CLIP-ViT-g} & QFormer & $4.84\mathrm{B}$ & $3.74\mathrm{B}$ & $985.95\mathrm{M}$ & $106.71\mathrm{M}$ \\
    \href{https://huggingface.co/01-ai/Yi-VL-6B}{Yi-VL 6B}~[\citenum{ai012024yi}] & \href{https://huggingface.co/01-ai/Yi-6B-Chat}{Yi-6B-Chat} & \href{https://huggingface.co/laion/CLIP-ViT-H-14-laion2B-s32B-b79K}{CLIP-ViT-H-14} & MLP & $6.71\mathrm{B}$ & $5.80\mathrm{B}$ & $631.75\mathrm{M}$ & $22.04\mathrm{M}$ \\
    \href{https://huggingface.co/liuhaotian/llava-v1.6-vicuna-7b}{LLaVA 1.6 7B}~[\citenum{liu2023improvedllava}] & \href{https://huggingface.co/lmsys/vicuna-7b-v1.5}{Vicuna-7B-v1.5} & \href{https://huggingface.co/openai/clip-vit-large-patch14}{CLIP-ViT-L} & MLP & $6.76\mathrm{B}$ & $6.61\mathrm{B}$ & $303.51\mathrm{M}$ & $20.98\mathrm{M}$ \\
    \href{https://huggingface.co/llava-hf/llava-1.5-7b-hf}{LLaVA 1.5 7B}~[\citenum{liu2023llava}] &  \href{https://huggingface.co/lmsys/vicuna-7b-v1.5}{Vicuna-7B-v1.5} & \href{https://huggingface.co/openai/clip-vit-large-patch14}{CLIP-ViT-L} & MLP & $7.06\mathrm{B}$ & $6.74\mathrm{B}$ & $303.51\mathrm{M}$ & $20.98\mathrm{M}$ \\
    % \href{https://huggingface.co/liuhaotian/llava-v1.6-mistral-7b}{LLaVA 1.6 Mistral}~[\citenum{liu2023improvedllava}] &  \href{https://huggingface.co/mistralai/Mistral-7B-Instruct-v0.2}{Mistral 7B v0.2} & \href{https://huggingface.co/openai/clip-vit-large-patch14}{CLIP-ViT-L} & MLP & $7.26\mathrm{B}$ & $7.11\mathrm{B}$ & $303.51\mathrm{M}$ & $20.98\mathrm{M}$ \\
    \href{https://huggingface.co/llava-hf/bakLlava-v1-hf}{BakLLaVA}~[\citenum{liu2023llava}] & \href{https://huggingface.co/mistralai/Mistral-7B-v0.1}{Mistral 7B v0.1} & \href{https://huggingface.co/openai/clip-vit-large-patch14}{CLIP-ViT-L} & MLP & $7.57\mathrm{B}$ & $7.24\mathrm{B}$ & $303.51\mathrm{M}$ & $20.98\mathrm{M}$ \\
    \href{https://huggingface.co/Gregor/mblip-bloomz-7b}{mBliP BloomZ}~[\citenum{geigle2023mblip}] & \href{https://huggingface.co/bigscience/bloomz-7b1}{BloomZ 7B} & \href{https://huggingface.co/QuanSun/EVA-CLIP/blob/main/EVA01_g_psz14.pt}{EVA01 CLIP-ViT-g} & QFormer & $8.16\mathrm{B}$ & $7.07\mathrm{B}$ & $985.95\mathrm{M}$ & $108.29\mathrm{M}$ \\
    \href{https://huggingface.co/Qwen/Qwen-VL-Chat}{Qwen-VL}~[\citenum{bai2023qwenvl}] & \href{https://huggingface.co/Qwen/Qwen-7B-Chat}{Qwen-7B} & \href{https://huggingface.co/laion/CLIP-ViT-bigG-14-laion2B-39B-b160k}{CLIP-VIT-bigG} & CrossAttn & $9.66\mathrm{B}$ & $7.10\mathrm{B}$ & $1.94\mathrm{B}$ & $80.00\mathrm{M}$ \\
    \hline
    \href{https://huggingface.co/openbmb/OmniLMM-12B}{OmniLMM 12B}~[\citenum{yu2023rlhf}] & \href{https://huggingface.co/HuggingFaceH4/zephyr-7b-beta}{Zephyr 7B $\beta$} & \href{https://huggingface.co/QuanSun/EVA-CLIP/blob/main/EVA02_CLIP_E_psz14_s4B.pt}{EVA02 CLIP ViT-E} & MLP & $11.61\mathrm{B}$ & $7.24\mathrm{B}$ & $4.28\mathrm{B}$ & $93.36\mathrm{M}$ \\
    \href{https://huggingface.co/liuhaotian/llava-v1.6-vicuna-13b}{LLaVA 1.6 13B}~[\citenum{liu2023improvedllava}] &  \href{lmsys/vicuna-13b-v1.5}{Vicuna-13B-v1.5} & \href{https://huggingface.co/openai/clip-vit-large-patch14}{CLIP-ViT-L} & MLP & $13.05\mathrm{B}$ & $12.85\mathrm{B}$ & $303.51\mathrm{M}$ & $31.47\mathrm{M}$ \\
    \href{https://huggingface.co/llava-hf/llava-1.5-13b-hf}{LLaVA 1.5 13B}~[\citenum{liu2023llava}] &  \href{lmsys/vicuna-13b-v1.5}{Vicuna-13B-v1.5} & \href{https://huggingface.co/openai/clip-vit-large-patch14}{CLIP-ViT-L} & MLP & $13.35\mathrm{B}$ & $13.02\mathrm{B}$ & $303.51\mathrm{M}$ & $31.47\mathrm{M}$ \\
    \href{https://huggingface.co/THUDM/cogvlm-chat-hf}{CogVLM}~[\citenum{wang2023cogvlm}] & \href{https://huggingface.co/lmsys/vicuna-7b-v1.5}{Vicuna-7B-v1.5} & \href{https://huggingface.co/QuanSun/EVA-CLIP/blob/main/EVA02_CLIP_E_psz14_s4B.pt}{EVA02 CLIP ViT-E} & CrossAttn & $17.64\mathrm{B}$ & $6.74\mathrm{B}$ & $4.28\mathrm{B}$ & $6.62\mathrm{B}$ \\
    \href{https://huggingface.co/OpenGVLab/InternVL-Chat-Chinese-V1-1}{InternVL V1.1}
    ~[\citenum{chen2023internvl}] &  \href{https://huggingface.co/meta-llama/Llama-2-13b}{Llama-2-13B} & \href{https://huggingface.co/OpenGVLab/InternViT-6B-224px}{InternViT 6B} & MLP & $19.11\mathrm{B}$ & $13.12\mathrm{B}$ & $5.91\mathrm{B}$ & $91.79\mathrm{M}$ \\
    \hline
    \href{https://huggingface.co/liuhaotian/llava-v1.6-34b}{LLaVA 1.6 34B}~[\citenum{liu2023improvedllava}] & \href{https://huggingface.co/NousResearch/Nous-Hermes-2-Yi-34B}{Nous-Hermes-2-Yi-34B} & \href{https://huggingface.co/openai/clip-vit-large-patch14}{CLIP-ViT-L} & MLP & $34.45\mathrm{B}$ & $33.93\mathrm{B}$ & $303.51\mathrm{M}$ & $58.73\mathrm{M}$ \\
    \href{https://huggingface.co/01-ai/Yi-VL-34B}{Yi-VL 34B}~[\citenum{ai012024yi}] & \href{https://huggingface.co/01-ai/Yi-34B-Chat}{Yi-34B-Chat} & \href{https://huggingface.co/laion/CLIP-ViT-H-14-laion2B-s32B-b79K}{CLIP-ViT-H} & MLP & $35.08\mathrm{B}$ & $33.93\mathrm{B}$ & $631.75\mathrm{M}$ & $60.60\mathrm{M}$ \\
    \href{https://huggingface.co/OpenGVLab/InternVL-Chat-Chinese-V1-2-Plus}{InternVL V1.2+}~[\citenum{chen2023internvl}] & \href{https://huggingface.co/NousResearch/Nous-Hermes-2-Yi-34B}{Nous-Hermes-2-Yi-34B} & \href{https://huggingface.co/OpenGVLab/InternViT-6B-448px-V1-2}{InternViT-6B V1-2} & MLP & $40.07\mathrm{B}$ & $34.39\mathrm{B}$ & $5.54\mathrm{B}$ & $143.17\mathrm{M}$ \\
    \hline
    Gemini Pro Vision~[\citenum{anil2023gemini}] &  ? & ? & ? & ? & ? & ? & ? \\
    GPT-4 Vision~[\citenum{openai2023gpt4v}] &  ? & ? & ? & ? & ? & ? & ? \\
    \bottomrule
    \end{tabular}
\end{table*}
\newpage
\onecolumn
\section{Results Details}
\label{appendix:result_details}
\subsection{General Results}
\label{appendix:result_details_overview}
\subsubsection{xGQA}
\label{appendix:result_details_xgqa}
\begin{figure*}[ht!]
    \centering
    \includegraphics[width=1.\linewidth]{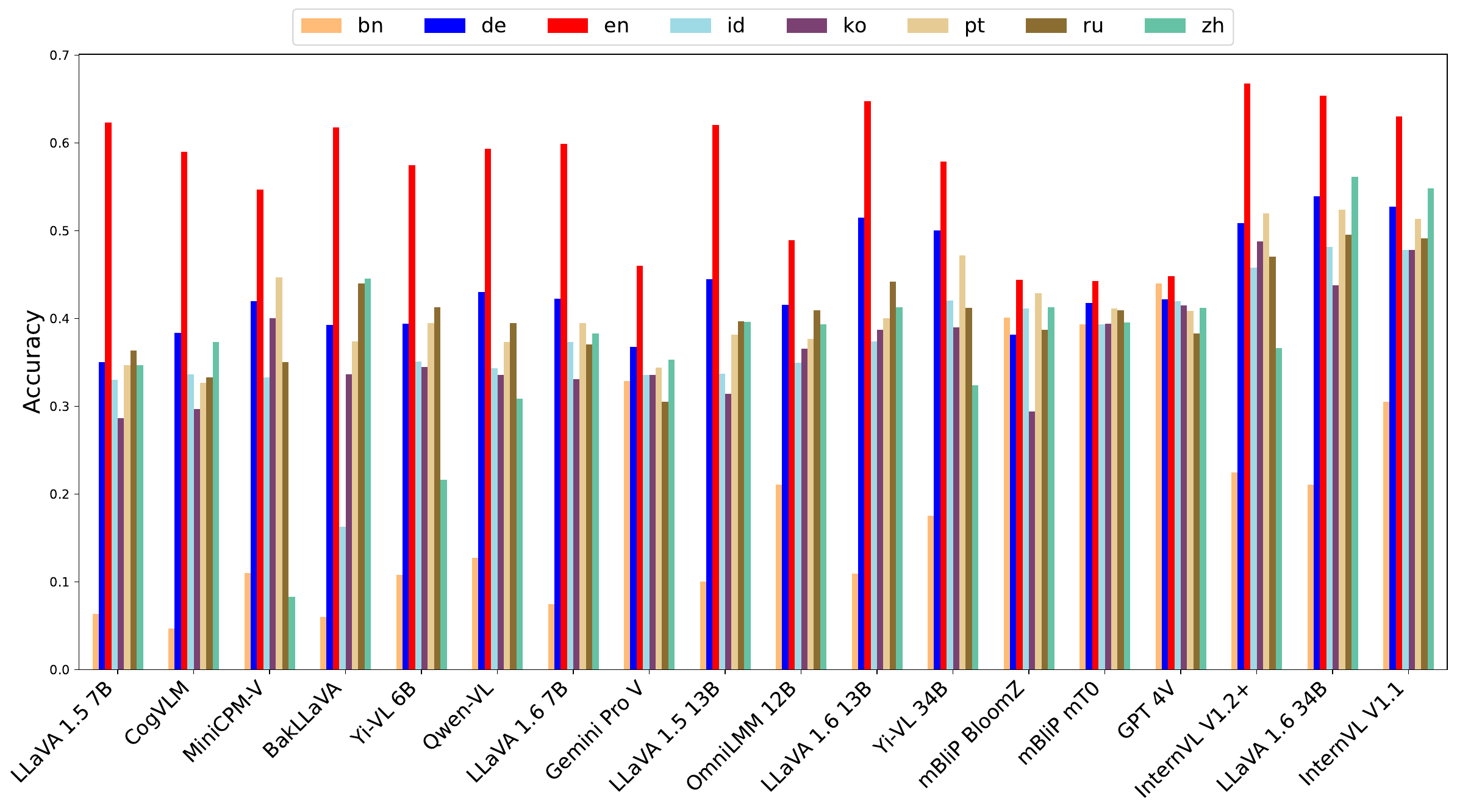}
    \caption{A bar plot showing the average accuracy per language and model on the xGQA dataset. The models on the x-Axis are ordered by their average score across all languages so that the best performing model is on the right and the worst is on the left.}
    \label{fig:result_details_xgqa_plot}
\end{figure*}
\begin{table}[ht!]
    \centering
    \small
    \caption{The average accuracy per language and model on the xGQA dataset. The column ``NEA'' stands for the average of Non-English languages.}
    \label{tab:result_details_xgqa_tab}
    \begin{tabular}{@{}lcccccccccc@{}}
    \toprule
    \multicolumn{1}{c}{Model} & \multicolumn{9}{c}{Language} \\
     & bn & de & en & id & ko & pt & ru & zh & NEA \\
    \midrule
    LLaVA 1.5 7B & $0.06$ & $0.35$ & $0.62$ & $0.33$ & $0.29$ & $0.35$ & $0.36$ & $0.35$ & $0.30$ \\
    CogVLM & $0.05$ & $0.38$ & $0.59$ & $0.34$ & $0.30$ & $0.33$ & $0.33$ & $0.37$ & $0.30$ \\
    MiniCPM-V & $0.11$ & $0.42$ & $0.55$ & $0.33$ & $0.40$ & $0.45$ & $0.35$ & $0.08$ & $0.31$ \\
    BakLLaVA & $0.06$ & $0.39$ & $0.62$ & $0.16$ & $0.34$ & $0.37$ & $0.44$ & $0.45$ & $0.32$ \\
    Yi-VL 6B & $0.11$ & $0.39$ & $0.57$ & $0.35$ & $0.34$ & $0.39$ & $0.41$ & $0.22$ & $0.32$ \\
    Qwen-VL & $0.13$ & $0.43$ & $0.59$ & $0.34$ & $0.34$ & $0.37$ & $0.39$ & $0.31$ & $0.33$ \\
    LLaVA 1.6 7B & $0.07$ & $0.42$ & $0.60$ & $0.37$ & $0.33$ & $0.39$ & $0.37$ & $0.38$ & $0.34$ \\
    Gemini Pro V & $0.33$ & $0.37$ & $0.46$ & $0.34$ & $0.34$ & $0.34$ & $0.31$ & $0.35$ & $0.34$ \\
    LLaVA 1.5 13B & $0.10$ & $0.44$ & $0.62$ & $0.34$ & $0.31$ & $0.38$ & $0.40$ & $0.40$ & $0.34$ \\
    OmniLMM 12B & $0.21$ & $0.42$ & $0.49$ & $0.35$ & $0.37$ & $0.38$ & $0.41$ & $0.39$ & $0.36$ \\
    LLaVA 1.6 13B & $0.11$ & $0.52$ & $0.65$ & $0.37$ & $0.39$ & $0.40$ & $0.44$ & $0.41$ & $0.38$ \\
    Yi-VL 34B & $0.18$ & $0.50$ & $0.58$ & $0.42$ & $0.39$ & $0.47$ & $0.41$ & $0.32$ & $0.38$ \\
    mBliP BloomZ & $0.40$ & $0.38$ & $0.44$ & $0.41$ & $0.29$ & $0.43$ & $0.39$ & $0.41$ & $0.39$ \\
    mBliP mT0 & $0.39$ & $0.42$ & $0.44$ & $0.39$ & $0.39$ & $0.41$ & $0.41$ & $0.40$ & $0.40$ \\
    GPT 4V & $0.44$ & $0.42$ & $0.45$ & $0.42$ & $0.41$ & $0.41$ & $0.38$ & $0.41$ & $0.41$ \\
    InternVL V1.2+ & $0.22$ & $0.51$ & $0.67$ & $0.46$ & $0.49$ & $0.52$ & $0.47$ & $0.37$ & $0.43$ \\
    LLaVA 1.6 34B & $0.21$ & $0.54$ & $0.65$ & $0.48$ & $0.44$ & $0.52$ & $0.50$ & $0.56$ & $0.46$ \\
    InternVL V1.1 & $0.31$ & $0.53$ & $0.63$ & $0.48$ & $0.48$ & $0.51$ & $0.49$ & $0.55$ & $0.48$ \\ \midrule
    Average & $0.19$ & $0.43$ & $0.57$ & $0.37$ & $0.37$ & $0.41$ & $0.40$ & $0.37$ & $0.37$ \\
    \bottomrule
    \end{tabular}
\end{table}

\newpage
\subsubsection{MaXM}
\label{appendix:result_details_maxm}
\begin{figure*}[ht!]
    \centering
    \includegraphics[width=1.\linewidth]{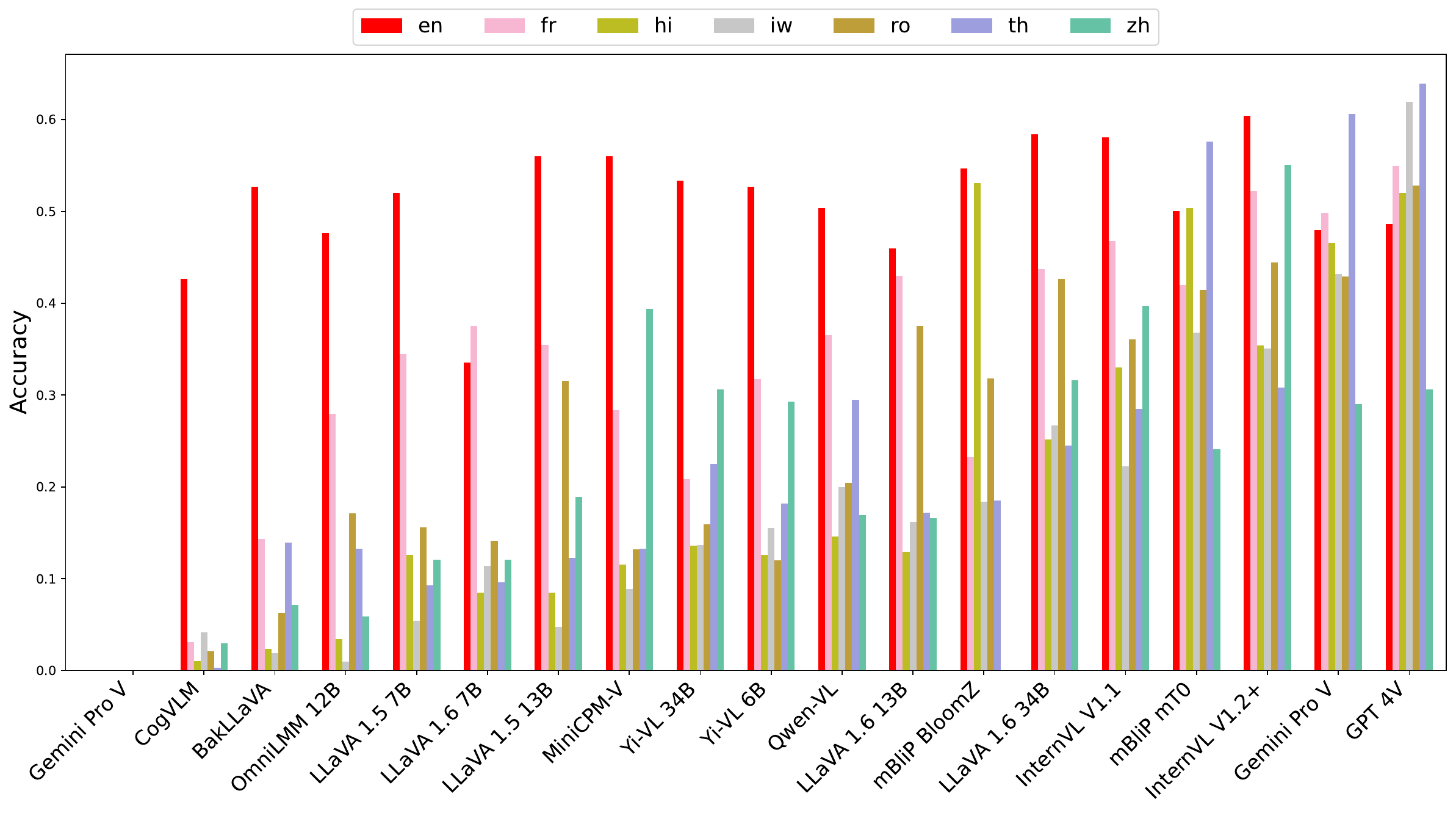}
    \caption{A bar plot showing the average accuracy per language and model on the MaXM dataset. The models on the x-Axis are ordered by their average score across all languages so that the best performing model is on the right and the worst is on the left.}
    \label{fig:result_details_maxm_plot}
\end{figure*}
\begin{table}[ht!]
    \centering
    \caption{The average accuracy per language and model on the MaXM dataset. The column ``NEA'' stands for the average of Non-English languages.}
    \label{tab:result_details_maxm_tab}
    \begin{tabular}{@{}lcccccccc@{}}
    \toprule
    \multicolumn{1}{c}{Model} & \multicolumn{8}{c}{Language} \\
     & en & fr & hi & iw & ro & th & zh & NEA \\
    \midrule
    CogVLM & $0.43$ & $0.03$ & $0.01$ & $0.04$ & $0.02$ & $0.00$ & $0.03$ & $0.02$ \\
    BakLLaVA & $0.53$ & $0.14$ & $0.02$ & $0.02$ & $0.06$ & $0.14$ & $0.07$ & $0.08$ \\
    OmniLMM 12B & $0.48$ & $0.28$ & $0.03$ & $0.01$ & $0.17$ & $0.13$ & $0.06$ & $0.11$ \\
    LLaVA 1.5 7B & $0.52$ & $0.34$ & $0.13$ & $0.05$ & $0.16$ & $0.09$ & $0.12$ & $0.15$ \\
    LLaVA 1.6 7B & $0.34$ & $0.38$ & $0.09$ & $0.11$ & $0.14$ & $0.10$ & $0.12$ & $0.16$ \\
    LLaVA 1.5 13B & $0.56$ & $0.35$ & $0.09$ & $0.05$ & $0.32$ & $0.12$ & $0.19$ & $0.19$ \\
    MiniCPM-V & $0.56$ & $0.28$ & $0.12$ & $0.09$ & $0.13$ & $0.13$ & $0.39$ & $0.19$ \\
    Yi-VL 34B & $0.53$ & $0.21$ & $0.14$ & $0.14$ & $0.16$ & $0.23$ & $0.31$ & $0.20$ \\
    Yi-VL 6B & $0.53$ & $0.32$ & $0.13$ & $0.16$ & $0.12$ & $0.18$ & $0.29$ & $0.20$ \\
    Qwen-VL & $0.50$ & $0.37$ & $0.15$ & $0.20$ & $0.20$ & $0.29$ & $0.17$ & $0.23$ \\
    LLaVA 1.6 13B & $0.46$ & $0.43$ & $0.13$ & $0.16$ & $0.38$ & $0.17$ & $0.17$ & $0.24$ \\
    mBliP BloomZ & $0.55$ & $0.23$ & $0.53$ & $0.18$ & $0.32$ & $0.19$ & $0.42$ & $0.31$ \\
    LLaVA 1.6 34B & $0.58$ & $0.44$ & $0.25$ & $0.27$ & $0.43$ & $0.25$ & $0.32$ & $0.32$ \\
    InternVL V1.1 & $0.58$ & $0.47$ & $0.33$ & $0.22$ & $0.36$ & $0.28$ & $0.40$ & $0.34$ \\
    mBliP mT0 & $0.50$ & $0.42$ & $0.50$ & $0.37$ & $0.41$ & $0.58$ & $0.24$ & $0.42$ \\
    InternVL V1.2+ & $0.60$ & $0.52$ & $0.35$ & $0.35$ & $0.44$ & $0.31$ & $0.55$ & $0.42$ \\
    Gemini Pro V & $0.48$ & $0.50$ & $0.47$ & $0.43$ & $0.43$ & $0.61$ & $0.29$ & $0.45$ \\
    GPT 4V & $0.49$ & $0.55$ & $0.52$ & $0.62$ & $0.53$ & $0.64$ & $0.31$ & $0.53$ \\ \midrule
    Average & 0.51 & 0.35 & 0.22 & 0.19 & 0.27 & 0.25 & 0.24 & 0.25 \\
    \bottomrule
    \end{tabular}
\end{table}

\newpage
\subsubsection{XVNLI}
\label{appendix:result_details_xvnli}
\begin{figure*}[ht!]
    \centering
    \includegraphics[width=1.\linewidth]{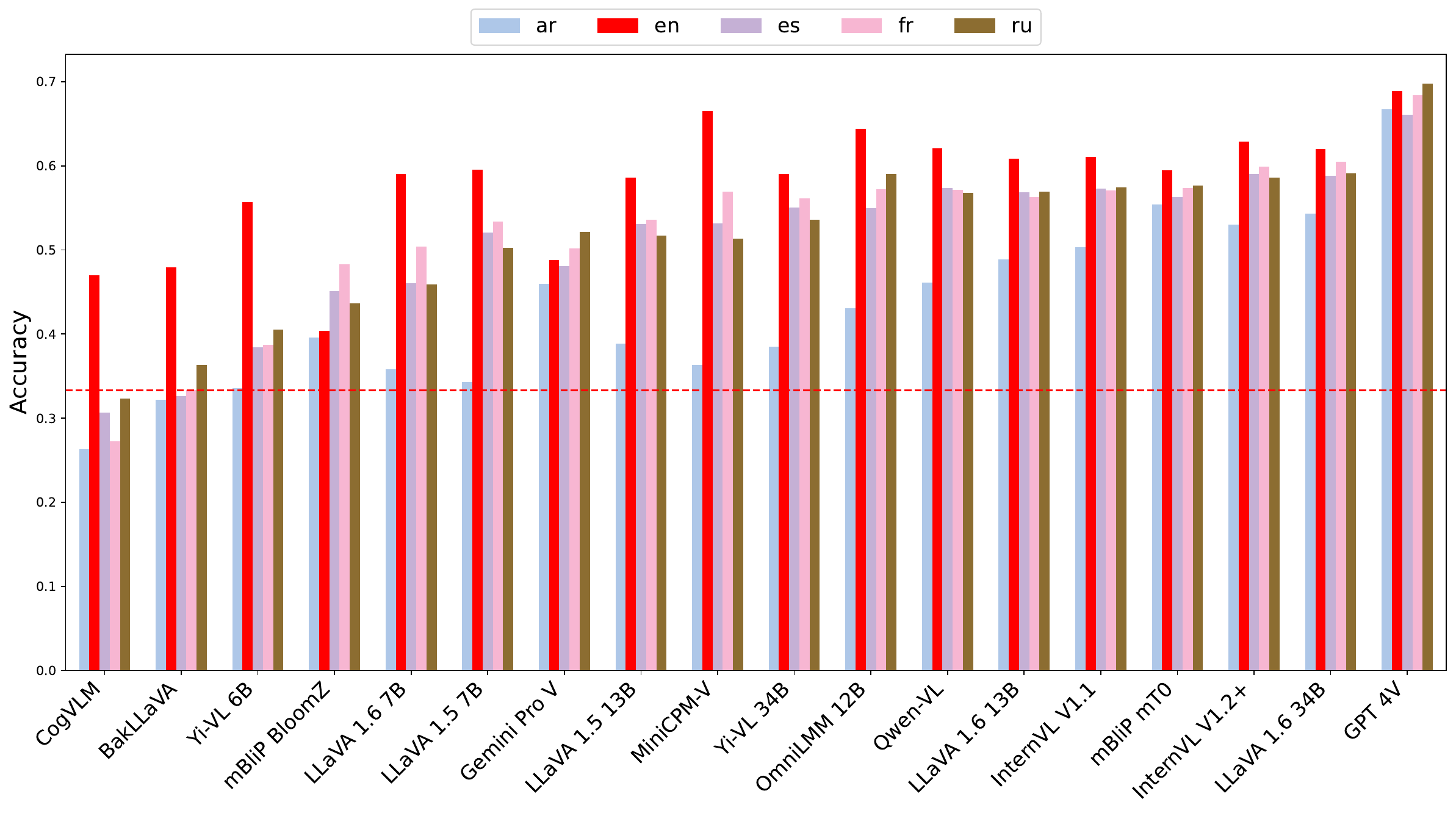}
    \caption{A bar plot showing the average accuracy per language and model on the XVNLI dataset. The models on the x-Axis are ordered by their average score across all languages so that the best performing model is on the right and the worst is on the left.}
    \label{fig:result_details_xvnli_plot}
\end{figure*}
\begin{table}[ht!]
    \centering
    \caption{The average accuracy per language and model on the XVNLI dataset. The column ``NEA'' stands for the average of Non-English languages.}
    \label{tab:result_details_xvnli_tab}
    \begin{tabular}{@{}lccccccc@{}}
    \toprule
    \multicolumn{1}{c}{Model} & \multicolumn{6}{c}{Language} \\
     & ar & en & es & fr & ru & NEA \\
    \midrule
    CogVLM & $0.26$ & $0.47$ & $0.31$ & $0.27$ & $0.32$ & $0.29$ \\
    BakLLaVA & $0.32$ & $0.48$ & $0.33$ & $0.33$ & $0.36$ & $0.34$ \\
    Yi-VL 6B & $0.34$ & $0.56$ & $0.38$ & $0.39$ & $0.41$ & $0.38$ \\
    mBliP BloomZ & $0.40$ & $0.40$ & $0.45$ & $0.48$ & $0.44$ & $0.44$ \\
    LLaVA 1.6 7B & $0.36$ & $0.59$ & $0.46$ & $0.50$ & $0.46$ & $0.45$ \\
    LLaVA 1.5 7B & $0.34$ & $0.60$ & $0.52$ & $0.53$ & $0.50$ & $0.47$ \\
    Gemini Pro V & $0.46$ & $0.49$ & $0.48$ & $0.50$ & $0.52$ & $0.49$ \\
    LLaVA 1.5 13B & $0.39$ & $0.59$ & $0.53$ & $0.54$ & $0.52$ & $0.49$ \\
    MiniCPM-V & $0.36$ & $0.66$ & $0.53$ & $0.57$ & $0.51$ & $0.49$ \\
    Yi-VL 34B & $0.39$ & $0.59$ & $0.55$ & $0.56$ & $0.54$ & $0.51$ \\
    OmniLMM 12B & $0.43$ & $0.64$ & $0.55$ & $0.57$ & $0.59$ & $0.54$ \\
    Qwen-VL & $0.46$ & $0.62$ & $0.57$ & $0.57$ & $0.57$ & $0.54$ \\
    LLaVA 1.6 13B & $0.49$ & $0.61$ & $0.57$ & $0.56$ & $0.57$ & $0.55$ \\
    InternVL V1.1 & $0.50$ & $0.61$ & $0.57$ & $0.57$ & $0.57$ & $0.56$ \\
    mBliP mT0 & $0.55$ & $0.59$ & $0.56$ & $0.57$ & $0.58$ & $0.57$ \\
    InternVL V1.2+ & $0.53$ & $0.63$ & $0.59$ & $0.60$ & $0.59$ & $0.58$ \\
    LLaVA 1.6 34B & $0.54$ & $0.62$ & $0.59$ & $0.60$ & $0.59$ & $0.58$ \\
    GPT 4V & $0.67$ & $0.69$ & $0.66$ & $0.68$ & $0.70$ & $0.68$ \\ \midrule
    Average & $0.43$ & $0.58$ & $0.51$ & $0.52$ & $0.52$ & $0.50$ \\
    \bottomrule
    \end{tabular}
\end{table}

\newpage
\subsubsection{MaRVL}
\label{appendix:result_details_marvl}
\begin{figure*}[ht!]
    \centering
    \includegraphics[width=1.\linewidth]{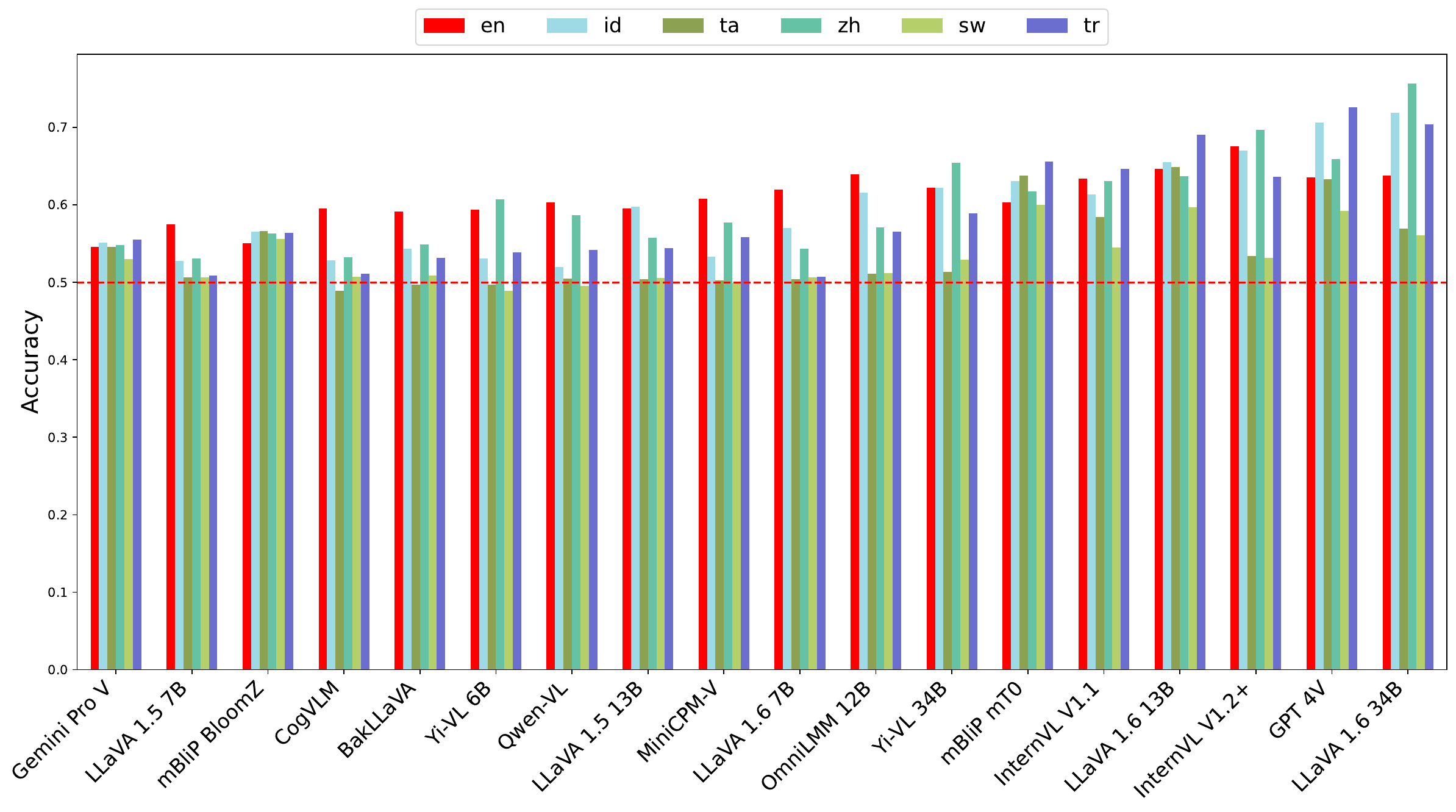}
    \caption{A bar plot showing the average accuracy per language and model on the MaRVL dataset. Note that MaRVL does not contain English data originally and we machine-translated English from the other languages and averaged the results. The models on the x-Axis are ordered by their average score across all languages so that the best performing model is on the right and the worst is on the left.}
    \label{fig:result_details_marvl_plot}
\end{figure*}
\begin{table}[ht!]
    \centering
    \small
    \caption{The average accuracy per language and model on the MaRVL dataset. Note that MaRVL does not contain English data originally and we machine-translated English from the other languages and averaged the results. The column ``NEA'' stands for the average of Non-English languages.}
    \label{tab:result_details_marvl_tab}
    \begin{tabular}{@{}lccccccc@{}}
    \toprule
    \multicolumn{1}{c}{Model} & \multicolumn{7}{c}{Language} \\
     & en & id & sw & ta & tr & zh & NEA \\
    \midrule
    CogVLM & $0.60$ & $0.53$ & $0.51$ & $0.49$ & $0.51$ & $0.53$ & $0.51$ \\
    LLaVA 1.5 7B & $0.57$ & $0.53$ & $0.51$ & $0.51$ & $0.51$ & $0.53$ & $0.52$ \\
    BakLLaVA & $0.59$ & $0.54$ & $0.51$ & $0.50$ & $0.53$ & $0.55$ & $0.53$ \\
    LLaVA 1.6 7B & $0.62$ & $0.57$ & $0.51$ & $0.50$ & $0.51$ & $0.54$ & $0.53$ \\
    Qwen-VL & $0.60$ & $0.52$ & $0.50$ & $0.50$ & $0.54$ & $0.59$ & $0.53$ \\
    Yi-VL 6B & $0.59$ & $0.53$ & $0.49$ & $0.50$ & $0.54$ & $0.61$ & $0.53$ \\
    MiniCPM-V & $0.61$ & $0.53$ & $0.50$ & $0.50$ & $0.56$ & $0.58$ & $0.53$ \\
    LLaVA 1.5 13B & $0.60$ & $0.60$ & $0.51$ & $0.50$ & $0.54$ & $0.56$ & $0.54$ \\
    Gemini Pro V & $0.55$ & $0.55$ & $0.53$ & $0.55$ & $0.56$ & $0.55$ & $0.55$ \\
    OmniLMM 12B & $0.64$ & $0.62$ & $0.51$ & $0.51$ & $0.57$ & $0.57$ & $0.56$ \\
    mBliP BloomZ & $0.55$ & $0.57$ & $0.56$ & $0.57$ & $0.56$ & $0.56$ & $0.56$ \\
    Yi-VL 34B & $0.62$ & $0.62$ & $0.53$ & $0.51$ & $0.59$ & $0.65$ & $0.58$ \\
    InternVL V1.1 & $0.63$ & $0.61$ & $0.54$ & $0.58$ & $0.65$ & $0.63$ & $0.60$ \\
    InternVL V1.2+ & $0.68$ & $0.67$ & $0.53$ & $0.53$ & $0.64$ & $0.70$ & $0.61$ \\
    mBliP mT0 & $0.60$ & $0.63$ & $0.60$ & $0.64$ & $0.66$ & $0.62$ & $0.63$ \\
    LLaVA 1.6 13B & $0.65$ & $0.66$ & $0.60$ & $0.65$ & $0.69$ & $0.64$ & $0.65$ \\
    LLaVA 1.6 34B & $0.64$ & $0.72$ & $0.56$ & $0.57$ & $0.70$ & $0.76$ & $0.66$ \\
    GPT 4V & $0.64$ & $0.71$ & $0.59$ & $0.63$ & $0.73$ & $0.66$ & $0.66$ \\ \midrule
    Average & $0.61$ & $0.60$ & $0.53$ & $0.54$ & $0.59$ & $0.60$ & $0.57$ \\
    \bottomrule
    \end{tabular}
\end{table}

\newpage
\subsubsection{M5-VGR}
\label{appendix:result_details_m5b_vgr}
\begin{figure*}[ht!]
    \centering
    \includegraphics[width=1.\linewidth]{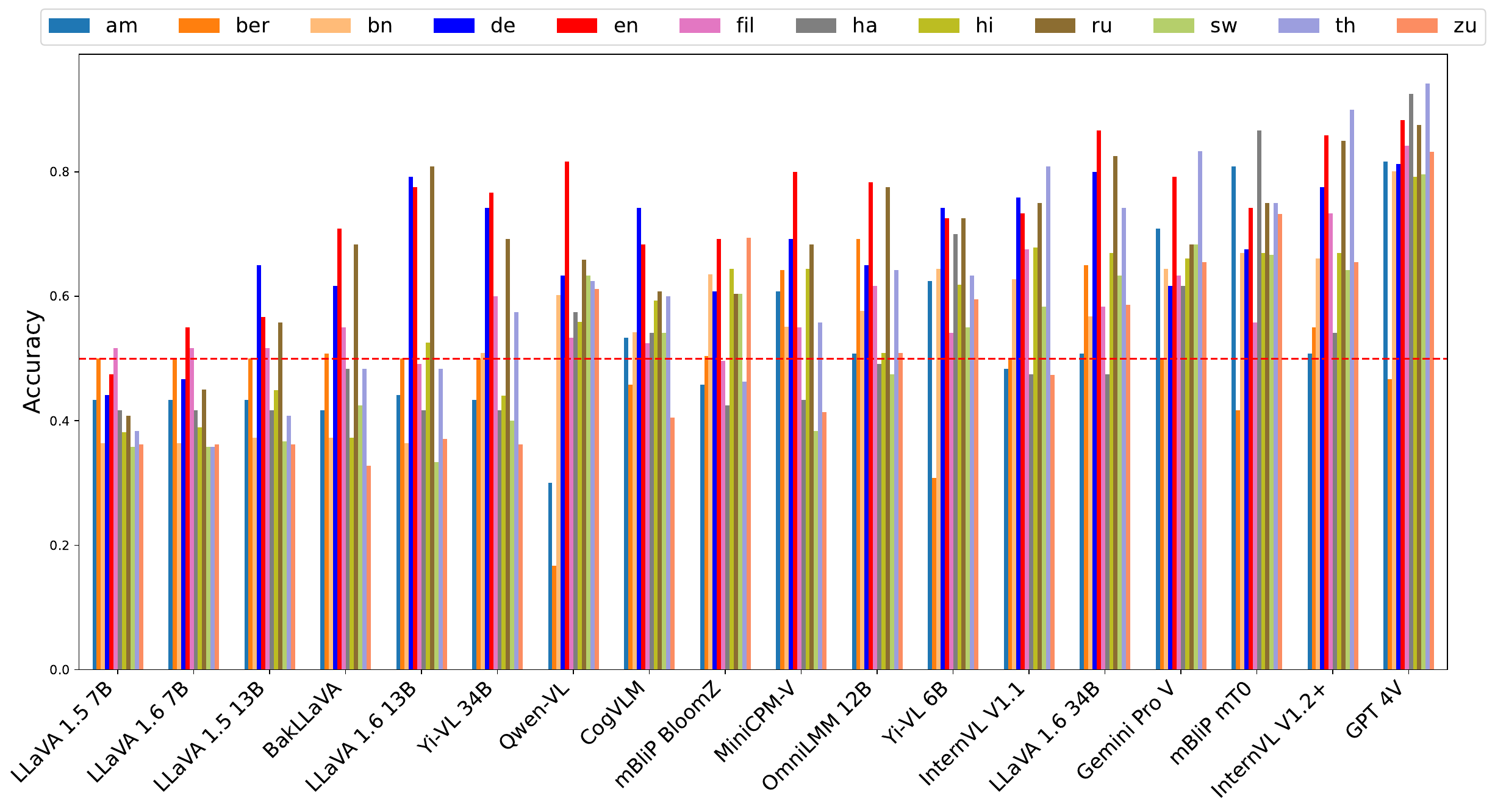}
    \caption{A bar plot showing the average accuracy per language and model on the M5-VGR dataset. The models on the x-Axis are ordered by their average score across all languages so that the best performing model is on the right and the worst is on the left.} 
    \label{fig:result_details_m5b_vgr_plot}
\end{figure*}
\begin{table}[ht!]
    \centering
    \caption{The average accuracy per language and model on the M5-VGR dataset. The column ``NEA'' stands for the average of Non-English languages.}
    \label{tab:result_details_m5b_vgr_tab}
    \addtolength{\tabcolsep}{-0.3em}
    \begin{tabular}{@{}lcccccccccccccc@{}}
    \toprule
    \multicolumn{1}{c}{Model} & \multicolumn{13}{c}{Language} \\
    & am & ber & bn & de & en & fil & ha & hi & ru & sw & th & zu & NEA \\
    \midrule
    LLaVA 1.5 7B & $0.43$ & $0.50$ & $0.36$ & $0.44$ & $0.47$ & $0.52$ & $0.42$ & $0.38$ & $0.41$ & $0.36$ & $0.38$ & $0.36$ & $0.42$ \\
    LLaVA 1.6 7B & $0.43$ & $0.50$ & $0.36$ & $0.47$ & $0.55$ & $0.52$ & $0.42$ & $0.39$ & $0.45$ & $0.36$ & $0.36$ & $0.36$ & $0.42$ \\
    LLaVA 1.5 13B & $0.43$ & $0.50$ & $0.37$ & $0.65$ & $0.57$ & $0.52$ & $0.42$ & $0.45$ & $0.56$ & $0.37$ & $0.41$ & $0.36$ & $0.46$ \\
    BakLLaVA & $0.42$ & $0.51$ & $0.37$ & $0.62$ & $0.71$ & $0.55$ & $0.48$ & $0.37$ & $0.68$ & $0.42$ & $0.48$ & $0.33$ & $0.48$ \\
    LLaVA 1.6 13B & $0.44$ & $0.50$ & $0.36$ & $0.79$ & $0.78$ & $0.49$ & $0.42$ & $0.53$ & $0.81$ & $0.33$ & $0.48$ & $0.37$ & $0.50$ \\
    Yi-VL 34B & $0.43$ & $0.50$ & $0.51$ & $0.74$ & $0.77$ & $0.60$ & $0.42$ & $0.44$ & $0.69$ & $0.40$ & $0.57$ & $0.36$ & $0.52$ \\
    Qwen-VL & $0.30$ & $0.17$ & $0.60$ & $0.63$ & $0.82$ & $0.53$ & $0.57$ & $0.56$ & $0.66$ & $0.63$ & $0.62$ & $0.61$ & $0.54$ \\
    CogVLM & $0.53$ & $0.46$ & $0.54$ & $0.74$ & $0.68$ & $0.53$ & $0.54$ & $0.59$ & $0.61$ & $0.54$ & $0.60$ & $0.41$ & $0.55$ \\
    mBliP BloomZ & $0.46$ & $0.50$ & $0.64$ & $0.61$ & $0.69$ & $0.50$ & $0.42$ & $0.64$ & $0.60$ & $0.60$ & $0.46$ & $0.69$ & $0.56$ \\
    MiniCPM-V & $0.61$ & $0.64$ & $0.55$ & $0.69$ & $0.80$ & $0.55$ & $0.43$ & $0.64$ & $0.68$ & $0.38$ & $0.56$ & $0.41$ & $0.56$ \\
    OmniLMM 12B & $0.51$ & $0.69$ & $0.58$ & $0.65$ & $0.78$ & $0.62$ & $0.49$ & $0.51$ & $0.78$ & $0.47$ & $0.64$ & $0.51$ & $0.59$ \\
    Yi-VL 6B & $0.62$ & $0.31$ & $0.64$ & $0.74$ & $0.72$ & $0.54$ & $0.70$ & $0.62$ & $0.72$ & $0.55$ & $0.63$ & $0.59$ & $0.61$ \\
    InternVL V1.1 & $0.48$ & $0.50$ & $0.63$ & $0.76$ & $0.73$ & $0.68$ & $0.47$ & $0.68$ & $0.75$ & $0.58$ & $0.81$ & $0.47$ & $0.62$ \\
    LLaVA 1.6 34B & $0.51$ & $0.65$ & $0.57$ & $0.80$ & $0.87$ & $0.58$ & $0.47$ & $0.67$ & $0.82$ & $0.63$ & $0.74$ & $0.59$ & $0.64$ \\
    Gemini Pro V & $0.71$ & $0.50$ & $0.64$ & $0.62$ & $0.79$ & $0.63$ & $0.62$ & $0.66$ & $0.68$ & $0.68$ & $0.83$ & $0.66$ & $0.66$ \\
    InternVL V1.2+ & $0.51$ & $0.55$ & $0.66$ & $0.78$ & $0.86$ & $0.73$ & $0.54$ & $0.67$ & $0.85$ & $0.64$ & $0.90$ & $0.66$ & $0.68$ \\
    mBliP mT0 & $0.81$ & $0.42$ & $0.67$ & $0.68$ & $0.74$ & $0.56$ & $0.87$ & $0.67$ & $0.75$ & $0.67$ & $0.75$ & $0.73$ & $0.69$ \\
    GPT 4V & $0.82$ & $0.47$ & $0.80$ & $0.81$ & $0.88$ & $0.84$ & $0.93$ & $0.79$ & $0.88$ & $0.80$ & $0.94$ & $0.83$ & $0.81$ \\ \midrule
    Average & $0.53$ & $0.49$ & $0.55$ & $0.68$ & $0.73$ & $0.58$ & $0.53$ & $0.57$ & $0.69$ & $0.52$ & $0.62$ & $0.52$ & $0.57$ \\
    \bottomrule
    \end{tabular}
\end{table}
\newpage
\subsubsection{M5-VLOD}
\begin{figure*}[ht!]
    \centering
    \includegraphics[width=1.\linewidth]{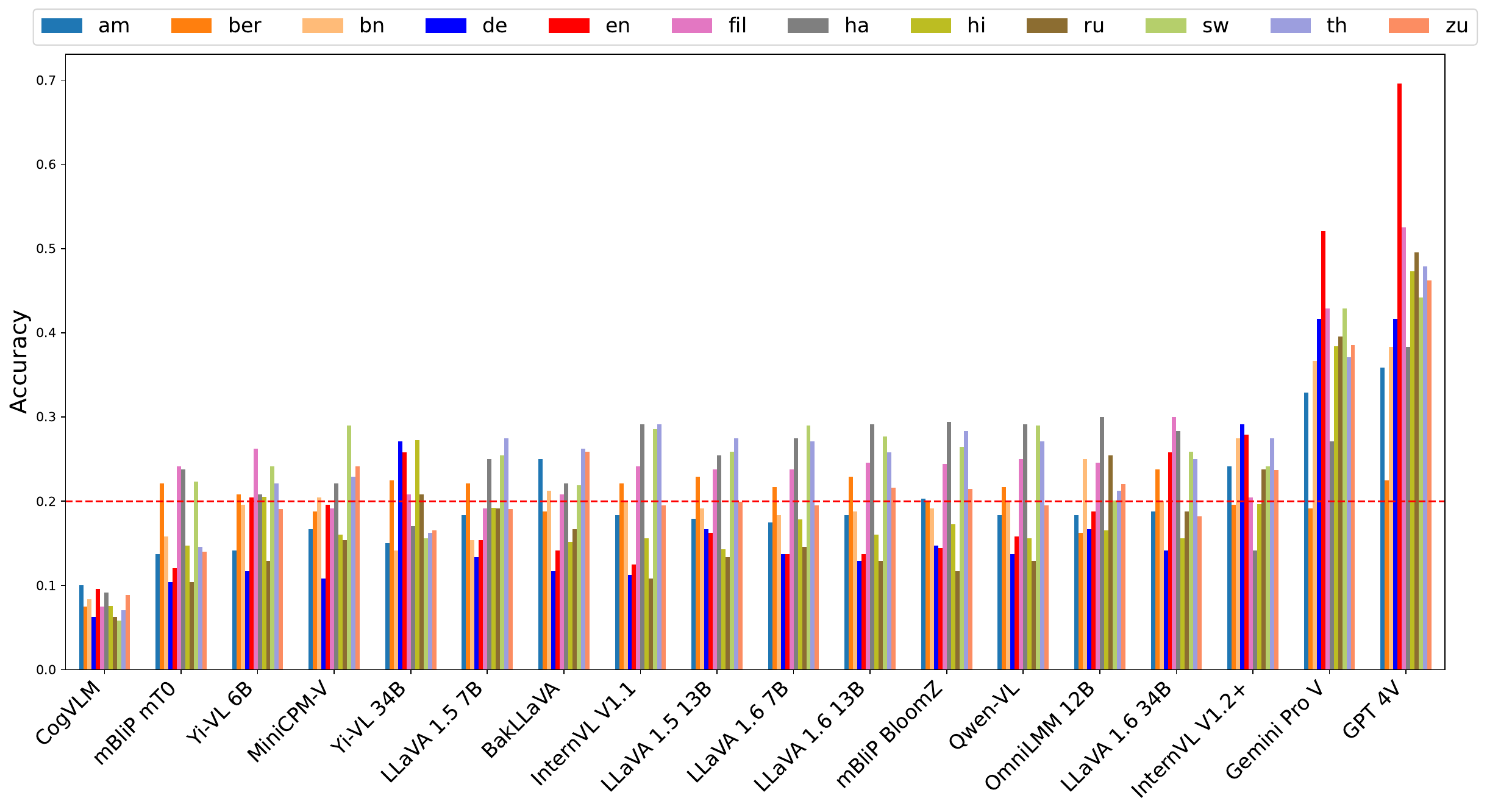}
    \caption{A bar plot showing the average accuracy per language and model on the M5-VLOD dataset. The models on the x-Axis are ordered by their average score across all languages so that the best performing model is on the right and the worst is on the left.}
    \label{fig:result_details_m5b_vlod_plot }
\end{figure*}
\begin{table}[ht!]
    \centering
    \caption{The average accuracy per language and model on the M5-VLOD dataset. The column ``NEA'' stands for the average of Non-English languages.}
    \label{tab:result_details_m5b_vlod_tab}
    \addtolength{\tabcolsep}{-0.3em}
    \begin{tabular}{@{}lcccccccccccccc@{}}
    \toprule
    \multicolumn{1}{c}{Model} & \multicolumn{13}{c}{Language} \\
    & am & ber & bn & de & en & fil & ha & hi & ru & sw & th & zu & NEA \\
    \midrule
    CogVLM & $0.10$ & $0.07$ & $0.08$ & $0.06$ & $0.10$ & $0.07$ & $0.09$ & $0.08$ & $0.06$ & $0.06$ & $0.07$ & $0.09$ & $0.08$ \\
    mBliP mT0 & $0.14$ & $0.22$ & $0.16$ & $0.10$ & $0.12$ & $0.24$ & $0.24$ & $0.15$ & $0.10$ & $0.22$ & $0.15$ & $0.14$ & $0.17$ \\
    Yi-VL 6B & $0.14$ & $0.21$ & $0.20$ & $0.12$ & $0.20$ & $0.26$ & $0.21$ & $0.21$ & $0.13$ & $0.24$ & $0.22$ & $0.19$ & $0.19$ \\
    Yi-VL 34B & $0.15$ & $0.22$ & $0.14$ & $0.27$ & $0.26$ & $0.21$ & $0.17$ & $0.27$ & $0.21$ & $0.16$ & $0.16$ & $0.17$ & $0.19$ \\
    MiniCPM-V & $0.17$ & $0.19$ & $0.20$ & $0.11$ & $0.20$ & $0.19$ & $0.22$ & $0.16$ & $0.15$ & $0.29$ & $0.23$ & $0.24$ & $0.20$ \\
    LLaVA 1.5 7B & $0.18$ & $0.22$ & $0.15$ & $0.13$ & $0.15$ & $0.19$ & $0.25$ & $0.19$ & $0.19$ & $0.25$ & $0.27$ & $0.19$ & $0.20$ \\
    BakLLaVA & $0.25$ & $0.19$ & $0.21$ & $0.12$ & $0.14$ & $0.21$ & $0.22$ & $0.15$ & $0.17$ & $0.22$ & $0.26$ & $0.26$ & $0.20$ \\
    LLaVA 1.5 13B & $0.18$ & $0.23$ & $0.19$ & $0.17$ & $0.16$ & $0.24$ & $0.25$ & $0.14$ & $0.13$ & $0.26$ & $0.28$ & $0.20$ & $0.21$ \\
    InternVL V1.1 & $0.18$ & $0.22$ & $0.20$ & $0.11$ & $0.12$ & $0.24$ & $0.29$ & $0.16$ & $0.11$ & $0.29$ & $0.29$ & $0.19$ & $0.21$ \\
    LLaVA 1.6 7B & $0.17$ & $0.22$ & $0.18$ & $0.14$ & $0.14$ & $0.24$ & $0.27$ & $0.18$ & $0.15$ & $0.29$ & $0.27$ & $0.19$ & $0.21$ \\
    LLaVA 1.6 13B & $0.18$ & $0.23$ & $0.19$ & $0.13$ & $0.14$ & $0.25$ & $0.29$ & $0.16$ & $0.13$ & $0.28$ & $0.26$ & $0.22$ & $0.21$ \\
    Qwen-VL & $0.18$ & $0.22$ & $0.20$ & $0.14$ & $0.16$ & $0.25$ & $0.29$ & $0.16$ & $0.13$ & $0.29$ & $0.27$ & $0.19$ & $0.21$ \\
    mBliP BloomZ & $0.20$ & $0.20$ & $0.19$ & $0.15$ & $0.14$ & $0.24$ & $0.29$ & $0.17$ & $0.12$ & $0.26$ & $0.28$ & $0.21$ & $0.21$ \\
    OmniLMM 12B & $0.18$ & $0.16$ & $0.25$ & $0.17$ & $0.19$ & $0.25$ & $0.30$ & $0.17$ & $0.25$ & $0.20$ & $0.21$ & $0.22$ & $0.21$ \\
    LLaVA 1.6 34B & $0.19$ & $0.24$ & $0.20$ & $0.14$ & $0.26$ & $0.30$ & $0.28$ & $0.16$ & $0.19$ & $0.26$ & $0.25$ & $0.18$ & $0.22$ \\
    InternVL V1.2+ & $0.24$ & $0.20$ & $0.28$ & $0.29$ & $0.28$ & $0.20$ & $0.14$ & $0.20$ & $0.24$ & $0.24$ & $0.28$ & $0.24$ & $0.23$ \\
    Gemini Pro V & $0.33$ & $0.19$ & $0.37$ & $0.42$ & $0.52$ & $0.43$ & $0.27$ & $0.38$ & $0.40$ & $0.43$ & $0.37$ & $0.39$ & $0.36$ \\
    GPT 4V & $0.36$ & $0.22$ & $0.38$ & $0.42$ & $0.70$ & $0.53$ & $0.38$ & $0.47$ & $0.50$ & $0.44$ & $0.48$ & $0.46$ & $0.42$ \\ \midrule
    Average & $0.20$ & $0.20$ & $0.21$ & $0.18$ & $0.22$ & $0.25$ & $0.25$ & $0.20$ & $0.19$ & $0.26$ & $0.26$ & $0.22$ & $0.22$ \\
    \bottomrule
    \end{tabular}
\end{table}
\newpage
\subsubsection{xFlickrCO}
\begin{figure*}[ht!]
    \centering
    \includegraphics[width=1.\linewidth]{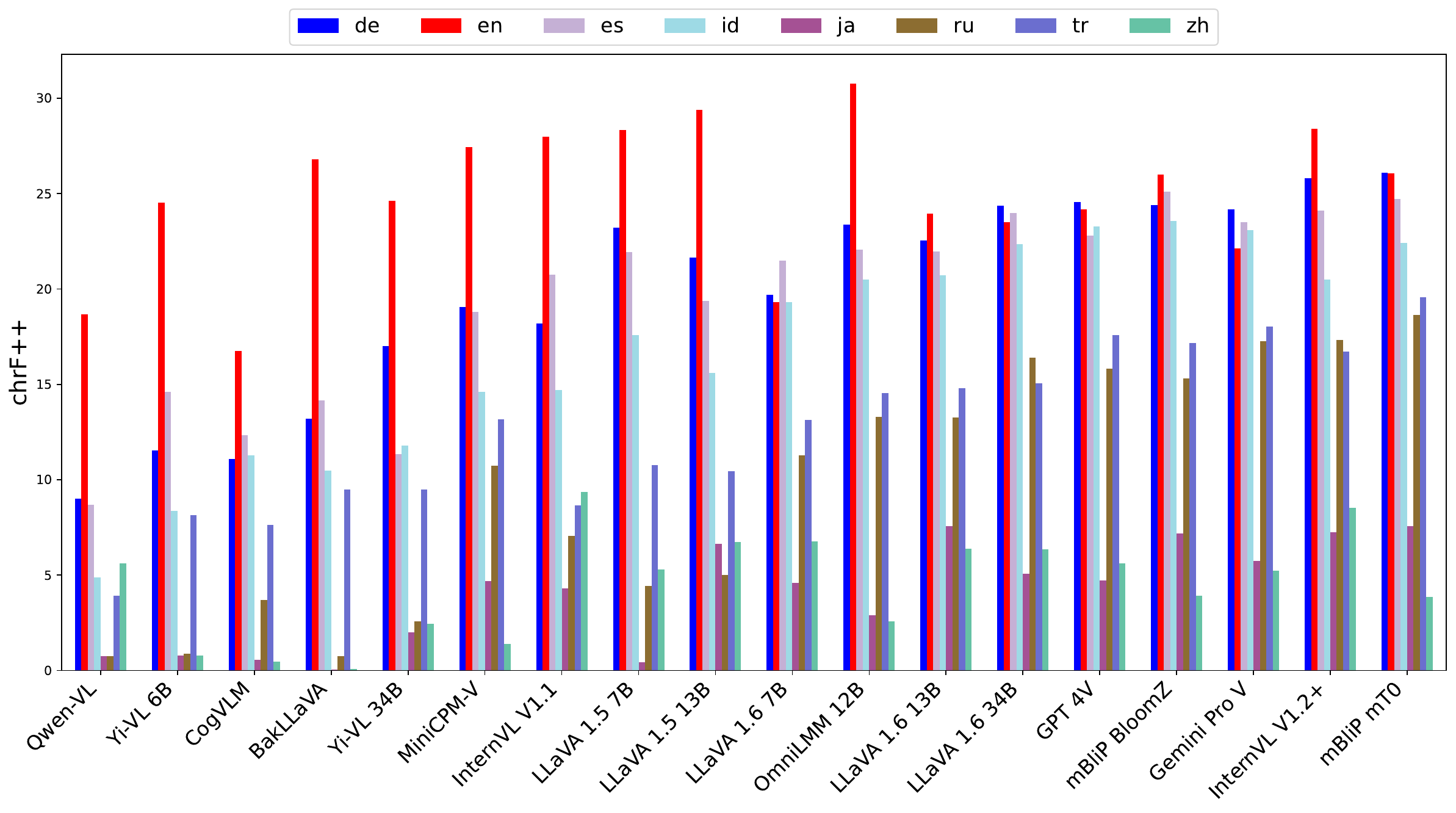}
    \caption{A bar plot showing the average chrF++ score per language and model on the xFlickrCO dataset. The models on the x-Axis are ordered by their average score across all languages so that the best performing model is on the right and the worst is on the left.}
    \label{fig:result_details_xflickrco_plot}
\end{figure*}
\begin{table}[ht!]
    \centering
    \caption{The average chrF++ score per language and model on the xFlickrCO dataset. The column ``NEA'' stands for the average of Non-English languages.}
    \label{tab:result_details_xflickrco_tab}
    \begin{tabular}{@{}lccccccccc@{}}
    \toprule
    \multicolumn{1}{c}{Model} & \multicolumn{9}{c}{Language} \\
    & de & en & es & id & ja & ru & tr & zh & NEA \\
    \midrule
    Qwen-VL & $9.00$ & $18.68$ & $8.69$ & $4.88$ & $0.77$ & $0.74$ & $3.91$ & $5.62$ & $4.80$ \\
    Yi-VL 6B & $11.53$ & $24.54$ & $14.61$ & $8.37$ & $0.78$ & $0.90$ & $8.15$ & $0.79$ & $6.45$ \\
    CogVLM & $11.08$ & $16.76$ & $12.32$ & $11.27$ & $0.56$ & $3.71$ & $7.62$ & $0.46$ & $6.72$ \\
    BakLLaVA & $13.21$ & $26.79$ & $14.17$ & $10.48$ & $0.06$ & $0.75$ & $9.49$ & $0.09$ & $6.89$ \\
    Yi-VL 34B & $17.02$ & $24.62$ & $11.36$ & $11.79$ & $2.00$ & $2.57$ & $9.50$ & $2.44$ & $8.10$ \\
    MiniCPM-V & $19.05$ & $27.43$ & $18.81$ & $14.62$ & $4.69$ & $10.73$ & $13.18$ & $1.40$ & $11.78$ \\
    InternVL V1.1 & $18.21$ & $27.98$ & $20.74$ & $14.69$ & $4.31$ & $7.07$ & $8.67$ & $9.38$ & $11.87$ \\
    LLaVA 1.5 7B & $23.22$ & $28.32$ & $21.95$ & $17.58$ & $0.44$ & $4.45$ & $10.77$ & $5.29$ & $11.96$ \\
    LLaVA 1.5 13B & $21.66$ & $29.39$ & $19.37$ & $15.59$ & $6.63$ & $5.02$ & $10.45$ & $6.72$ & $12.21$ \\
    LLaVA 1.6 7B & $19.70$ & $19.31$ & $21.48$ & $19.32$ & $4.60$ & $11.27$ & $13.14$ & $6.78$ & $13.75$ \\
    OmniLMM 12B & $23.39$ & $30.76$ & $22.05$ & $20.50$ & $2.89$ & $13.29$ & $14.55$ & $2.59$ & $14.18$ \\
    LLaVA 1.6 13B & $22.55$ & $23.94$ & $21.98$ & $20.73$ & $7.57$ & $13.26$ & $14.79$ & $6.39$ & $15.33$ \\
    LLaVA 1.6 34B & $24.38$ & $23.52$ & $23.98$ & $22.36$ & $5.08$ & $16.40$ & $15.05$ & $6.34$ & $16.23$ \\
    GPT 4V & $24.56$ & $24.17$ & $22.82$ & $23.29$ & $4.73$ & $15.82$ & $17.58$ & $5.60$ & $16.34$ \\
    mBliP BloomZ & $24.39$ & $25.99$ & $25.12$ & $23.56$ & $7.18$ & $15.31$ & $17.16$ & $3.93$ & $16.67$ \\
    Gemini Pro V & $24.17$ & $22.13$ & $23.50$ & $23.10$ & $5.75$ & $17.28$ & $18.03$ & $5.24$ & $16.73$ \\
    InternVL V1.2+ & $25.81$ & $28.41$ & $24.13$ & $20.48$ & $7.25$ & $17.34$ & $16.73$ & $8.54$ & $17.18$ \\
    mBliP mT0 & $26.10$ & $26.07$ & $24.74$ & $22.41$ & $7.56$ & $18.64$ & $19.58$ & $3.87$ & $17.56$ \\ \midrule
    Average & $19.95$ & $24.93$ & $19.55$ & $16.95$ & $4.05$ & $9.70$ & $12.69$ & $4.53$ & $12.49$ \\
    \bottomrule
    \end{tabular}
\end{table}
\newpage
\subsubsection{XM3600}
\begin{figure*}[ht!]
    \centering
    \includegraphics[width=1.\linewidth]{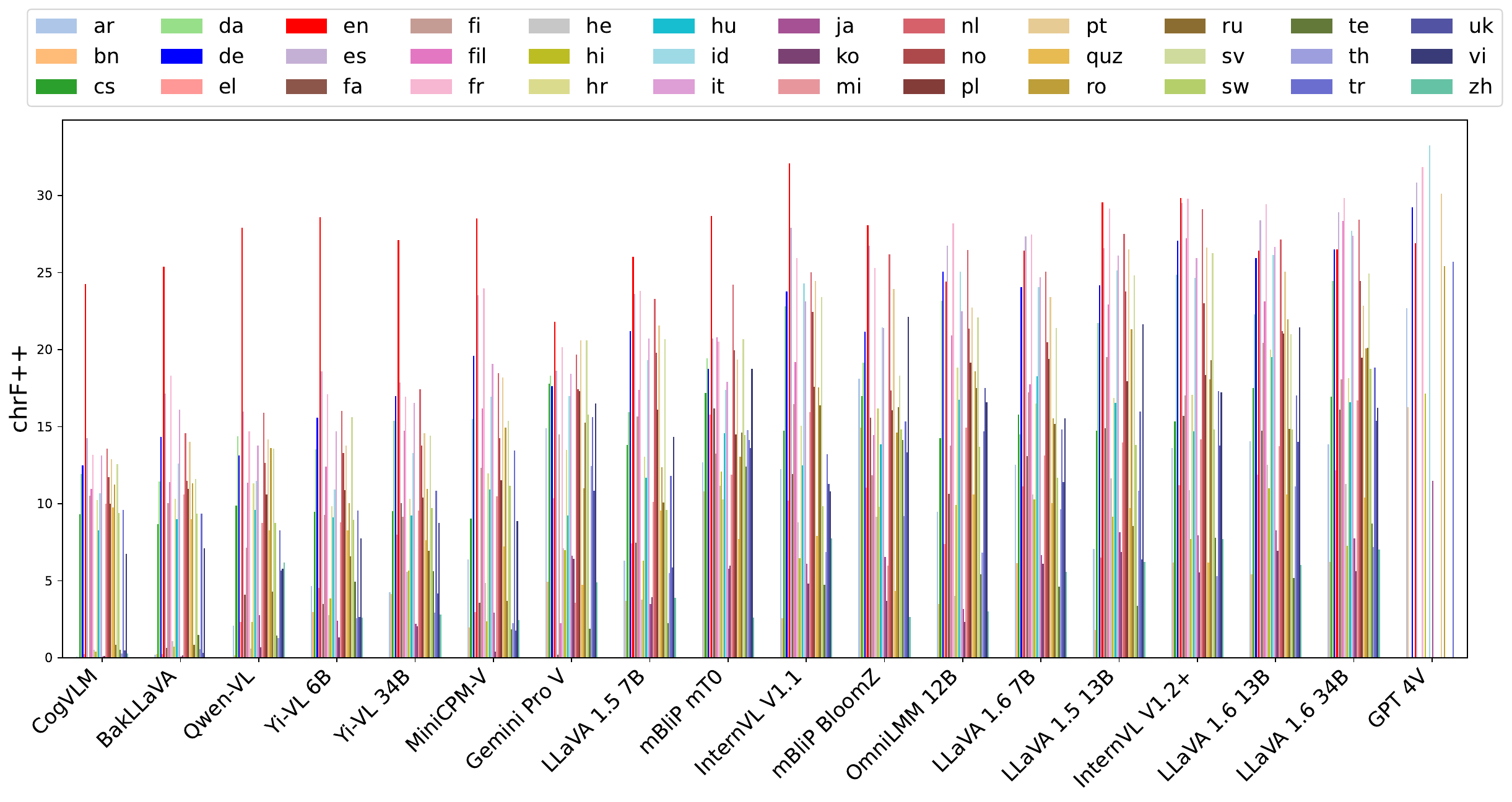}
    \caption{A bar plot showing the average chrF++ score per language and model on the XM3600 dataset. Due to resource restrictions, we evaluated GPT 4V only on a subset of languages. The models on the x-Axis are ordered by their average score across all languages so that the best performing model is on the right and the worst is on the left.}
    \label{fig:result_details_xm3600_plot}
\end{figure*}
\newpage
\begin{table}[ht!]
    \scriptsize
    \centering
    \caption{The average chrF++ score per language and model on the XM3600 dataset. Due to resource restrictions, we evaluated GPT 4V only on a subset of languages. The column ``NEA'' stands for the average of Non-English languages.}
    \label{tab:result_details_xm3600_tab}
    \begin{tabular}{@{}lcccccccccccc@{}}
        \toprule
        \multicolumn{1}{c}{Model} & \multicolumn{12}{c}{Language} \\
        & ar & bn & cs & da & de & el & en & es & fa & fi & fil & fr \\
        \midrule
        CogVLM & $0.07$ & $0.04$ & $9.30$ & $11.92$ & $12.50$ & $0.25$ & $24.26$ & $14.25$ & $0.02$ & $10.52$ & $10.96$ & $13.18$ \\
        BakLLaVA & $0.21$ & $0.22$ & $8.65$ & $11.45$ & $14.33$ & $0.24$ & $25.39$ & $17.13$ & $0.64$ & $10.02$ & $11.41$ & $18.33$ \\
        Qwen-VL & $2.08$ & $0.17$ & $9.89$ & $14.38$ & $13.14$ & $2.32$ & $27.89$ & $16.00$ & $4.09$ & $7.13$ & $11.36$ & $14.70$ \\
        Yi-VL 6B & $4.65$ & $2.98$ & $9.48$ & $13.55$ & $15.58$ & $4.54$ & $28.59$ & $18.58$ & $3.50$ & $9.29$ & $12.42$ & $17.12$ \\
        Yi-VL 34B & $4.24$ & $4.14$ & $9.52$ & $15.40$ & $17.00$ & $8.00$ & $27.11$ & $17.86$ & $10.06$ & $9.17$ & $14.73$ & $16.93$ \\
        MiniCPM-V & $6.38$ & $1.96$ & $9.05$ & $15.52$ & $19.60$ & $2.98$ & $28.53$ & $23.54$ & $3.57$ & $12.33$ & $16.19$ & $23.98$ \\
        Gemini Pro V & $14.90$ & $4.94$ & $17.79$ & $18.32$ & $17.63$ & $10.36$ & $21.81$ & $18.64$ & $0.21$ & $14.50$ & $2.25$ & $20.15$ \\
        LLaVA 1.5 7B & $6.30$ & $3.71$ & $13.80$ & $15.93$ & $21.18$ & $7.42$ & $26.02$ & $23.60$ & $7.45$ & $15.67$ & $17.38$ & $23.83$ \\
        mBliP mT0 & $12.68$ & $10.79$ & $17.20$ & $19.43$ & $18.74$ & $15.76$ & $28.68$ & $20.71$ & $16.19$ & $13.26$ & $20.79$ & $20.52$ \\
        InternVL V1.1 & $12.23$ & $2.55$ & $14.74$ & $22.82$ & $23.77$ & $10.20$ & $32.10$ & $27.91$ & $11.94$ & $16.47$ & $19.20$ & $25.95$ \\
        mBliP BloomZ & $18.10$ & $14.92$ & $16.99$ & $19.16$ & $21.17$ & $11.03$ & $28.05$ & $26.73$ & $15.59$ & $11.86$ & $14.47$ & $25.28$ \\
        OmniLMM 12B & $9.48$ & $3.51$ & $14.24$ & $23.15$ & $25.05$ & $7.37$ & $24.42$ & $26.75$ & $10.65$ & $13.78$ & $20.92$ & $28.18$ \\
        LLaVA 1.6 7B & $12.52$ & $6.13$ & $15.79$ & $14.50$ & $24.06$ & $11.11$ & $26.41$ & $27.37$ & $13.07$ & $17.23$ & $17.76$ & $27.48$ \\
        LLaVA 1.5 13B & $7.07$ & $1.80$ & $14.75$ & $21.74$ & $24.15$ & $6.49$ & $29.55$ & $26.59$ & $14.90$ & $19.51$ & $22.91$ & $29.14$ \\
        InternVL V1.2+ & $13.59$ & $6.19$ & $15.34$ & $24.85$ & $27.05$ & $11.20$ & $29.84$ & $29.50$ & $15.69$ & $17.01$ & $27.22$ & $29.80$ \\
        LLaVA 1.6 13B & $14.07$ & $5.42$ & $17.51$ & $22.30$ & $25.95$ & $11.90$ & $26.42$ & $28.39$ & $14.72$ & $20.44$ & $23.14$ & $29.42$ \\
        LLaVA 1.6 34B & $13.85$ & $6.20$ & $16.94$ & $24.44$ & $26.51$ & $12.17$ & $26.52$ & $28.90$ & $16.09$ & $18.08$ & $28.35$ & $29.83$ \\
        GPT 4V & $22.67$ & $16.27$ & - & - & $29.24$ & - & $26.89$ & $30.86$ & - & - & - & $31.82$ \\ \midrule
        Average & $9.73$ & $5.11$ & $12.83$ & $17.16$ & $20.92$ & $7.41$ & $27.14$ & $23.52$ & $8.80$ & $13.13$ & $16.19$ & $23.65$ \\
        \bottomrule
    \end{tabular}
    \begin{tabular}{@{}lcccccccccccc@{}}
        \toprule
        \multicolumn{1}{c}{Model} & \multicolumn{12}{c}{Language} \\
        & he & hi & hr & hu & id & it & ja & ko & mi & nl & no & pl \\
        \midrule
        CogVLM & $0.52$ & $0.38$ & $10.25$ & $8.25$ & $10.70$ & $13.11$ & $0.07$ & $0.13$ & $10.00$ & $13.59$ & $11.73$ & $9.98$ \\
        BakLLaVA & $1.07$ & $0.71$ & $10.33$ & $8.98$ & $12.59$ & $16.12$ & $0.07$ & $0.16$ & $10.62$ & $14.56$ & $11.48$ & $10.97$ \\
        Qwen-VL & $0.58$ & $2.32$ & $11.33$ & $9.60$ & $11.50$ & $13.76$ & $2.75$ & $0.70$ & $8.73$ & $15.91$ & $12.64$ & $10.59$ \\
        Yi-VL 6B & $2.78$ & $3.86$ & $9.82$ & $9.12$ & $10.90$ & $14.69$ & $2.40$ & $1.32$ & $8.81$ & $16.04$ & $13.30$ & $10.88$ \\
        Yi-VL 34B & $5.58$ & $5.64$ & $10.31$ & $9.23$ & $13.30$ & $16.55$ & $2.21$ & $2.02$ & $9.55$ & $17.43$ & $13.79$ & $10.40$ \\
        MiniCPM-V & $4.86$ & $2.36$ & $11.96$ & $10.91$ & $16.94$ & $19.06$ & $2.92$ & $0.39$ & $10.49$ & $18.47$ & $14.27$ & $11.51$ \\
        Gemini Pro V & $7.12$ & $6.98$ & $13.48$ & $9.22$ & $16.98$ & $18.44$ & $6.63$ & $6.43$ & $3.55$ & $19.67$ & $17.43$ & $17.29$ \\
        LLaVA 1.5 7B & $3.76$ & $6.29$ & $13.05$ & $11.69$ & $19.33$ & $20.73$ & $3.48$ & $3.93$ & $10.10$ & $23.30$ & $19.79$ & $16.10$ \\
        mBliP mT0 & $11.16$ & $12.08$ & $10.26$ & $14.59$ & $17.39$ & $17.92$ & $5.79$ & $6.00$ & $11.88$ & $24.20$ & $19.97$ & $14.49$ \\
        InternVL V1.1 & $8.80$ & $6.47$ & $15.05$ & $12.49$ & $24.31$ & $23.13$ & $6.09$ & $4.83$ & $15.93$ & $25.02$ & $22.45$ & $17.58$ \\
        mBliP BloomZ & $9.16$ & $16.18$ & $9.78$ & $13.84$ & $21.44$ & $21.39$ & $6.53$ & $3.67$ & $5.99$ & $26.17$ & $17.35$ & $16.07$ \\
        OmniLMM 12B & $3.99$ & $9.91$ & $18.84$ & $16.72$ & $25.07$ & $22.50$ & $3.16$ & $2.31$ & $14.94$ & $26.47$ & $21.36$ & $19.16$ \\
        LLaVA 1.6 7B & $10.61$ & $10.26$ & $16.52$ & $18.26$ & $24.05$ & $24.71$ & $6.66$ & $6.09$ & $13.12$ & $25.07$ & $20.49$ & $19.38$ \\
        LLaVA 1.5 13B & $11.63$ & $9.13$ & $16.87$ & $16.54$ & $25.13$ & $26.11$ & $8.16$ & $6.86$ & $13.98$ & $27.52$ & $23.77$ & $17.96$ \\
        InternVL V1.2+ & $10.88$ & $7.69$ & $17.07$ & $14.70$ & $24.65$ & $25.94$ & $7.96$ & $5.53$ & $14.17$ & $29.11$ & $23.02$ & $18.37$ \\
        LLaVA 1.6 13B & $12.54$ & $11.00$ & $19.99$ & $19.52$ & $26.15$ & $26.66$ & $8.27$ & $6.95$ & $13.73$ & $27.15$ & $21.19$ & $21.03$ \\
        LLaVA 1.6 34B & $11.30$ & $7.27$ & $18.16$ & $16.57$ & $27.69$ & $27.40$ & $7.75$ & $5.60$ & $16.69$ & $28.42$ & $24.45$ & $19.49$ \\
        GPT 4V & - & $17.16$ & - & - & $33.24$ & - & $11.46$ & - & - & - & - & - \\\midrule
        Average & $6.46$ & $7.54$ & $12.95$ & $12.23$ & $20.08$ & $19.35$ & $5.13$ & $3.50$ & $10.68$ & $21.01$ & $17.14$ & $14.51$ \\
        \bottomrule
    \end{tabular}
    \addtolength{\tabcolsep}{-.1425em}
    \begin{tabular}{@{}lccccccccccccc@{}}
        \toprule
        \multicolumn{1}{c}{Model} & \multicolumn{13}{c}{Language} \\
        & pt & quz & ro & ru & sv & sw & te & th & tr & uk & vi & zh & NEA \\
        \midrule
        CogVLM & $12.87$ & $9.75$ & $11.23$ & $0.86$ & $12.57$ & $9.41$ & $0.51$ & $0.26$ & $9.58$ & $0.46$ & $6.74$ & $0.29$ & $7.04$ \\
        BakLLaVA & $14.00$ & $9.00$ & $11.30$ & $0.85$ & $11.61$ & $9.37$ & $1.47$ & $0.57$ & $9.36$ & $0.31$ & $7.11$ & $0.03$ & $7.58$ \\
        Qwen-VL & $14.17$ & $8.25$ & $13.60$ & $4.30$ & $13.59$ & $8.75$ & $1.44$ & $1.28$ & $8.26$ & $5.66$ & $5.76$ & $6.20$ & $8.20$ \\
        Yi-VL 6B & $13.77$ & $8.25$ & $10.04$ & $6.57$ & $15.64$ & $8.94$ & $4.93$ & $2.57$ & $9.55$ & $2.65$ & $7.76$ & $2.61$ & $8.82$ \\
        Yi-VL 34B & $14.57$ & $7.64$ & $10.95$ & $6.95$ & $14.42$ & $9.71$ & $5.62$ & $2.92$ & $10.84$ & $4.19$ & $8.74$ & $2.82$ & $9.78$ \\
        MiniCPM-V & $18.21$ & $7.21$ & $14.94$ & $3.69$ & $15.36$ & $11.16$ & $1.83$ & $2.24$ & $13.47$ & $1.74$ & $8.88$ & $2.46$ & $10.30$ \\
        Gemini Pro V & $20.60$ & $4.72$ & $10.98$ & $15.27$ & $20.60$ & $15.80$ & $1.87$ & $12.45$ & $15.62$ & $10.82$ & $16.48$ & $4.88$ & $12.37$ \\
        LLaVA 1.5 7B & $21.57$ & $9.55$ & $12.38$ & $10.08$ & $20.68$ & $9.59$ & $2.23$ & $5.51$ & $11.78$ & $5.84$ & $14.34$ & $3.87$ & $12.44$ \\
        mBliP mT0 & $19.35$ & $7.70$ & $13.05$ & $14.63$ & $20.66$ & $14.45$ & $12.42$ & $14.76$ & $14.13$ & $13.60$ & $18.73$ & $2.59$ & $14.80$ \\
        InternVL V1.1 & $24.47$ & $7.91$ & $17.55$ & $16.39$ & $23.40$ & $9.82$ & $4.73$ & $6.85$ & $13.22$ & $11.26$ & $10.80$ & $7.76$ & $14.97$ \\
        mBliP BloomZ & $23.93$ & $4.32$ & $14.59$ & $16.25$ & $18.31$ & $14.82$ & $14.12$ & $9.19$ & $15.34$ & $13.35$ & $22.14$ & $2.65$ & $15.20$ \\
        OmniLMM 12B & $22.75$ & $10.61$ & $18.61$ & $17.49$ & $22.09$ & $13.68$ & $5.41$ & $6.84$ & $14.68$ & $17.49$ & $16.58$ & $3.00$ & $15.34$ \\
        LLaVA 1.6 7B & $23.42$ & $10.04$ & $15.55$ & $15.18$ & $21.42$ & $11.69$ & $4.60$ & $9.62$ & $14.81$ & $11.40$ & $15.54$ & $5.58$ & $15.46$ \\
        LLaVA 1.5 13B & $26.51$ & $9.70$ & $21.33$ & $8.53$ & $24.80$ & $13.81$ & $3.39$ & $10.84$ & $15.98$ & $6.36$ & $21.66$ & $6.22$ & $16.05$ \\
        InternVL V1.2+ & $26.63$ & $6.20$ & $18.06$ & $19.30$ & $26.27$ & $14.83$ & $7.79$ & $5.30$ & $17.30$ & $13.79$ & $17.22$ & $7.71$ & $17.05$ \\
        LLaVA 1.6 13B & $25.07$ & $10.60$ & $21.96$ & $14.86$ & $21.01$ & $14.80$ & $5.18$ & $11.11$ & $17.03$ & $14.03$ & $21.44$ & $6.02$ & $17.44$ \\
        LLaVA 1.6 34B & $22.85$ & $10.39$ & $20.08$ & $20.11$ & $24.92$ & $18.73$ & $8.70$ & $7.19$ & $18.83$ & $15.36$ & $16.23$ & $7.02$ & $17.79$ \\
        GPT 4V & $30.13$ & - & $25.41$ & - & - & - & - & - & $25.70$ & - & - & - & $24.91$ \\\midrule
        Average & $20.83$ & $7.88$ & $15.65$ & $10.63$ & $18.19$ & $11.63$ & $4.79$ & $6.08$ & $14.19$ & $8.24$ & $13.12$ & $3.98$ & $13.64$ \\
        \bottomrule
    \end{tabular}
\end{table}
\subsection{Language Fidelity Analysis}
\begin{table}[!ht]
\centering
\caption{Pearson correlation coefficients between language fidelity on xFlickrCO and Performance on other datasets.}
\label{tab:results_language_fidelity_analyses_correlations}
\begin{tabular}{@{}lccccccccc@{}}
\toprule
\textbf{Dataset} & \multicolumn{9}{c}{\textbf{Language}} \\
\midrule
& Avg.   
& zh   
& en    
& de    
& id   
& ja   
& ru    
& es   
& tr  \\
\midrule
xFlickrCO        & .91  & .85 & .65  & 0.86 & .88 & .91  & .92  & .90  & .84 \\
XM3600           & .81  & .74 & .63  & 0.63 & .69 & .74  & .76  & .67  & .82 \\
MaXM             & .55  & .17 & .43  & -    & -   & -    & -    & -    & -    \\
XVNLI            & .51  & -   & .46  & -    & -   & -    & .47  & .20  & -    \\
MaRVL            & .46  & .21 & .41  & -    & .50 & -    & -    & -    & .50 \\
M5-VGR          & .34  & -   & .11  & 0.15 & -   & -    & .42  & -    & -    \\
xGQA             & .21  & .35 & .47  & 0.08 & .37 & -    & -.04 & -    & -    \\
M5-VLOD         & .14  & -   & .44  & 0.20 & -   & -    & .14  & -    & -   \\
\bottomrule
\end{tabular}
\end{table}
\end{document}